\documentclass[fleqn,10pt]{wlscirep}

\usepackage{mdframed, multirow}
\newcommand{\probP}{\text{I\kern-0.15em P}}
\usepackage{natbib}

\setcitestyle{numbers}
\usepackage{algorithm}
\usepackage{algpseudocode}
\usepackage{xr}

\usepackage{tcolorbox} 
\tcbuselibrary{skins, breakable}

\newcommand{\munc}{M$_{\text{unc}}$ }
\newcommand{\mcov}{M$_{\text{cov}}$ }
\newcommand{\cunc}{C$_{\text{unc}}$ }
\newcommand{\ccov}{C$_{\text{cov}}$ }
\newcounter{mybox}

\title{Generating crossmodal gene expression from cancer histopathology improves multimodal AI predictions}

\author[1]{Samiran Dey}
\author[2,3]{Christopher R.S. Banerji}
\author[1]{Partha Basuchowdhuri}
\author[4]{Sanjoy K. Saha}
\author[2,5,6]{Deepak Parashar}
\author[2,7,*]{Tapabrata Chakraborti}

\affil[1]{School of Mathematical \& Computational Sciences, Indian Association for the Cultivation of Science, Kolkata, India}
\affil[2]{The Alan Turing Institute, London, UK}
\affil[3]{Comprehensive Cancer Center, King's College London, London, UK}
\affil[4]{Department of Computer Science and Engineering, Jadavpur University, Kolkata, India}
\affil[5]{MRC Biostatistics Unit, University of Cambridge, Cambridge, UK}
\affil[6]{Department of Biostatistics, Bioinformatics and Biomathematics, Georgetown University, Washington DC, USA}
\affil[7]{UCL Cancer Institute, Dept of Medical Physics \& Biomedical Engineering, University College London, London, UK}

\affil[*]{Correspondence to: Tapabrata Chakraborti (\href{tchakraborti@turing.ac.uk}{tchakraborty@turing.ac.uk; t.chakraborty@ucl.ac.uk})}

\begin{abstract}
Emerging research has highlighted that artificial intelligence-based multimodal fusion of digital pathology and transcriptomic features can improve cancer diagnosis (grading/subtyping) and prognosis (survival risk) prediction. However, such direct fusion is impractical in clinical settings, where histopathology remains the gold standard and transcriptomic tests are rarely requested in public healthcare. We experiment on two publicly available multimodal datasets, The Cancer Genomic Atlas and the Clinical Proteomic Tumor Analysis Consortium, spanning four independent cohorts: glioma-glioblastoma, renal, uterine, and breast, and observe significant performance gains in gradation and risk estimation ($p$-value $<0.05$) when incorporating synthesized transcriptomic data with WSIs. Also, predictions using synthesized features were statistically close to those obtained with real transcriptomic data ($p$-value $>0.05$), consistently across cohorts. Here we show that with our diffusion based crossmodal generative AI model, PathGen, gene expressions synthesized from digital histopathology jointly predict cancer grading and patient survival risk with high accuracy (state-of-the-art performance), certainty (through conformal coverage guarantee) and interpretability (through distributed co-attention maps). PathGen code is available on GitHub at \url{https://github.com/Samiran-Dey/PathGen} for open use.

\end{abstract}

\begin{document}

\maketitle

\section*{INTRODUCTION}
\label{sec_intro}

Histopathological assessment is the gold standard for cancer diagnosis and provides information essential for accurate staging to inform prognosis. However, aging populations, strained healthcare services and the increasing diagnostic and prognostic complexity of modern medicine are contributing to an unsustainable workload for practising histopathologists \cite{workload}. Modern artificial intelligence (AI) algorithms, particularly deep learning models, can predict cancer grade and sub-types from digital hematoxylin and eosin (H\&E) stained pathology slides, complementing and supporting pathology workflows and potentially reducing the high workload burden \cite{grader1, grader2, grader4, grader5, grader6, grader7}. Very recently the emergence of pathology foundation models using large vision transformers (ViT) has greatly increased model flexibility, enabling support with a wide variety of downstream tasks and providing greater utility for pathologists \cite{uni, mi_zero, conch}.

Classically, histopathology has relied on microscopic assessment of H\&E-stained tissue sections, often complemented by targeted panels of immunohistochemical and other special stains, to confirm and refine diagnoses. As technology has advanced, a broader range of molecular tests has become routine for assessing certain tumours. These include genetic analyses such as IDH mutation status, which is required to appropriately subtype gliomas according to the WHO \cite{who} and EGFR mutation status which is essential to guide therapy decisions in non-small cell lung cancers \cite{egfr}. More recently, a range of transcriptomic tests have been developed, which show clear evidence for guiding treatment decisions in specific malignancies \cite{chemo}. Following extensive clinical evaluation, several of these tests have been granted regulatory approval for clinical use \cite{preci}. However, guidelines, especially in public healthcare systems, strictly limit the application of these transcriptomic tests to a small minority of patients, largely due to their high financial cost and infrastructural requirements \cite{cost}.

Recently, AI models employing a multimodal fusion of digital pathology and transcriptomic features, trained on large public datasets, like The Cancer Genomic Atlas (TCGA), have led to significant improvements in the predictive accuracy of both grading and survival risk estimation, compared to unimodal models employing digital H\&E slides alone \cite{riskr5, pathomic, mcat}. Though these powerful models have the potential to convey substantial health benefits, they are developed in a setting disjoint from the current reality of the clinic. Translation of these models into most healthcare systems is infeasible at present, as the required transcriptomic data is very seldom collected. Future translation requires rigorous, costly and widespread clinical evaluation to determine which patients would benefit from these models sufficiently to justify the cost of transcriptomic data. 

Crossmodal generative AI models can be trained to synthesize one unavailable data modality (transcriptomics) from routinely available modalities (digital pathology) \cite{trans_gen, SEQUOIA, trans_hist}. Coupled with a multimodal AI model, this could facilitate a cost-effective screening tool to estimate the added value of collecting transcriptomic data for diagnostic/prognostic assessment of each individual patient. Such an approach could support cost-efficient clinical evaluation, and sustainable translation of multimodal AI models to support pathology workflows.

In this work, we introduce a diffusion based crossmodal generative model, PathGen \cite{code}, that synthesizes transcriptomic data from digital images of H$\&$E-stained slides. We evaluate our model on two publicly available multimodal datasets: The Cancer Genomic Atlas (TCGA) \cite{tcga} and the Clinical Proteomic Tumor Analysis Consortium (CPTAC) \cite{cptac}, using data from four independent cancer cohorts: glioma-glioblastoma, renal, uterine and breast. We find that the generated transcriptomic data are highly similar to the corresponding real transcriptomic data in a hold-out test set. We further demonstrate that the predictive performance of both diagnosis (cancer grading) and prognosis (survival risk) significantly improves when the synthesised transcriptomic data is combined with features learned from the corresponding digital pathology image, using a state-of-the-art ViT-based model through a co-attention mechanism. 
For practical utility, our model must be trusted by clinicians and attune to emerging regulation \cite{eu_ai_act}. We thus provide transparency via detailed attention based heatmaps for both grading and survival risk prediction, allowing pathologists to visually affirm the clinical relevance of the regions contributing most towards the AI decision. We provide uncertainty quantification, via conformal analysis \cite{conformal}, to allow pathologist users a measure of certainty in the added value of the synthesized transcriptomic data for each patient. Finally, to evaluate algorithmic fairness, we apply our conformal uncertainty quantification to assess model performance analysis across various patient demographics. Our work builds on preliminary studies in crossmodal prediction \cite{uni, pathomic, mcat}, to provide a road map for sustainable, safe and cost-effective translation of multimodal AI models supporting cancer diagnostics/prognostics, to the clinic.

\section*{RESULTS}
\label{sec_result}

\subsection*{Dataset and study design}
\label{sec_data}
Our experiments use publicly available data from TCGA and CPTAC, specifically from four independent cancer cohorts - Glioblastoma Multiforme and Brain Lower Grade Glioma (TCGA-GBM, TCGA-LGG, CPTAC-GBM), Kidney Renal Clear Cell Carcinoma (TCGA-KIRC), Uterine Corpus Endometrial Carcinoma (TCGA-UCEC, CPTAC-UCEC), and Breast Invasive Carcinoma (TCGA-BRCA). For TCGA-GBMLGG, 912 whole slide images (WSIs) coming from 745 cases are used, of which 183 cases belong to TCGA-GBM \cite{tcga_gbm} and 562 cases belong to TCGA-LGG \cite{tcga_lgg}. We consider 485 WSIs from 462 cases for TCGA-KIRC \cite{tcga_kirc}, 294 WSIs from 267 cases for TCGA-UCEC \cite{tcga_ucec}, 1010 WSIs from 946 cases for TCGA-BRCA \cite{tcga-brca}, 242 WSIs from 62 cases for CPTAC-GBM \cite{cptac-gbm}, and 364 WSIs from 71 cases for CPTAC-UCEC \cite{cptac-ucec} for our experiments. The slides have been split into training, validation, calibration, and test sets (Supplementary Table \ref{s-data_split}). The models trained using the TCGA data have been evaluated on the corresponding CPTAC cohorts, without any further training or finetuning. Cases with missing grade, survival data or transcriptomic data were excluded. The data distribution across various demographic groups like gender, age, censorship status, etc. varies for all cohorts (Supplementary Table \ref{s-data_demo}). 

Transcriptomic features are selected to match those employed in the successful multimodal models \cite{pathomic, mcat}. As proposed previously \cite{mcat}, transcriptomic data for each patient is divided into 6 broad categories of genes - tumour suppressor genes, oncogenes, protein kinases, cell differentiation markers, transcription factors, and cytokines and growth factors. However, a gene may occur in multiple gene groups. The number of available gene expression levels for each gene group differs across cohorts but remains same for the TCGA and CPTAC datasets for the same cancer cohort (Supplementary Table \ref{s-data_trans}).  When discussing gene expression levels throughout this paper, we refer to transcript abundance, rather than protein levels.

\begin{figure}[!ht]
    \centerline{\includegraphics[scale=0.17]{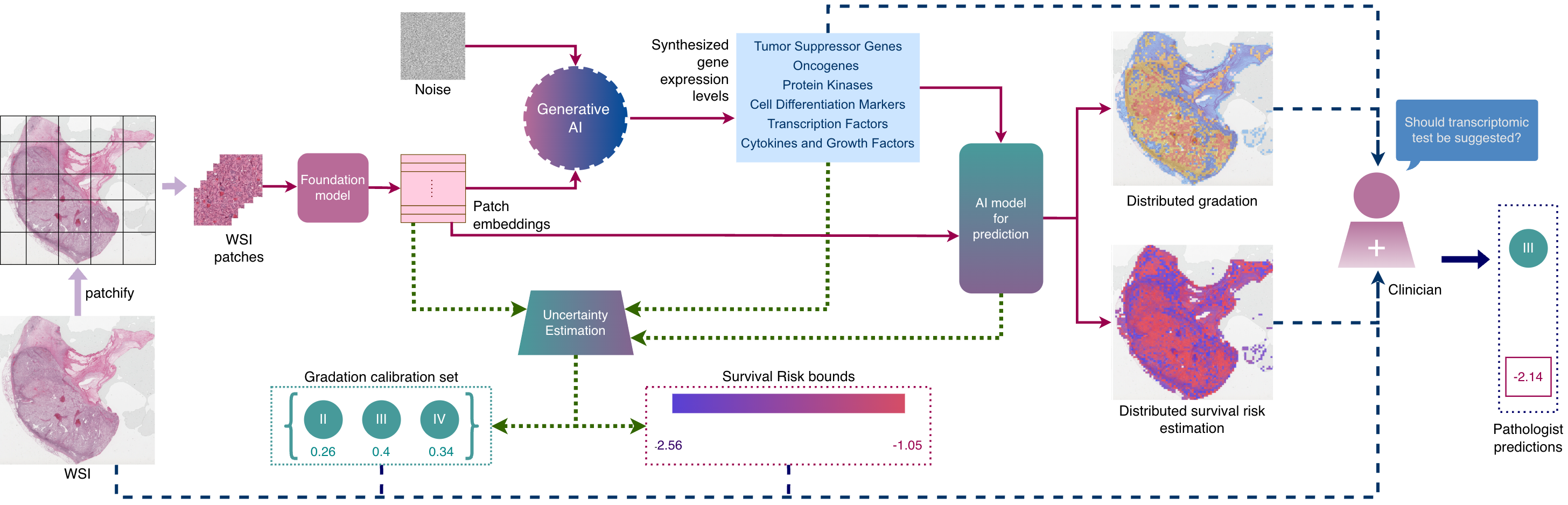}}
    \caption{\textbf{Methodology overview.} Pipeline for the proposed methodology. Our diffusion-based crossmodal generative model, PathGen, synthesizes transcriptomic features from whole slide image patch embeddings obtained using a state-of-the-art foundation model from the whole slide images. Multimodal prediction is performed for both diagnosis (cancer grading) and prognosis (survival risk) using the synthesized transcriptomic data and histopathology images to predict the added value of transcriptomic data, for each patient. Uncertainty quantification provides a patient-level estimate of model reliability and provides clinicians with survival risk bounds and tumour grade prediction sets which are guaranteed to contain the ground truth with a specified probability. Distributed predictions also provide a window to understand the intra-tumour heterogeneity.}
    \label{fig_overview}
\end{figure}

Our methodology synthesizes transcriptomic data from WSIs using our diffusion based generative model, PathGen. Synthesized transcriptomes are then used for automated cancer gradation and survival risk estimation to assess whether transcriptomic tests should be performed (Figure \ref{fig_overview}.c). Transcriptomic data (both real and synthesized) is used at two stages in the predictive model (MCAT\_GR) - to co-attend to the WSI patch embeddings and learn a combined feature map that is suitable to predict both grade and survival, and to predict the survival risk of a patient in a learnable ensemble method along with the co-attended feature maps. We perform extensive experiments using synthesized transcriptomic data to decide on this model setting (Supplementary Table \ref{s-result_gen_use}). The model is trained end-to-end for performing both gradation and survival risk estimation together using the loss function  
$L = \lambda \times L_{survival} + (1-\lambda) \times L_{grade}$, 
where $L_{survival}$ is the loss for survival risk estimation, $L_{grade}$ is the loss for gradation and $\lambda$ is the loss coefficient to balance the emphasis between survival risk estimation and gradation. Experiments show that at $\lambda=0.3$ the model performance is optimal with close and high scores for both gradation and survival risk estimation together across glioblastoma-glioma and renal cohorts (Supplementary Figure \ref{s-fig_lambda} and Supplementary Table \ref{s-result_gen_use}). Thus, for all results reported in the paper, we use $\lambda=0.3$. For uncertainty estimation using conformal prediction, we choose the error rate $\alpha=0.1$. Further details of model architecture, loss function and training methodology are provided in Methods.

\subsection*{Similarity evaluation of synthesized and real transcriptomic data}
\label{sec_realness}
The transcriptomic data synthesized by our model, PathGen, is compared to the real patient transcriptomes to assess similarity. The Spearman correlation coefficient for all gene groups between real and synthesized transcriptomic features evaluates to 0.713 for TCGA-GBMLGG, 0.717 for TCGA-KIRC, 0.436 for TCGA-UCEC, 0.642 for TCGA-BRCA, 0.662 for CPTAC-GBM, and 0.669 for CPTAC-UCEC (Figure \ref{fig_res}.a.i and Supplementary Table \ref{s-result_trans}). Thus, showing highly significant correlation between real and synthesized transcriptomic data (Spearman correlation $p$-value $< 0.05$). The corresponding normalized mean absolute error (nMAE) over all gene groups evaluates to 0.141, 0.160, 0.173, 0.155, 0.167 and 0.178 for TCGA-GBMLGG, TCGA-KIRC, TCGA-UCEC, TCGA-BRCA, CPTAC-GBM and CPTAC-UCEC respectively (Figure \ref{fig_res}.a.i and Supplementary Table \ref{s-result_trans}). Though there exists a high correlation and low MAE between synthetic and real transcriptomic data, there are some variations in correlation across the gene sets and data cohorts (Figures \ref{fig_res}.a.ii, \ref{fig_res}.a.iii and Supplementary Table \ref{s-result_trans}). Notably, synthesized and real oncogene expression levels are highly correlated for glioblastoma-glioma, renal and breast cohorts (Figure \ref{fig_res}.a.iii and Supplementary Table \ref{s-result_trans}). This is particularly beneficial for cancer grading or survival estimation because oncogenes when aberrantly active can serve as biomarkers that drive tumour progression, facilitating the prediction of cancer prognosis \cite{oncogenes}. Thus we conclude that the synthesized transcriptomic data are highly correlated with the real transcriptomic data and we gain confidence towards using the former in the predictive pipeline.

\subsection*{Performance evaluation of synthesized transcriptomic data in multimodal prediction}
\label{sec_mainres}
\begin{figure} [!ht]
    \centerline{\includegraphics[scale=0.084]{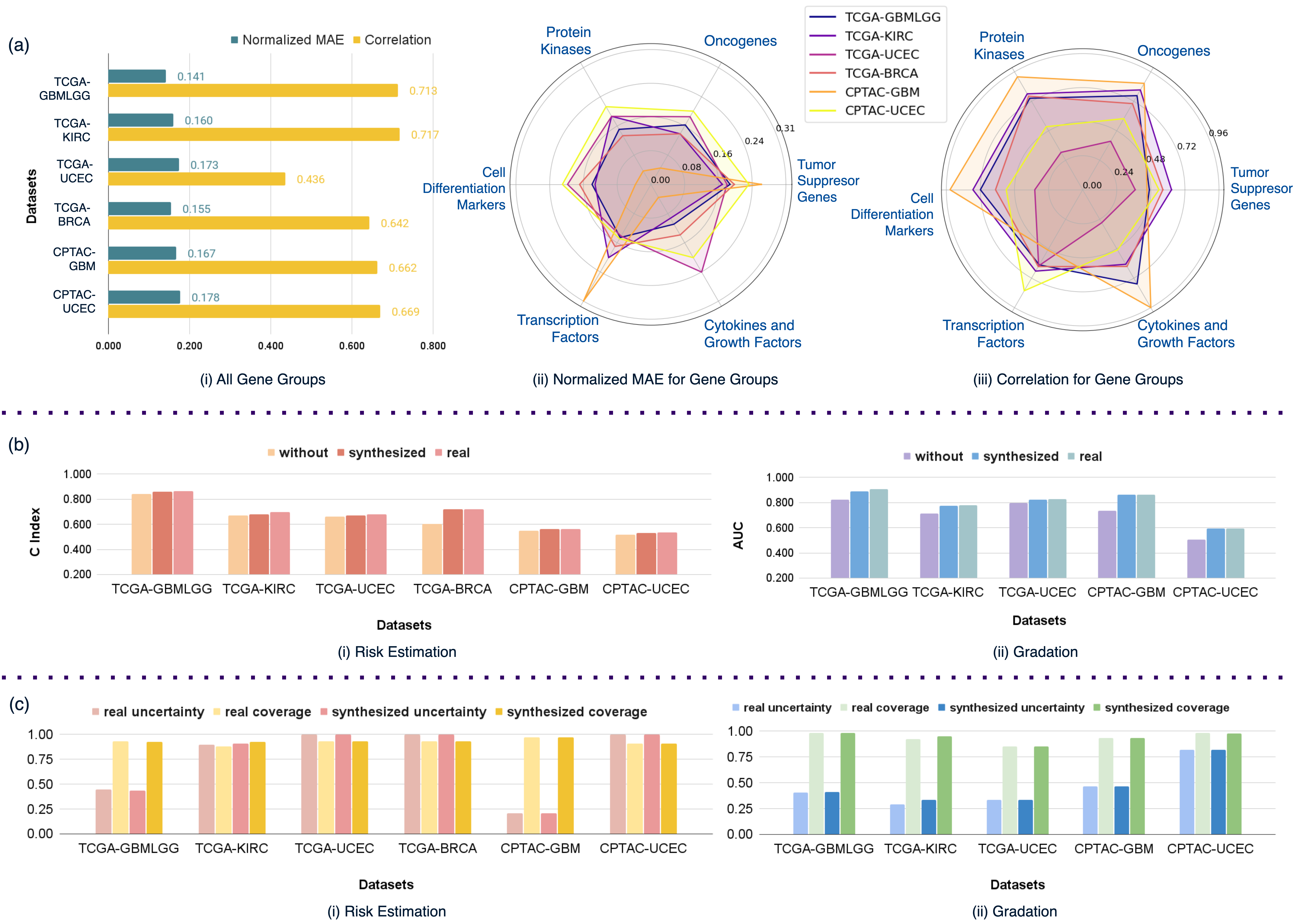}}
    \caption{\textbf{Evaluation of synthesized transcriptomic data}. (a) Comparison between real and synthesized gene expression levels over all genes and for different gene groups. (b) Plot of significant performance improvement on using synthesized transcriptomic data over using WSIs alone (marked as without), and closeness of using synthesized and real transcriptomic data for survival risk estimation and gradation. (c) Plot for comparison of uncertainty and obtained coverage for survival risk and grade predictions using real and synthesized transcriptomic data. Source data are provided as a Source Data file.}
    \label{fig_res}
\end{figure}

We next analyse whether PathGen synthesised transcriptomes significantly add value to automated gradation and survival risk estimation, over and above whole slide images (WSIs) alone. Evaluation metrics used for grade prediction and survival risk estimation are the Area Under Curve (AUC) and the concordance index (C Index) respectively. For survival risk estimation the C Index obtained using WSIs alone are 0.842, 0.671, 0.663, 0.603, 0.547, 0.518 for TCGA-GBMLGG, TCGA-KIRC, TCGA-UCEC, TCGA-BRCA, CPTAC-GBM and CPTAC-UCEC repectively (Figure \ref{fig_res}.b.i and Supplementary Table \ref{s-result_main}). And for gradation, we obtain an AUC of 0.823 for TCGA-GBMLGG, 0.714 for TCGA-KIRC, 0.796 for TCGA-UCEC, 0.736 for CPTAC-GBM, and 0.508 for CPTAC-UCEC (Figure \ref{fig_res}.b.ii and Supplementary Table \ref{s-result_main}). Since TCGA-BRCA does not report cancer grades in the clinical data, only survival risk estimation could be performed. The performance significantly improves when PathGen synthesized transcriptomics are used alongside WSIs (Wilcoxon rank-sum test $p$-value $<0.05$), resulting in an AUC of 0.890 and C Index of 0.861 for TCGA-GBMLGG, AUC of 0.773 and C Index of 0.681 for TCGA-KIRC, AUC of 0.821 and C Index of 0.673 for TCGA-UCEC, C Index of 0.720 for TCGA-BRCA, AUC of 0.865 and C Index of 0.565 for CPTAC-GBM, and AUC of 0.593 and C Index of 0.530 for CPTAC-UCEC (Figures \ref{fig_res}.b.i, \ref{fig_res}.b.ii and Supplementary Table \ref{s-result_main}). A significant improvement was also noted on evaluating jointly over survival and gradation for using synthesized transcriptomic data with WSIs for prediction (ANOSIM test $p$-value $=0.001$, Supplementary Table \ref{s-result_main}), consistent over datasets and cohorts. Thus, in the absence of real transcriptomic data, adding synthesized transcriptomic data significantly improves diagnostic and prognostic predictions compared to using only WSI features.

We further compare gradation and survival risk predictions using WSIs and PathGen synthesized transcriptomic data to predictions using WSIs and real transcriptomic data. Combining real transcriptomic data with WSIs yields an AUC of 0.907 and C Index of 0.866 for TCGA-GBMLGG jointly, AUC of 0.778 and C Index of 0.697 for TCGA-KIRC, AUC of 0.828 and C Index of 0.680 for TCGA-UCEC, C Index of 0.720 for TCGA-BRCA, AUC of 0.863 and C Index of 0.564 for CPTAC-GBM, and AUC of 0.593 and C Index of 0.533 for CPTAC-UCEC and the corresponding result using synthesized transcriptomic data is provided above (Figures \ref{fig_res}.b.i, \ref{fig_res}.b.ii and Supplementary Table \ref{s-result_main}).  Remarkably, model performance on using real transcriptomic data is not significantly different to using PathGen synthesized transcriptomes (Wilcoxon rank-sum test $p$-value $> 0.05$ and ANOSIM test $p$-value $> 0.05$) for gradation and survival risk estimation, both separately and jointly, in all considered TCGA and CPTAC cohorts (Supplementary Table \ref{s-result_main}). Thus, the transcriptomic data synthesized by our model, PathGen, consistently performs comparably for gradation and survival risk estimation with no significant difference on average from real transcriptomic data.

We also compare the performance of our methodology with the state-of-the-art models for using real transcriptomic data and WSI patches. For TCGA-GBMLGG, the C Index evaluated for GSCNN \cite{GSCNN} is 0.781, DeepAttnMISL \cite{DeepAttnMISL} is 0.734, MCAT \cite{mcat} is 0.817, Pathomic fusion \cite{pathomic} is 0.826, whereas our methodology yields a C Index of 0.866 using real transcriptomic data and a C Index of 0.861 using synthesized transcriptomic data for survival risk estimation. PathGen-X \cite{PathoGen_X} obtains a C-Index of 0.81 on the TCGA-GBM cohort; however, their inference stage is unimodal and only considers a section of patients (Grade IV).  For gradation, Pathomic fusion \cite{pathomic} has an AUC of 0.906, whereas our methodology has an AUC of 0.907 on real transcriptomic data and an AUC of 0.890 on synthesized transcriptomic data. For TCGA-KIRC, the C Index evaluated for Pathomic fusion \cite{pathomic} is 0.720 whereas for our methodology is 0.697 on real transcriptomic data and 0.681 on synthesized transcriptomic data. However, Pathomic fusion uses ROIs annotated by expert pathologists for predicting grades, while we use transcriptomic features to co-attend to the WSI patch embeddings without requiring any expert intervention, thus reducing expert annotation overheads. For TCGA-UCEC, DeepSets\cite{deepsets}, DeepAttnMISL\cite{DeepAttnMISL}, and MCAT\cite{mcat} achieve a C Index of 0.593, 0.586, and 0.622, respectively, whereas our methodology yields a C Index of 0.680 and 0.673 using real and synthesized transcriptomes, respectively, for survival risk estimation. On TCGA-BRCA, our methodology achieved a C-index of 0.720 for both rel and synthesized transcriptomic profiles, compared to 0.522, 0.577, and 0.580 for DeepSets\cite{deepsets}, DeepAttnMISL\cite{DeepAttnMISL}, and MCAT\cite{mcat}, respectively. We note that the considered SOTA models, DeepSets, DeepAttnMISL, and MCAT, do not perform gradation, so a direct comparison on grade prediction was not possible. For similar reasons, we could only compare the performance of our model with Pathomic fusion for TCGA-GBMLGG gradation and TCGA-KIRC survival risk estimation. Since our model performs better or at least comparably with the state-of-the-art models using real transcriptomic data, and we observe that the predictive performance between using real or synthesized transcriptomic data is very close with no significant difference for the test set population (Wilcoxon rank-sum test $p$-value $>0.05$, Supplementary Table \ref{s-result_main}), we may conclude that we achieve state-of-the-art performance using gene expression levels synthesized by our model, PathGen.

\subsection*{Transparency evaluation for clinical interpretability}
\label{sec_disres}
\begin{figure}[!ht]
    \centerline{\includegraphics[scale=0.155]{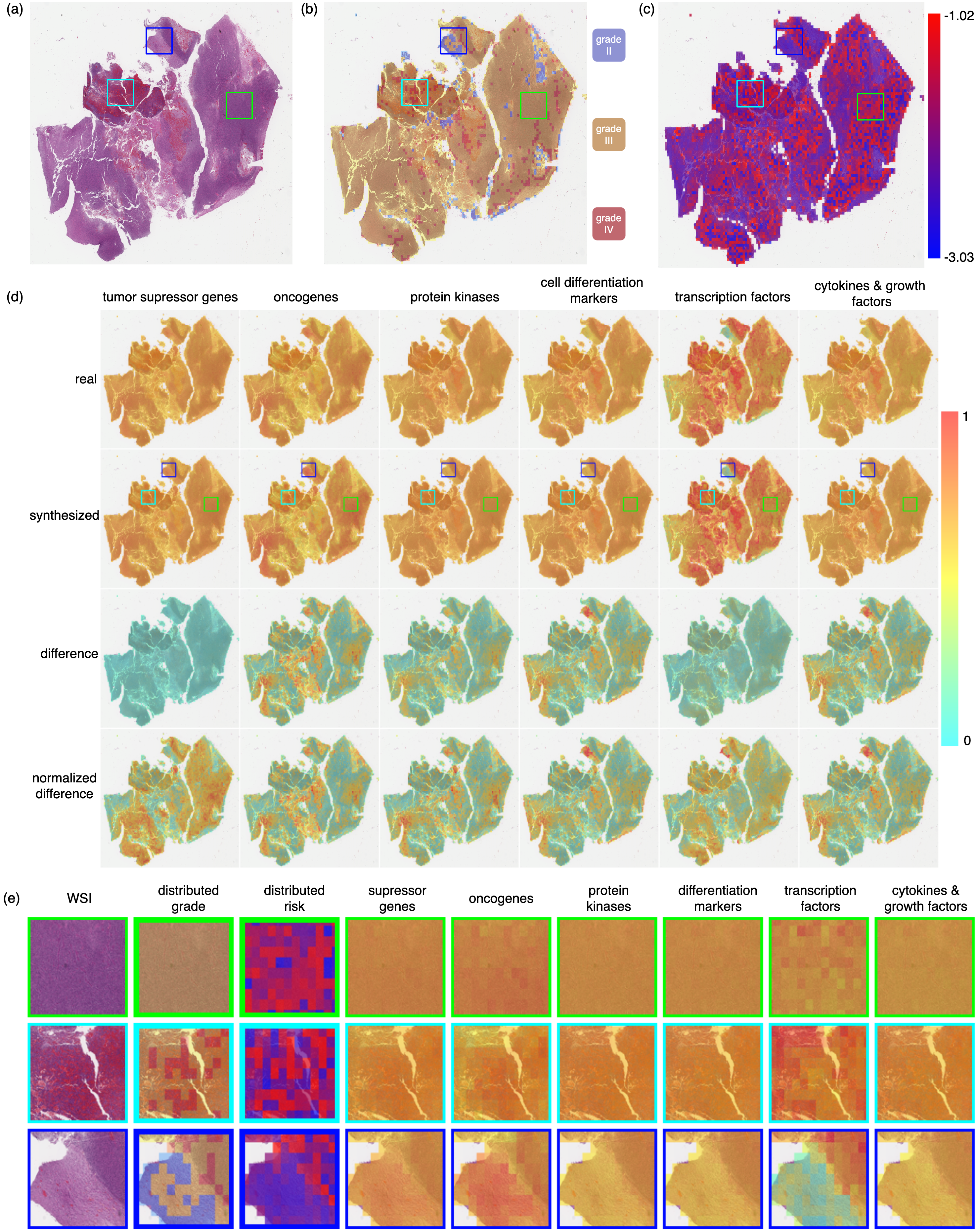}}
    \caption{\textbf{Explainability maps for a male patient of age 58, ground truth grade III, survival time 10.71 months, and survival status alive from the TCGA-LGG cohort}. (a) whole slide image (WSI) (b) intra-tumour gradation heterogeneity (c) intra-tumour survival heterogeneity (d) co-attention maps for real and synthesized transcriptomic data and WSI patches for different gene groups (e) comparative study of intra-tumour heterogeneity and corresponding co-attention maps obtained for prediction using synthesized transcriptomic data for chosen regions marked with different coloured rectangles. }
    \label{fig_gbmlgg}
\end{figure}

\begin{figure}[!ht]
    \centerline{\includegraphics[scale=0.13]{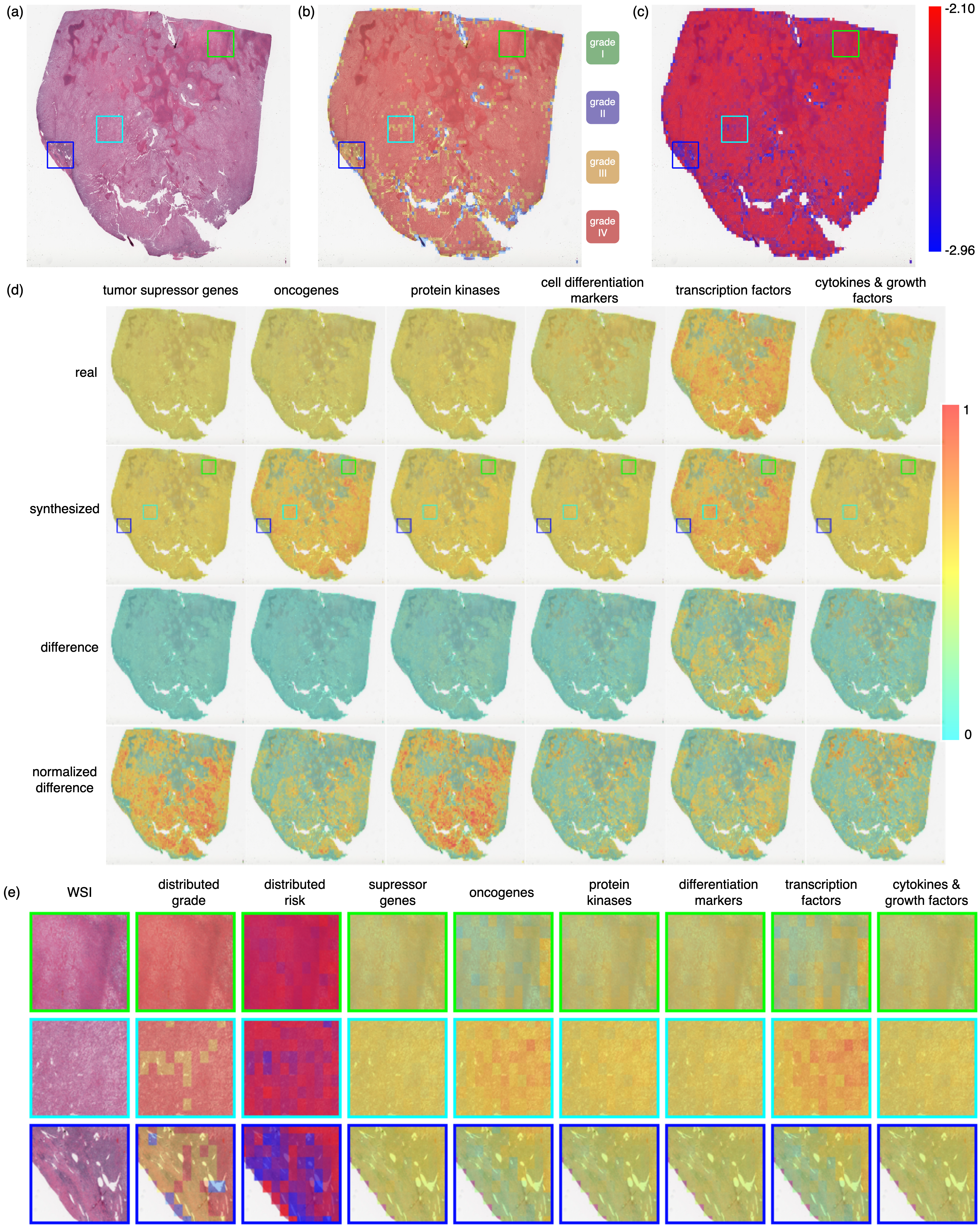}}
    \caption{\textbf{Explainability maps for a male patient of age 40, ground truth grade IV, survival time 33.99 months, and survival status deceased from the TCGA-KIRC cohort}. (a) whole slide image (WSI) (b) intra-tumour gradation heterogeneity (c) intra-tumour survival heterogeneity (d) co-attention maps for real and synthesized transcriptomic data and WSI patches for different gene groups (e) comparative study of intra-tumour heterogeneity and corresponding co-attention maps obtained for prediction using synthesized transcriptomic data for chosen regions marked with different coloured rectangles. }
    \label{fig_kirc}
\end{figure}

To visualize intra-tumour heterogeneity, we use our trained model to predict grade and survival risk for individual WSI patches independently. The distributed prediction allows the identification of regions of varied grades within a WSI, highlighting the heterogeneity for clinical assistance (Figures \ref{fig_gbmlgg}.b, \ref{fig_kirc}.b and Supplementary Figures \ref{s-fig_gbmlgg1}.b, \ref{s-fig_kirc1}.b, \ref{s-fig_cptac_ucec}.b). It is evident that higher grade regions are correlated to higher risks, as is expected (Figure \ref{fig_kirc}.e). By observing distributed predictions at different magnification levels corresponding to a WSI, we find that using WSIs at higher resolution helps to obtain more precise predictions for tissue regions, as is expected (Supplementary Figure \ref{s-fig_mag}). 

\begin{figure} [!ht]
    \centerline{\includegraphics[scale=0.082]{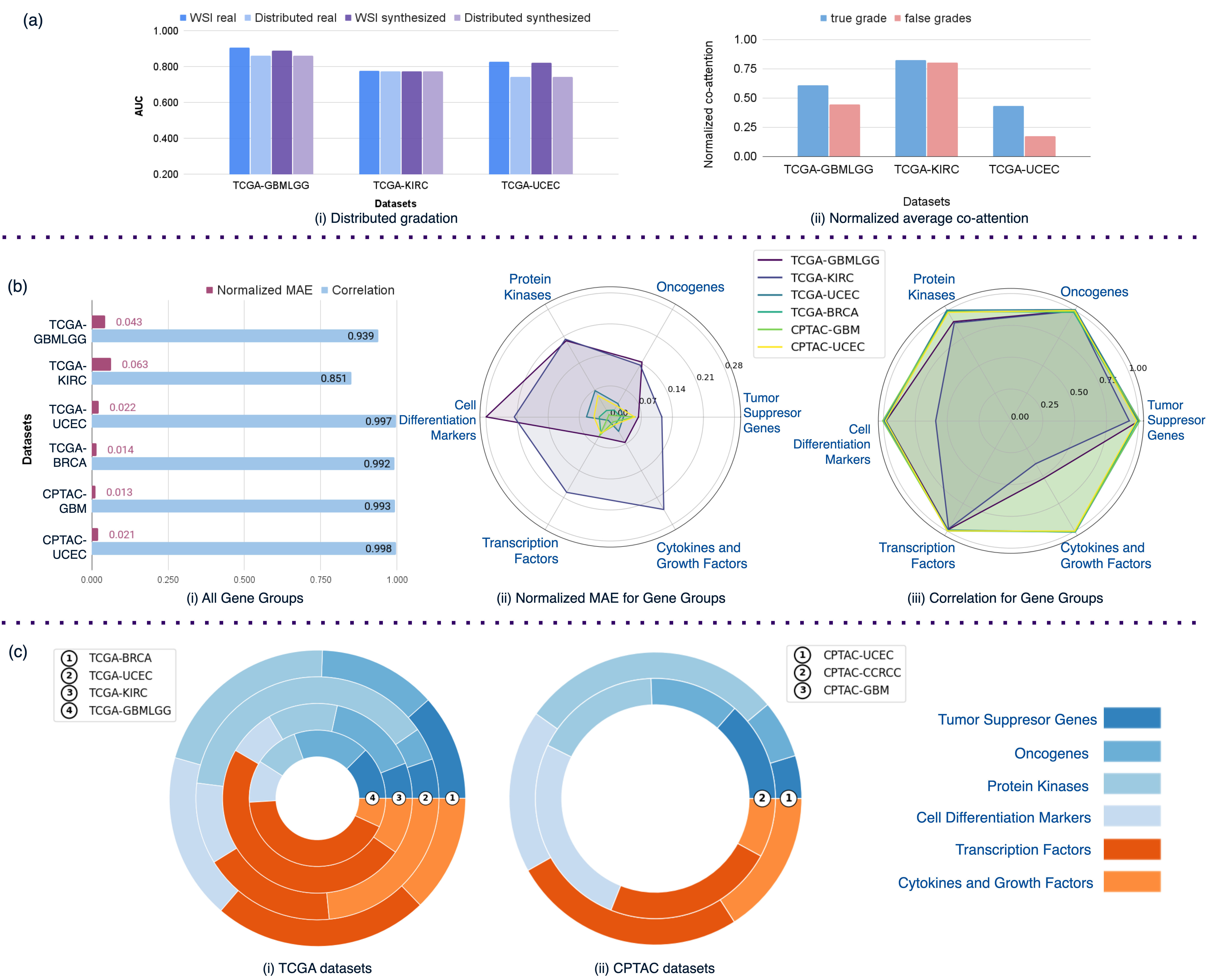}}
    \caption{\textbf{Explainability analysis}. (a) (i) Plot for comparison between distributed and non-distributed WSI predictions for gradation using both real and synthesized transcriptomic data. (ii) Plot showing normalized average co-attention value for the true grade is higher than that of false grades. (b) Comparison between co-attention maps obtained using real and synthesized transcriptomic data over all genes and for different gene groups. (c) Plot illustrating the percentage contribution of the gene groups in co-attention for different TCGA and CPTAC data cohorts. Source data are provided as a Source Data file.}
    \label{fig_dist_res}
\end{figure}

Further, we consider the aggregate of the heterogeneous (distributed) predictions and compare it with the grade predicted by the model, considering all WSI patches as input (non-distributed prediction). For the TCGA cohorts, the distributed gradation performance is not significantly different from a one-time prediction with all WSI patches (Wilcoxon ranksum test $p$-value $>$ 0.05, Figure \ref{fig_dist_res}.a.i). Thus, we see from the results that there is indeed a distributed variation of grade when predicted locally for patches and mapped onto the WSIs as a heatmap, both within and across cohorts, which may provide an additional visual presentation of tumour heterogeneity to the clinician expert (pathologist) to choose whether or not to draw any further intuition from them.

To add explainability, we next study how transcriptomic data co-attends to the corresponding WSI patched. We compare the co-attention maps obtained using real and synthesized gene expression levels using Spearman correlation coefficient and normalized mean absolute error (nMAE). A higher correlation with a high corresponding nMAE would denote that the range of co-attention values is large for the particular gene group but the co-attention map is still correlated. TCGA-GBMLGG obtains a correlation of 0.939 and nMAE of 0.043, TCGA-KIRC obtains a correlation of 0.851 and nMAE of 0.063, TCGA-UCEC obtains a correlation of 0.997 and nMAE of 0.022, TCGA-BRCA obtains a correlation of 0.992 and nMAE of 0.014, CPTAC-GBM obtains a correlation of 0.993 and nMAE of 0.013, and CPTAC-UCEC obtains a correlation of 0.998 and nMAE of 0.021 over all gene groups (Figure \ref{fig_dist_res}.b.i). These significantly high correlation values (Spearman correlation $p$-value $<0.05$) and low nMAEs between co-attention maps for real and synthesized transcriptomic data explains the reason for no significant difference between using the same for gradation and survival risk estimation across cohorts and datasets (Wilcoxon ranksum test $p$-value $>0.05$, ANOSIM test  $p$-value $>0.05$, Supplementary Table \ref{s-result_main}). For all cohorts, co-attention maps attributing to oncogenes are most correlated with a low nMAE, followed by transcription factor genes and tumour suppressor genes (Figures \ref{fig_dist_res}.b.ii, \ref{fig_dist_res}.b.iii). Oncogenes usually drive cell division and survival, thus promoting cancer. Inactivation of tumour suppressor genes can lead to cancer by failing to regulate uncontrolled cell growth \cite{hallmark}. So the general co-attention observations above are not unexpected, although a rigorous inference would need a detailed analysis of gene set functional significance, which is outside the scope of this work. 

We further study the contribution of synthesized transcriptomic features in co-attending to the WSI patch embeddings and observe that there are variances in the percentage contribution of individual gene groups in co-attention for all cohorts and datasets (Figures \ref{fig_dist_res}.c.i,\ref{fig_dist_res}.c.ii, and Supplementary Table \ref{s-result_coattn_perc}). For most cohorts, the transcription factor genes and oncogenes have the highest co-attention values with the WSI patches for both real and synthesized gene expression levels and hence contribute the most to the prediction (Figures \ref{fig_gbmlgg}.d, \ref{fig_kirc}.d, and Supplementary Figures \ref{s-fig_gbmlgg1}.d, \ref{s-fig_kirc1}.d,  \ref{s-fig_cptac_ucec}.d). Since activation of oncogenes is known to drive tumour proliferation \cite{oncogenes}, having these as one of the higher contributors to co-attention would be aligned with clinical expectations.

We next visualise specific regions of the WSI and compare the distributed grades, distributed risk and the co-attention maps between synthesized transcriptomic features and the WSI patches (Figures \ref{fig_gbmlgg}.e, \ref{fig_kirc}.e and Supplementary Figure \ref{s-fig_gbmlgg1}.e, \ref{s-fig_kirc1}.e, \ref{s-fig_cptac_ucec}.e). We observe that ground truth grade regions correspond to higher attention scores and other grade regions correspond to low attention values (Figures \ref{fig_gbmlgg}, \ref{fig_kirc}, \ref{fig_dist_res}.a.ii).

\subsection*{Uncertainty quantification for clinical reliability}
\label{sec_uncres}

We observe that the uncertainty profile of using synthesized transcriptomic data matches that of using real transcriptomic data for both gradation and survival risk estimation, for different patient demographics and stratification across cohorts, by performing conformal prediction (CP) as described in Methods. For TCGA-GBMLGG, the marginal uncertainty (\munc) for the overall data population evaluates to 0.407 and 0.413 for gradation, and 0.446 and 0.438 for survival risk estimation, on using real and synthesized transcriptomic data, respectively (Figure \ref{fig_res}.c). For TCGA-UCEC, a \munc of 0.337 for gradation and 1.0 for survival is obtained, for both real and synthesized transcriptomic data (Figure \ref{fig_res}.c). For TCGA-KIRC, gradation \munc increases from  0.292 to 0.334, and the survival \munc increases from 0.898 to 0.909 on using synthesized transcriptomic data over their real counterparts (Figure \ref{fig_res}.c). However the changes in \munc are not significant, proving the similarity in the uncertainty profile between the use of real and synthesized transcriptmic data (Wilcoxon rank sum test $p$-value $>0.05$, Supplementary Tables \ref{s-result_GBMLGG}, \ref{s-result_GBM}, \ref{s-result_KIRC}, \ref{s-result_TUCEC}, \ref{s-result_CUCEC}). 

Furthermore, we observe that the uncertainty profile remains similar across different demographic groups for real and synthesised transcriptomic data, with no significant difference in uncertainty, by performing stratified conditional conformal prediction as described in Methods (Wilcoxon rank-sum test $p$-value $>0.05$, Supplementary Tables \ref{s-result_GBMLGG}, \ref{s-result_GBM}, \ref{s-result_KIRC}, \ref{s-result_TUCEC}, \ref{s-result_CUCEC}). Additionally, stratifying by demographic categories helps achieve fairness in the distribution of coverage across the population by ensuring each group within the category independently obtains the desired coverage. We find variations in both conditional uncertainty (\cunc) across demographic categories and marginal uncertainty (\munc) across groups in a category, which is expected (Supplementary Tables \ref{s-result_GBMLGG}, \ref{s-result_GBM}, \ref{s-result_KIRC}, \ref{s-result_TUCEC}, \ref{s-result_CUCEC}). For instance, in comparison to the overall \munc of 0.413, when stratified by gender, \cunc decreases to 0.312 (real) and 0.323 (synthesized), and when stratified by time bin, it decreases to 0.335 (real) and 0.331 (synthesized), for TCGA-GBMLGG (Supplementary Table \ref{s-result_GBMLGG}).

We also find that models trained on the TCGA datasets adapt and calibrate well over the corresponding CPTAC cohorts, not always with a drastic increase in uncertainty, which otherwise is expected. The model trained on TCGA-GBMLGG when calibrated and tested over CPTAC-GBM has a \munc of 0.465 for gradation and 0.206 for survival for both real and synthesized transcriptomic data (Figure \ref{fig_res}.c). On calibrating and testing the model trained on TCGA-UCEC over CPTAC-UCEC, a gradation \munc of 0.821 for real, and 0.818 for synthesized transcriptomic data and a survival \munc of 1.0 for both real and synthesized transcriptomes is observed (Figure \ref{fig_res}.c).

It may be observed that for a few population groups across some cohorts, the obtained coverage does not always match the desired coverage because of fewer samples in the calibration set. These coverage slacks are expected and can be theoretically computed as a function of the cardinality of the calibration set \cite{conformal}. We find that all obtained coverages that are less than the desired coverage of 0.9 fall within the expected deviation for the corresponding calibration set size of the demographic group (Supplementary Tables \ref{s-data_demo}, \ref{s-table_deviation}). For many such cases, using synthesized data has increased the coverage at the cost of an insignificant increase in uncertainty (Wilcoxon rank-sum test $p$-value $>0.05$, Supplementary Tables \ref{s-result_GBMLGG}, \ref{s-result_GBM}, \ref{s-result_KIRC}, \ref{s-result_TUCEC}, \ref{s-result_CUCEC}). As risk estimation directly uses the transcriptomic data for prediction and not just for co-attention, such differences in obtained coverage are expected even for minimal differences in the gene expression levels between the synthesized and real transcriptomic features. Moreover, the uncertainty estimates are derived from conformal sets for gradation and risk bounds for survival estimation, which helps to understand the extent to which we may rely on the model predictions. Additionally, the gradation conformal sets provide a range of alternative grades to consider before making the final decision.

\begin{figure} [!ht]
    \centerline{\includegraphics[scale=0.129]{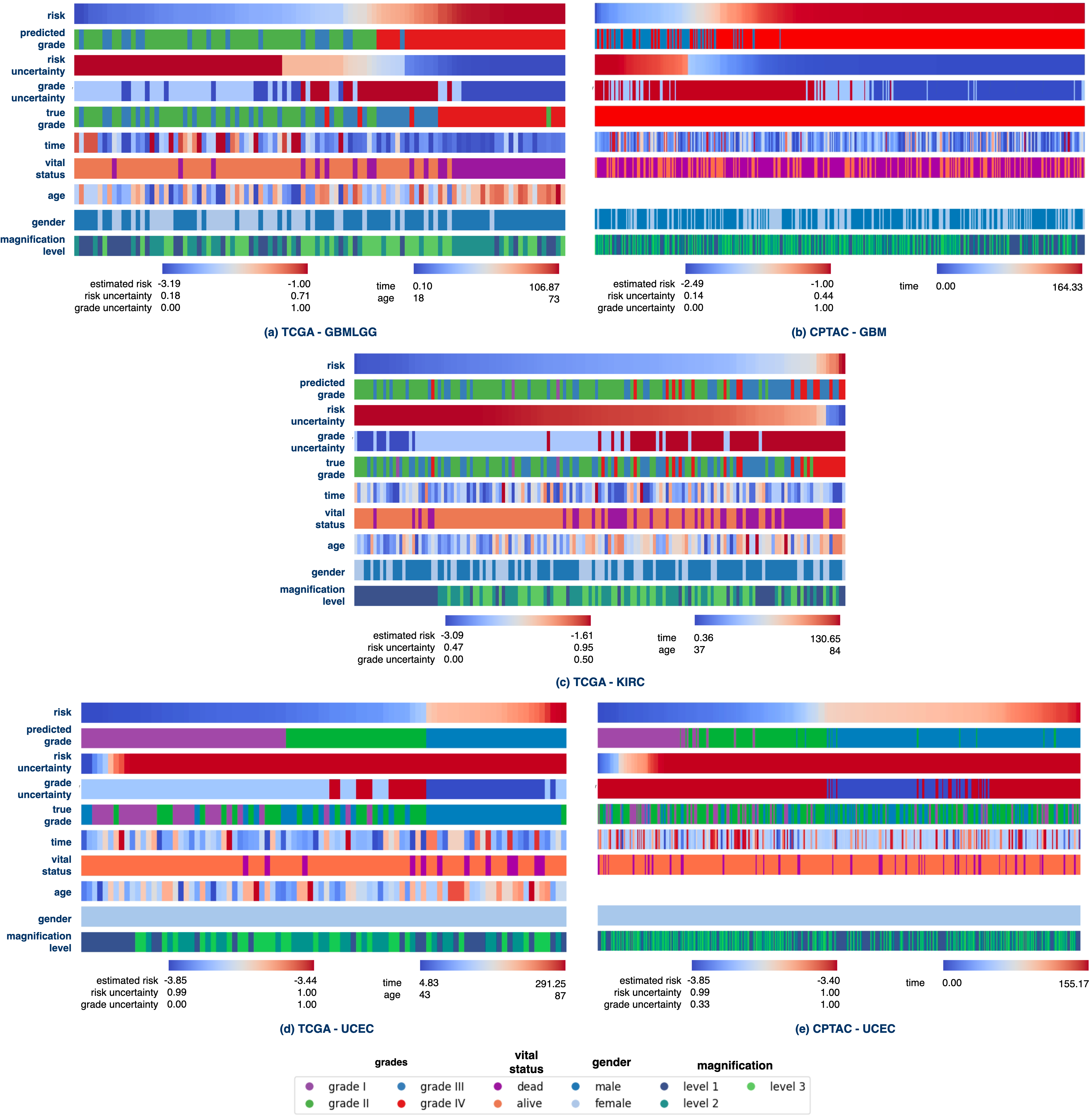}}
    \caption{\textbf{Illustrative result for evaluating fairness in model predictions across demographic categories and data cohorts.} The cases have been arranged in ascending order of predicted patient risk. The estimated uncertainty in survival risk and grade prediction is also shown. Source data are provided as a Source Data file.}
    \label{fig_fairness}
\end{figure}

Next, we investigate algorithmic fairness by assessing whether the performance in gradation and survival risk predictions shows demographic biases. We observe that the predicted risk is less for patients who are reported to be alive and more for deceased patients, which is as expected (Figures \ref{fig_fairness}.a, \ref{fig_fairness}.c, \ref{fig_fairness}.d). Also, the uncertainty of survival risk estimation decreases with increasing risk (Figures \ref{fig_fairness}.a, \ref{fig_fairness}.b, \ref{fig_fairness}.c), which is expected because the actual survival time of patients who are alive is beyond the reported survival time. Thus, the model estimated survival risk should correspond to a time beyond the reported survival time and have a higher uncertainty measure when compared in terms of reported survival time.  It may also be observed that the model tends to correlate higher risks with the highest grade, for all data cohorts (Figure \ref{fig_fairness}). In general, the uncertainty of survival risk estimation is always greater if the survival time reported for a living patient is less (Figure \ref{fig_fairness}). Also, the uncertainty of prediction for alive patients is mostly high, except for gradation in TCGA-KIRC (Figure \ref{fig_fairness} and Supplementary Tables \ref{s-result_GBMLGG}, \ref{s-result_GBM}, \ref{s-result_KIRC}, \ref{s-result_TUCEC}, \ref{s-result_CUCEC}). The uncertainty of grade and survival prediction is more for male patients than females except for gradation in TCGA-GBMLGG (Supplementary Tables \ref{s-result_GBMLGG}, \ref{s-result_GBM}, \ref{s-result_KIRC}). The uncertainty of risk estimation decreases with age for all TCGA cohorts (Supplementary Tables \ref{s-result_GBMLGG}, \ref{s-result_KIRC}, \ref{s-result_TUCEC}). Thus, these results provide an analysis from a health equity and predictive fairness point of view as to how the performance varies across patient demographics.

\section*{DISCUSSION}
\label{sec_discussion} 

In this paper, we present PathGen, a crossmodal generative diffusion model for synthesizing transcriptomic features from WSIs of H$\&$E stained slides. Our results are based on assessment of two publicly available cancer datasets - TCGA and CPTAC, using data from four cancer cohorts: glioma-glioblastoma, renal, uterine and breast. We demonstrated that PathGen synthesized transcriptomic features are highly correlated to true patient transcriptomes. By using a multimodal AI model to predict cancer grade and survival risk from WSIs and transcriptomic features, with state-of-the-art performance, we demonstrated that combining the features learnt from WSIs using a vision transformer based foundation model with the synthesized transcriptomic features from PathGen, significantly improved predictive accuracy compared to WSIs alone. Remarkably, we found that there was no significant difference in accuracy when real patient transcriptomic data were used instead of PathGen synthesized transcriptomic data in the multimodal model, demonstrating that PathGen captures the independent gradation/survival information in true patient gene expressions. To further check for generalizability, the models trained on the TCGA dataset were evaluated on the same CPTAC cohorts. We observed that our claims that synthesized transcriptomic data provides significant added value over WSI-only predictions, and that predictions based on real and synthesized transcriptomic data remain significantly close, generalizes well across independent datasets and cohorts. 

H$\&$E stained slides of tumour tissue are essential for diagnostic and prognostic assessment of malignancy, and are collected routinely. However, transcriptomic tests, though they have the potential to inform prognosis and guide treatment selection \cite{chemo, preci} are not routinely performed, especially in the public sector \cite{cost}. This is due to a number of reasons, including economic cost, infrastructural requirements and incomplete knowledge about patient benefit. PathGen can synthesize realistic and informative patient transcriptomic features, from routine and inexpensive WSIs of H$\&$E stained slides. Also, we note the average inference time for cross-modal synthesis of transcriptomic data from WSIs using PathGen is 21.15 seconds, and the average time for estimating survival and grade is 0.71 seconds, which varies over the magnification level and the size of the WSI, as expected. Thus, in the future, we envision that PathGen can be studied as a low cost screening tool, to guide the targeted collection of transcriptomic data modalities, by potentially selecting patients who would benefit from transcriptomic assessment for more accurate diagnosis/prognosis. This would of course need a dedicated follow up study which would include patient response to treatment when selected in this manner, that would entail a separate clinical validation project.

Moreover, to be useful in the clinic, AI models must be trusted \cite{explainability}. Thus, to add explainability to the model predictions, we have undertaken measures with design choices guided by emerging regulatory requirements (e.g., EU AI Act Articles 13–15) \cite{eu_ai_act} and in consultation with a practising histopathologist based at University College Hospitals, London. Firstly, by employing co-attention tools for model transparency, we visualize the role of the gene groups and WSI patches in predicting survival and grades, and by mapping distributed gradation and survival risk estimates back onto the WSI images, we provide a window into intra-tumour heterogeneity. These facilitate additional visual presentations to the clinician experts to choose whether or not to draw any further intuition from them. Secondly, we have employed conformal prediction for assessment of model reliability, providing not simply point estimates for gradation and survival risk, but ranges that contain true values with a specified certainty. This analysis has also allowed us to assess model fairness by comparing model reliability across patient demographics such as age and gender. Importantly, we found that there was no significant difference in model reliability or fairness when using PathGen synthesized transcriptomic features compared to real patient transcriptomes. For reproducibility of research and ethos of open science, we have made the code fully available online through our GitHub repository \cite{code}.

However, though the PatheGen synthesized transcriptomic features corresponding to the six considered gene groups (both synthesized and real) contain important diagnostic/prognostic information, they do not capture the full diversity of the transcriptome. There is likely a limit to how much of the full transcriptome can be synthesized from only knowledge of WSIs. Also, evaluating the clinically important but less abundant synthesized gene expressions would require substantial domain expertise. Hence, we do not claim that PathGen replaces the need for real transcriptomic data, but rather it provides a potential way of estimating the benefit of collecting the full data across the population. The increasing use of techniques such as spatial transcriptomics \cite{spatial} will provide insight into this question, as well as WSI patch level resolution of transcriptomics, to guide development of synthesized transcriptomic data with a greater number of features in the near future.

\section*{METHODS}
\label{sec_methods}

The paper aims to use synthesized transcriptomic data for the automated selection of whole slide image (WSI) features for risk estimation and gradation of cancer cases, and evaluate the value added by the use of synthesized transcriptomic data over WSI-only predictions and the difference in using real transcriptomic data instead (Figure \ref{fig_overview}). The WSIs are patchified and the corresponding patch embeddings are obtained using UNI \cite{uni}. Our   model, PathGen, generates transcriptomic features from the patch embeddings. Together, the patch embeddings and the gene embeddings obtained from the transcriptomic features are thus used by MCAT\_GR for gradation and risk estimation. In the following subsections, we discuss the steps of the proposed methodology in detail.

\subsection*{Preprocessing multimodal data}
\label{sec_prepro}

\textbf{Histopathology data.} The diagnostic WSIs for a patient are divided into 224 $\times$ 224 patches, as required for obtaining patch embeddings using the pre-trained foundation model, UNI. The magnification level of the WSI is randomly chosen from the available options to make the model magnification level agnostic. The patches having normalized mean intensity more than 0.8 are excluded. It is observed that such patches consist mostly of the background region and hence, they do not contribute to the prediction. We do not use stain normalization in preprocessing the WSIs in line with recent works that find no significant effect of using stain normalization as a preprocessing step \cite{stain_norm}. The 224 $\times$ 224 patches are provided as input to UNI to obtain the patch embeddings of dimension 1 $\times$ 1024. UNI \cite{uni} is a pre-trained vision encoder for histopathology developed using private datasets, and thus is not based on the public datasets used in our experiments.

\textbf{Transcriptomic data.} As proposed in MCAT \cite{mcat}, the corresponding transcriptomic data of the patient is divided into 6 broad categories of genes - tumour suppressor genes, oncogenes, protein kinases, cell differentiation markers, transcription factors, and cytokines and growth factors. The gene expression levels are z-score normalized if not already done. Our model, PathGen, is trained to synthesize such transcriptomic data from the corresponding WSIs. The gene encoders are trained to obtain embeddings of dimension 1 $\times$ 1024 for the gene expression levels of each category. 

\subsection*{Synthesizing transcriptomic data from histopathology images}
\label{sec_PathGen}
\subsubsection*{Background of diffusion}
\label{sec_backdiff}

The process of diffusion is represented as a Markov chain where, in the forward process, noise is added in each step to degrade the the data for a predetermined timestep $T$ \cite{sohl2015deep}. Diffusion models are trained to predict the noise introduced in each step from a noisy sample. In the reverse process, the noise is given as input to the diffusion model and denoising is performed for $T$ timesteps to obtain the synthesized data \cite{ho2020denoising}.

The forward process begins with a data sample $\mathbf{x}_0$ and gradually adds Gaussian noise at each timestep $t$ \cite{song2020score} following the equation,
\begin{equation}
    q(\mathbf{x}_t | \mathbf{x}_{t-1}) = \mathcal{N}(\mathbf{x}_t; \sqrt{1 - \beta_t} \mathbf{x}_{t-1}, \beta_t \mathbf{I}),
\end{equation}
where $\beta_t$ is a predefined noise variance schedule. For any given timestep $t$, the sample $\mathbf{x}_t$ is obtained directly from the initial data $\mathbf{x}_0$ as per the following equation,
\begin{equation}
    q(\mathbf{x}_t | \mathbf{x}_0) = \mathcal{N}(\mathbf{x}_t; \sqrt{\bar{\alpha}_t} \mathbf{x}_0, (1 - \bar{\alpha}_t) \mathbf{I}),
\end{equation}
where $\bar{\alpha}_t = \prod_{s=1}^t (1 - \beta_s)$ \cite{ho2020denoising}. This simplifies to the equation,
\begin{equation}
    \mathbf{x}_t = \sqrt{\bar{\alpha}_t} \mathbf{x}_0 + \sqrt{1 - \bar{\alpha}_t} \epsilon,
\end{equation}
with $\epsilon \sim \mathcal{N}(0, \mathbf{I})$.

The reverse process aims to denoise the sample back to the ground truth data. A learnable model is used to predict the noise at each step to obtain the sample $\mathbf{x}_{t-1}$ from $\mathbf{x}_t$ \cite{song2020score}.
\begin{equation}
    p_\theta(\mathbf{x}_{t-1} | \mathbf{x}_t) = \mathcal{N}(\mathbf{x}_{t-1}; \mu_\theta(\mathbf{x}_t, t), \sigma_\theta(\mathbf{x}_t, t)),
    \label{equ_diffinf}
\end{equation}
where $\mu_\theta$ and $\sigma_\theta$ are learned parameters and given the following equations, 
\begin{equation}
    \mu_\theta(\mathbf{x}_t, t) := \frac{1}{\sqrt{\alpha_t}} \left( \mathbf{x}_t - \frac{\beta_t}{\sqrt{1 - \bar{\alpha}_t}} \epsilon_\theta(\mathbf{x}_t, t) \right),
\end{equation}
\begin{equation}
    \sigma_\theta(\mathbf{x}_t, t) := \tilde{\beta}_t^{\frac{1}{2}}, \quad t \in \{T, T-1, \dots, 1\}.
\end{equation}

The model is trained to minimize the difference between the true noise $\epsilon$ and the predicted noise $\epsilon_\theta$ using the simplified equation,
\begin{equation}
    L_{diff} = \mathbb{E}_{t, \mathbf{x}_0, \epsilon} \left[ \|\epsilon - \epsilon_\theta(\mathbf{x}_t, t)\|^2 \right].
    \label{equ_diffloss}
\end{equation}

\subsubsection*{Architecture of PathGen}
\label{sec_archdiff}
\begin{figure}
    \centerline{\includegraphics[scale=0.262]{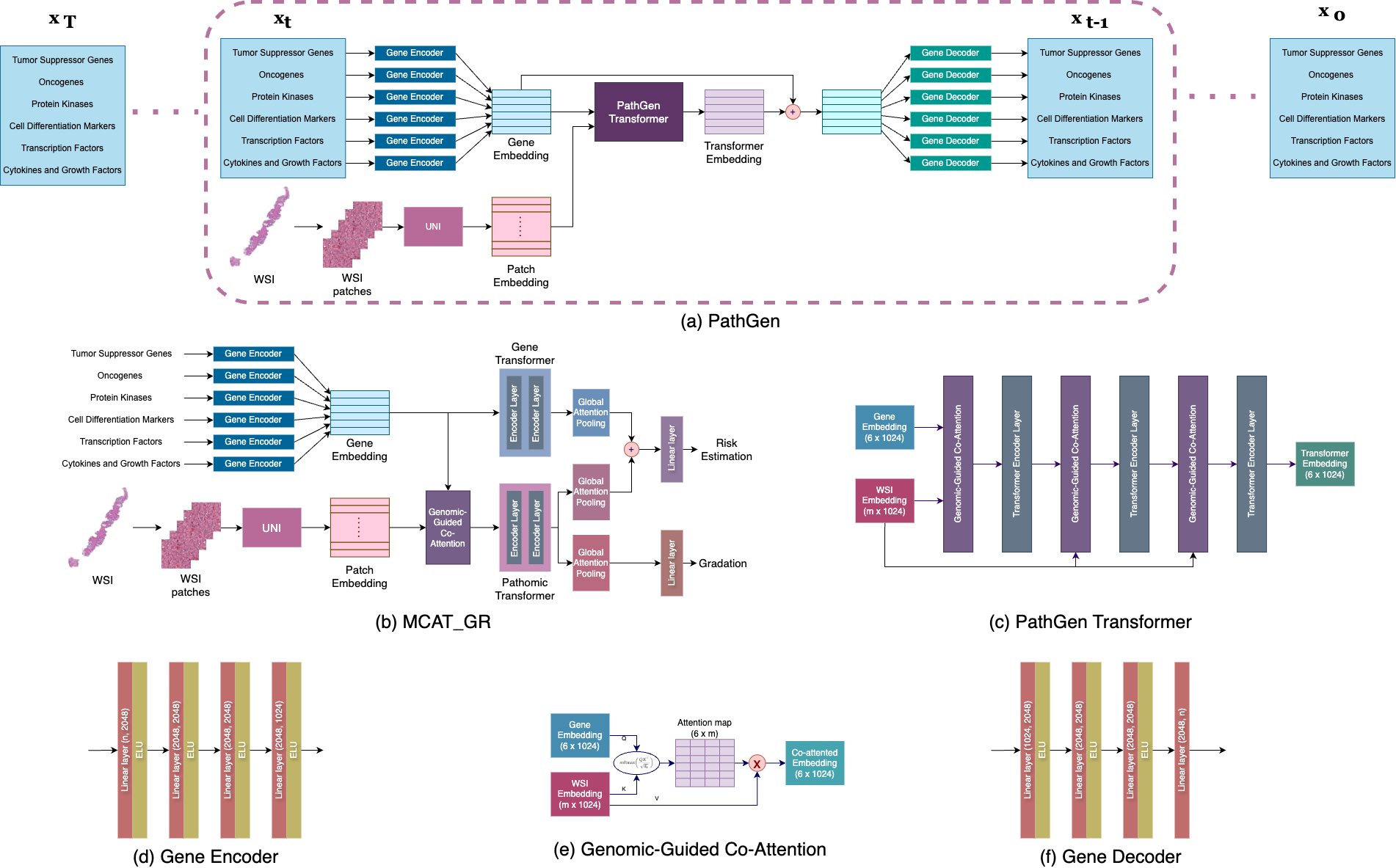}}
    \caption{\textbf{Model Architecture}. (a) Architecture of our diffusion model, PathGen, used for synthesizing gene expression levels from histopathology images. (b) Architecture of the modified MCAT model for gradation and risk estimation, MCAT\_GR. (c) Detailed architecture of PathGen transformer. (d) Architecture of gene encoder that converts gene expression level to gene embeddings. (e) Representation of genomic-guided co-attention function. (f) Architecture of gene decoder that converts gene embeddings back to gene expression levels.}
    \label{fig_model}
\end{figure}

The architecture of our model, PathGen \cite{code}, to generate transcriptomic features from whole slide images is illustrated in Figure \ref{fig_model}.a with further details in Figures \ref{fig_model}.c, \ref{fig_model}.d, \ref{fig_model}.e and \ref{fig_model}.f. To obtain the transcriptomic features at a timestep $t-1$, $\mathbf{x}_{t-1}$, the transcriptomic features at timestep $t$ and the WSI path embeddings are provided as input. The process is performed for T timesteps, where $\mathbf{x}_{T}$ is noise and $\mathbf{x}_{0}$ is the synthesized transcriptomic data. The embedding of the gene expression levels in the transcriptomic data at $\mathbf{x}_{t}$ is obtained using the gene encoder consisting of four linear layers with ELU activation as illustrated in Figure \ref{fig_model}.d. The gene embedding and the corresponding patch embedding obtained using UNI \cite{uni} goes as input to the PathGen transformer, illustrated in Figure \ref{fig_model}.c. The PathGen transformer comprises genomic-guided co-attention illustrated in Figure \ref{fig_model}.e., proposed in MCAT \cite{mcat}, and the transformer encoder layers \cite{attention}. Genomic-guided co-attention is inspired by the self-attention introduced in the transformer \cite{attention}. The gene embeddings attend to the WSI embeddings to produce co-attended embeddings, which comprise the most relevant combined features learned while training. Co-attention is performed thrice to ensure that the generated transcriptomic features have correspondence to the input WSI. Further, the transformer embeddings are decoded to gene expression levels using specific gene decoders for the different gene groups. The gene decoders comprise linear layers to ELU activation except the last, as illustrated in Figure \ref{fig_model}.f.

\subsubsection*{Training and inference}
\label{sec_traindiff}
Box \ref{algo_train} gives the training algorithm. The model is trained with the loss function in equation \ref{equ_diffloss} for $T=1000$ timesteps with a learning rate of $1\times10^{-4}$. For training, a sample is considered at a random intermediate timestep $t$ sampled uniformly. Box \ref{algo_box2} specifies the inference algorithm. The inference is performed for $T=1000$ timesteps by initially providing random noise sampled from mean mean-zero, unit-variance Gaussian distribution as input using equation \ref{equ_diffinf}. All experiments were performed with an NVIDIA A100 GPU, including training and inference.



\refstepcounter{mybox}   

\begin{tcolorbox}[
    colback=white,
    colframe=black,
    colbacktitle=blue!10!white,
    coltitle=black,
    title=\textbf{Box \themybox\ | PathGen training algorithm},
    fonttitle=\bfseries,
    boxrule=0.6pt,
    arc=2mm,
    breakable
]
\label{algo_train}

\begin{algorithmic}
    \Repeat
        \State $\mathbf{x}_0 \sim q(\mathbf{x}_0)$
        \State $t \sim \text{Uniform}(\{1, \dots, T\})$
        \State $\epsilon \sim \mathcal{N}(0, \mathbf{I})$
        \State Take gradient descent step on $\nabla_\theta \| \epsilon - \epsilon_\theta(\sqrt{\bar{\alpha}_t} \mathbf{x}_0
        + \sqrt{1 - \bar{\alpha}_t} \epsilon, t) \|^2$
    \Until{converged}
\end{algorithmic}

\end{tcolorbox}

\refstepcounter{mybox} 
\begin{tcolorbox}[
    colback=white,                
    colframe=black,               
    colbacktitle=blue!10!white,   
    coltitle=black,               
    title=\textbf{Box \themybox\ | PathGen training algorithm},
    fonttitle=\bfseries,
    boxrule=0.6pt,
    arc=2mm,
    breakable
]
\label{algo_box2}
\begin{algorithmic}
    \State $\mathbf{x}_T \sim \mathcal{N}(0, \mathbf{I})$
    \For{$t = T, \dots, 1$}
        \State $\mathbf{z} \sim \mathcal{N}(0, \mathbf{I})$ if $t > 1$, else $\mathbf{z} = 0$
        \State $\mathbf{x}_{t-1} = \frac{1}{\sqrt{\alpha_t}} \left( \mathbf{x}_t - \frac{1 - \alpha_t}{\sqrt{1 - \bar{\alpha}_t}} \epsilon_\theta(\mathbf{x}_t, t) \right) + \sigma_t \mathbf{z}$
    \EndFor
    \State \textbf{return} $\mathbf{x}_0$
\end{algorithmic}
\label{algo_inference}
\end{tcolorbox}




\subsection*{Multimodal cancer gradation and survival risk estimation}
\label{sec_gradrisk}
The PathGen synthesized transcriptomic data along with the corresponding WSIs, is used for gradation and risk estimation, using the predictive model MCAT\_GR.

\subsubsection*{Architecture}
\label{sec_mcatarch}
The architecture used for gradation and risk estimation using multi-modal histopathology and synthesized transcriptomic data is MCAT \cite{mcat} with an additional pipeline for gradation. The modified MCAT architecture, MCAT\_GR, is illustrated in Figure \ref{fig_model}.b. The co-attended features processed using the pathomic transformer of MCAT\_GR are used for gradation by passing through a separate global attention pooling. However, transcriptomic feature embeddings from the gene transformer are not used directly for gradation, as we experiment to see that not including the transcriptomic data for grade prediction performs better. The global attention pooling performed for risk estimation and gradation is separate to decrease the dependency between the grade and risk. 

\subsubsection*{Training details}
The loss used for training gradation is binary cross entropy loss \cite{bce} ($L_{grade}$), and the loss used for training risk estimation is negative log-likelihood survival loss \cite{mcat} ($L_{risk}$). Thus, the total loss for MCAT\_GR is given by the equation,
\begin{equation}
    L_{MCAT_{GR}} = \lambda \times L_{grade} + (1-\lambda) \times L_{risk}
    \label{equ_mcatloss}
\end{equation}
where $\lambda$ is a hyperparameter to control the contribution of gradation and survival loss in the training. We use $\lambda=0.3$ for our experiments. However, on unavailability of gradation annotations, models are trained for survival only using $L_{risk}$ ($\lambda=0$). The learning rate used is $2 \times 10^{-4}$.  The models were trained using an NVIDIA A100 GPU.

    




\refstepcounter{mybox} 
\begin{tcolorbox}[
    colback=white,                
    colframe=black,               
    colbacktitle=blue!10!white,   
    coltitle=black,               
    title=\textbf{Box \themybox\ - Algorithm of conformal prediction for gradation},
    fonttitle=\bfseries,
    boxrule=0.6pt,
    arc=2mm,
    breakable
]
\label{algo_box3}

\begin{algorithmic}
    \State \textbf{Input:} Calibration samples \( \{ \mathbf{x}_i, y_i \}_{i=1}^{n} \), error rate \( \alpha \) \Comment{$y_i$ is the true grade}
    
    \State \textbf{Step 1:} Compute conformal scores:
    \For{each calibration sample \( i \)}
        \State $\text{score}_i \gets 1 - p_{i_{y_i}}$ \Comment{where \( p_{i_{y_i}} \) is the predicted probability of the true class $y_i$}
    \EndFor

    \State \textbf{Step 2:} Compute adjusted quantile:
    \State $q \gets \frac{\lceil (n + 1) \cdot (1 - \alpha) \rceil}{n}$
    \State $\hat{q} \gets \text{quantile}(\{\text{score}_i\}, q)$

    \State \textbf{Step 3:} Form conformal sets:
    \For{each test sample}
        \If{$p_j \geq 1 - \hat{q}$} 
            \State Include class \( j \) in the conformal set \Comment{where $p_j$ is the predicted probability of belonging to class $j$}
        \EndIf
    \EndFor
\end{algorithmic}
\label{algo_cpgrade}
\end{tcolorbox}

\subsection*{Uncertainty estimation using conformal prediction}
\label{sec_cp}
\subsubsection*{Conformal for gradation}
\label{sec_cpgrade}
Conformal prediction is performed to estimate the reliability of model predicted grades and to provide clinicians with a conformal set of the most probable grade options. Marginal conformal prediction ensures that the coverage of prediction obtained for a population is more than $1-\alpha$, where $\alpha$ is a chosen error rate \cite{conformal}. The marginal coverage property for gradation is given by the equation,
\begin{equation}
    1 - \alpha \leq \probP{\,(Y_{test} \in \; C(X_{test})) }
    \label{equ_cov_grad}
\end{equation}
where $C$ represents the conformal set consisting of probable prediction classes to gain the desired coverage of at least $1-\alpha$. While marginal calibration ensures desired coverage for the entire population, it fails to ensure the fairness of coverage stratified over demographic groups. For instance, given a desired coverage of 90\%, female patients may have obtained 100\% coverage while male patients may have obtained 80\% coverage, still satisfying the marginal coverage property. Thus, we further ensure coverage stratified over groups for all demographic categories. The stratified conditional coverage property for gradation is given by the equation,
\begin{equation}
    1 - \alpha \leq \probP{\,(Y_{test} \in C(X_{test}) \; | \; X_{test} \in g_i}), \quad \forall\, g_i \in G
    \label{equ_condcov_grad}
\end{equation}
where G represents the demographic category (gender, age, etc.) and $g_i$ represents the groups (male, female, age $<40$, age $>60$, etc) in the category. Box \ref{algo_box3} specifies the procedure for performing conformal prediction for gradation \cite{conformal}. The uncertainty for gradation is computed using the predicted conformal set by the equation, 
\begin{equation}
    uncertainty = \frac{|C|}{N} \times \frac{\Delta_{max}}{N-1}
    \label{equ_unc_grad}
\end{equation}
where $|C|$ represents the cardinality of the conformal set, $N$ represents the number of grade classes, and  $\Delta_{max}$ stands for the maximum difference between the grades in the conformal set.

    





\refstepcounter{mybox} 
\begin{tcolorbox}[
    colback=white,                
    colframe=black,               
    colbacktitle=blue!10!white,   
    coltitle=black,               
    title=\textbf{Box \themybox\ - Algorithm of conformal prediction for survival risk estimation},
    fonttitle=\bfseries,
    boxrule=0.6pt,
    arc=2mm,
    breakable
]
\label{algo_box4}

\begin{algorithmic}
    \State \textbf{Input:} Calibration samples \( \{ \mathbf{x}_i, risk_i \}_{i=1}^{n} \), error rate \( \alpha \)
    \Comment{$risk_i$ is the risk corresponding to the ground truth time bin}
    
    \State \textbf{Step 1:} Compute conformal scores:
    \For{each calibration sample \( i \)}
        \State $\text{score}_i \gets | risk_i - \hat{risk_i} |$
        \Comment{where \( \hat{risk_i} \) is the predicted risk}
    \EndFor

    \State \textbf{Step 2:} Compute adjusted quantile:
    \State $q \gets \frac{\lceil (n + 1) \cdot (1 - \alpha) \rceil}{n}$
    \State $\hat{q} \gets \text{quantile}(\{\text{score}_i\}, q)$

    \State \textbf{Step 3:} Compute uncertainty bounds and form conformal sets:
    \For{each test sample}
        \State $risk_{lb} \gets \hat{risk} - \hat{q}$
        \State $risk_{ub} \gets \hat{risk} + \hat{q}$
        \If{$risk_{lb} \leq t_{lb}$ or $t_{ub} \leq risk_{ub}$}
            \State Include time bin \( t \) in the conformal set
            \Comment{$t_{lb}$ and $t_{ub}$ are the lower and upper bound risks for time bin $t$}
        \EndIf
    \EndFor
\end{algorithmic}
\label{algo_cprisk}
\end{tcolorbox}

\subsubsection*{Conformal for survival risk estimation}
\label{sec_cprisk}
For survival risk estimation, conformal prediction is performed to estimate the upper and lower bounds for the estimated risk. The marginal coverage property for risk estimation ensures that the ground truth survival is covered within the predicted bounds by a probability of more than $1-\alpha$, where $\alpha$ is a chosen error rate, and is given by the equation,
\begin{equation}
    1 - \alpha \leq \probP{\,(  Y_{test} \in \; [risk_{lb}, risk_{ub}])}
    \label{equ_cov_risk}
\end{equation}
where $risk_{lb}$ and $risk_{ub}$ are the conformalized lower and upper bounds of the risk predicted from $X_{test}$. For risk estimation, the survival time in months is divided into 4 time bins for a given population, considering the vital status. The time bins covered between the lower and upper bounds of the risk give the conformal set $C$. Similar to gradation, stratified conditional conformal prediction is performed to ensure fairness in coverage over different demographic groups. The stratified conditional coverage property for survival risk estimation is given by the equation,
\begin{equation}
    1 - \alpha \leq \probP{\,(  Y_{test} \in [risk_{lb}, risk_{ub}] \; | \; X_{test} \in g_i}), \quad \forall\, g_i \in G
    \label{equ_condcov_risk}
\end{equation}
where G represents the demographic category and $g_i$ represents the groups in the category. Box \ref{algo_box4} mentions the detailed conformal procedure for survival risk estimation. The uncertainty for risk estimation is further computed as,
\begin{equation}
    uncertainty = \frac{|C|}{N} \times \frac{|risk_{ub} - risk_{lb}|}{|risk_{high} - risk_{low}|}
    \label{equ_unc_risk}
\end{equation}
where N=4, the number of time bins, $risk_{high}=$ is the highest possible risk, and $risk_{low}=$ is the lowest possible risk score.\\

Both the uncertainty measures given in equation \ref{equ_unc_grad} and equation \ref{equ_unc_risk} are metrics to quantify uncertainty for gradation and survival risk estimation, respectively. 

\subsubsection*{Estimation of coverage slack}
However, though conformal properties (equations \ref{equ_cov_grad}, \ref{equ_condcov_grad}, \ref{equ_cov_risk} and \ref{equ_condcov_risk}) ensure that the obtained coverage is at least $1-\alpha$, the coverage of conformal prediction conditionally on the calibration set is a random quantity \cite{conformal}. The deviation in coverage is a function of the cardinality of the calibration set (n), and the error rate ($\alpha$),  and follows the distribution: $\mathrm{Beta}(n+1-l, \; l\;)$, where $l= \lfloor (n+1)\alpha \rfloor$ \cite{beta_conformal}. The expected deviation ($\varepsilon$) can be computed using the equation,
\begin{equation}
    \varepsilon(n,\alpha) = \max( (1-\alpha)-L,\; U-(1-\alpha),\; 0)
    \label{equ_n_eps}
\end{equation}
where L and U are the lower and upper bounds for the 90\% confidence-type interval around $1-\alpha$, for $\alpha=0.1$, and is given by,
\begin{equation}
    [L, U] = [F^{-1}_{\mathrm{Beta}(a,b)}(0.05), \; F^{-1}_{\mathrm{Beta}(a,b)}(0.95)]
    \label{equ_interval}
\end{equation}
Hence, conformalized coverage slack of $\alpha \; \pm \; \varepsilon$ is inherent for finite set calibration. The expected deviations in coverage have been computed for different calibration set sizes, using equation \ref{equ_n_eps}, and presented in Supplementary Table \ref{s-table_deviation}, for $\alpha=0.1$.

\section*{INCLUSION AND ETHICS}
The author team includes researchers from the UK and India, and contributions reflect international collaboration. Although the data sets analysed were publicly available, we acknowledge that patient cohorts may not fully represent global populations.

\section*{DATA AVAILABILITY}
For all our experiments, we use data from two anonymised publicly available open-access datasets - The Cancer Genomic Atlas (TCGA) and the Clinical Proteomic Tumor Analysis Consortium (CPTAC). The TCGA data can be accessed at {\scriptsize \url{https://www.cancer.gov/tcga}}, and the CPTAC data can be accessed at  {\scriptsize \url{https://gdc.cancer.gov/about-gdc/contributed-genomic-data-cancer-research/clinical-proteomic-tumor-analysis-consortium-cptac}}. We use the corresponding transcriptomic data downloaded from cBioPortal ({\scriptsize \url{https://www.cbioportal.org}}) for all cohorts and datasets. The gene categories used are found at {\scriptsize \url{https://github.com/mahmoodlab/MCAT/blob/master/datasets_csv_sig/signatures.csv}}. No primary data was collected as part of this study. Source data are provided with this paper. 

\section*{CODE AVAILABILITY}
For the reproducibility of research and ethos of open science, we have made the code fully available online through our GitHub repository at \url{https://github.com/Samiran-Dey/PathGen} \cite{code} under the CC-BY-NC-ND 4.0 license.

\renewcommand{\refname}{REFERENCES}   
\renewcommand{\bibname}{REFERENCES}   

\section*{ACKNOWLEDGEMENTS}
S Dey was supported by the IIT KGP AI4ICPS Chanakya Fellowship. P Basuchowdhuri is supported by the Indo-Swedish DBT-Vinnova project, BT/PR41025/Swdn/135/9/2020. CRSB was supported by the CRUK City of London Centre Award [CTRQQR-2021/100004]. T Chakraborti is supported by the Turing-Roche Strategic Partnership. We are grateful to Prof Benjamin MacArthur from the Alan Turing Institute and Mr. Chris Harbron from Roche Pharmaceuticals for their insightful comments and constructive feedback throughout this project.

\section*{AUTHOR CONTRIBUTIONS}
S. Dey has contributed to conceptualization, methodology development, experimental studies, and writing. P. Basuchowdhuri, S.K. Saha, D. Parashar contributed to overall feedback, knowledge sharing and writing/editing of the manuscript. T. Chakraborti has contributed to conceptualization, problem setting, overall research direction, project supervision and writing/editing the manuscript. C.R.S. Baernji has provided clinical expertise and feedback, and contributed to writing/editing the manuscript. 

\section*{COMPETING INTERESTS}
The authors declare no competing interests.

\section*{FIGURES}

\section*{\textbf{Generating crossmodal gene expression from cancer histopathology improves multimodal AI predictions}}
\subsection*{\centerline{SUPPLEMENTARY MATERIAL}}

\begin{table}[!ht]
\centering
\captionsetup{labelformat=default,labelsep=colon,name=Supplementary Table}
\caption{\label{data_split} Details of data splits across datasets (TCGA, CPTAC) and cancer cohorts used in the experiments. The cohorts include glioblastoma-glioma (TCGA-GBMLGG, CPTAC-GBM), renal (TCGA-KIRC), uterine (TCGA-UCEC, CPTAC-UCEC) and breast (TCGA-BRCA). Data from the CPTAC cohorts were excluded from training. Models trained on the TCGA cohorts were evaluated on the corresponding CPTAC datasets.\\}
\begin{tabular}{|l|cccc|cccc|}
\hline
\multicolumn{1}{|c|}{\multirow{2}{*}{\textbf{Datasets}}} & \multicolumn{4}{c|}{\textbf{Cases}}     & \multicolumn{4}{c|}{\textbf{Slides}}    \\ \cline{2-9} 
\multicolumn{1}{|c|}{}                                   & Training & Validation & Calibration & Testing & Training & Validation & Calibration & Testing \\ \hline
TCGA-GBMLGG                                              & 532   & 56         & 75          & 82   & 653   & 69         & 86          & 104  \\
TCGA-KIRC                                                & 337   & 34         & 45          & 46   & 344   & 42         & 48          & 51   \\
TCGA-UCEC                                                & 191   & 19         & 30          & 27   & 210   & 23         & 31          & 30   \\
TCGA-BRCA                                                & 676   & 78         & 97          & 95   & 729   & 80         & 102         & 99   \\
CPTAC-GBM                                                & -     & -          & 32          & 30   & -     & -          & 151         & 91   \\
CPTAC-UCEC                                               & -     & -          & 38          & 33   & -     & -          & 194         & 170  \\ \hline
\end{tabular}
\end{table}

\begin{table}[!ht]
\centering
\captionsetup{labelformat=default,labelsep=colon,name=Supplementary Table}
\caption{\label{data_trans} Table for number of genes in each gene group in the transcriptomic data. Same number of genes have been considered in each gene groups for same cohorts from different datasets. We note that there is repetition of genes across gene groups. The last row shows the additive result of the number of genes across gene groups including repetitions, and not the total number of genes considered for prediction.}
\resizebox{\textwidth}{!}{
\begin{tabular}{|l|cccccc|}
\hline
\multicolumn{1}{|c|}{\textbf{Gene type}} & \textbf{TCGA-GBMLGG} & \textbf{TCGA-KIRC} & \textbf{TCGA-UCEC} & \textbf{TCGA-BRCA} & \textbf{CPTAC-GBM} & \textbf{CPTAC-UCEC} \\ \hline
Tumor Suppressor Genes                 & 84                   & 145                & 152                & 227                & 84                 & 152                 \\
Oncogenes                              & 314                  & 522                & 554                & 959                & 314                & 554                 \\
Protein Kinases                        & 498                  & 669                & 700                & 934                & 498                & 700                 \\
Cell Differentiation Markers           & 424                  & 681                & 693                & 843                & 424                & 693                 \\
Transcription Factors                  & 1396                 & 2366               & 2489               & 3712               & 1396               & 2489                \\
Cytokines and Growth Factors           & 428                  & 990                & 965                & 1142               & 428                & 965                 \\ \hline
All gene groups                              & 3144                 & 5373               & 5553               & 7817               & 3144               & 5553    \\ \hline           
\end{tabular}
}
\end{table}

\begin{table}[!ht]
\centering
\captionsetup{labelformat=default,labelsep=colon,name=Supplementary Table}
\caption{\label{data_demo} Demographic details of the data used for calibration and testing for all the datasets and cohorts. Ground truth grades are absent for TCGA-BRCA and CPTAC-GBM datasets. Also, patient age is missing in the clinical data for the CPTAC cohorts. ncal denotes the number of samples in the calibration set and ntest denotes the number of samples in the test set.}
\resizebox{\textwidth}{!}{
\begin{tabular}{|l|l|cc|cc|cc|cc|cc|cc|}
\hline
\multicolumn{1}{|c|}{\multirow{2}{*}{\textbf{Category}}} & \multicolumn{1}{c|}{\multirow{2}{*}{\textbf{Group}}} & \multicolumn{2}{c|}{\textbf{TCGA - GBMLGG}} & \multicolumn{2}{c|}{\textbf{TCGA - KIRC}} & \multicolumn{2}{c|}{\textbf{TCGA - UCEC}} & \multicolumn{2}{c|}{\textbf{TCGA - BRCA}} & \multicolumn{2}{c|}{\textbf{CPTAC - GBM}} & \multicolumn{2}{c|}{\textbf{CPTAC - UCEC}} \\ \cline{3-14} 
\multicolumn{1}{|c|}{} & \multicolumn{1}{c|}{} & \multicolumn{1}{c|}{ncal} & ntest & \multicolumn{1}{c|}{ncal} & ntest & \multicolumn{1}{c|}{ncal} & ntest & \multicolumn{1}{c|}{ncal} & ntest & \multicolumn{1}{c|}{ncal} & ntest & \multicolumn{1}{c|}{ncal} & ntest \\ \hline
overall & - & \multicolumn{1}{c|}{86} & 104 & \multicolumn{1}{c|}{144} & 153 & \multicolumn{1}{c|}{93} & 90 & \multicolumn{1}{c|}{306} & 297 & \multicolumn{1}{c|}{453} & 273 & \multicolumn{1}{c|}{582} & 510 \\ \hline
\multirow{2}{*}{gender} & male & \multicolumn{1}{c|}{55} & 66 & \multicolumn{1}{c|}{102} & 99 & \multicolumn{1}{c|}{-} & - & \multicolumn{1}{c|}{282} & 120 & \multicolumn{1}{c|}{282} & 120 & \multicolumn{1}{c|}{-} & - \\ \cline{2-14} 
 & female & \multicolumn{1}{c|}{31} & 38 & \multicolumn{1}{c|}{42} & 54 & \multicolumn{1}{c|}{93} & 90 & \multicolumn{1}{c|}{171} & 153 & \multicolumn{1}{c|}{171} & 153 & \multicolumn{1}{c|}{582} & 510 \\ \hline
\multirow{3}{*}{age} & \textless 40 & \multicolumn{1}{c|}{33} & 43 & \multicolumn{1}{c|}{6} & 3 & \multicolumn{1}{c|}{-} & - & \multicolumn{1}{c|}{30} & 9 & \multicolumn{1}{c|}{-} & - & \multicolumn{1}{c|}{-} & - \\ \cline{2-14} 
 & 40 to 60 & \multicolumn{1}{c|}{28} & 37 & \multicolumn{1}{c|}{51} & 84 & \multicolumn{1}{c|}{42} & 39 & \multicolumn{1}{c|}{162} & 144 & \multicolumn{1}{c|}{-} & - & \multicolumn{1}{c|}{-} & - \\ \cline{2-14} 
 & \textgreater 60 & \multicolumn{1}{c|}{25} & 24 & \multicolumn{1}{c|}{87} & 66 & \multicolumn{1}{c|}{51} & 51 & \multicolumn{1}{c|}{114} & 144 & \multicolumn{1}{c|}{-} & - & \multicolumn{1}{c|}{-} & - \\ \hline
\multirow{3}{*}{magnification} & level 1 & \multicolumn{1}{c|}{27} & 29 & \multicolumn{1}{c|}{48} & 51 & \multicolumn{1}{c|}{31} & 30 & \multicolumn{1}{c|}{102} & 99 & \multicolumn{1}{c|}{151} & 91 & \multicolumn{1}{c|}{194} & 170 \\ \cline{2-14} 
 & level 2 & \multicolumn{1}{c|}{27} & 38 & \multicolumn{1}{c|}{48} & 51 & \multicolumn{1}{c|}{31} & 30 & \multicolumn{1}{c|}{102} & 99 & \multicolumn{1}{c|}{151} & 91 & \multicolumn{1}{c|}{194} & 170 \\ \cline{2-14} 
 & level 3 & \multicolumn{1}{c|}{32} & 37 & \multicolumn{1}{c|}{48} & 51 & \multicolumn{1}{c|}{31} & 30 & \multicolumn{1}{c|}{102} & 99 & \multicolumn{1}{c|}{151} & 91 & \multicolumn{1}{c|}{194} & 170 \\ \hline
\multirow{2}{*}{censorship} & alive & \multicolumn{1}{c|}{48} & 67 & \multicolumn{1}{c|}{75} & 102 & \multicolumn{1}{c|}{81} & 78 & \multicolumn{1}{c|}{279} & 279 & \multicolumn{1}{c|}{165} & 69 & \multicolumn{1}{c|}{555} & 447 \\ \cline{2-14} 
 & dead & \multicolumn{1}{c|}{38} & 37 & \multicolumn{1}{c|}{69} & 51 & \multicolumn{1}{c|}{12} & 12 & \multicolumn{1}{c|}{27} & 18 & \multicolumn{1}{c|}{288} & 204 & \multicolumn{1}{c|}{27} & 63 \\ \hline
\multirow{4}{*}{grade} & grade I & \multicolumn{1}{c|}{-} & - & \multicolumn{1}{c|}{3} & 3 & \multicolumn{1}{c|}{30} & 24 & \multicolumn{1}{c|}{-} & - & \multicolumn{1}{c|}{-} & - & \multicolumn{1}{c|}{225} & 99 \\ \cline{2-14} 
 & grade II & \multicolumn{1}{c|}{34} & 39 & \multicolumn{1}{c|}{57} & 63 & \multicolumn{1}{c|}{21} & 24 & \multicolumn{1}{c|}{-} & - & \multicolumn{1}{c|}{-} & - & \multicolumn{1}{c|}{207} & 270 \\ \cline{2-14} 
 & grade III & \multicolumn{1}{c|}{28} & 36 & \multicolumn{1}{c|}{60} & 63 & \multicolumn{1}{c|}{42} & 42 & \multicolumn{1}{c|}{-} & - & \multicolumn{1}{c|}{-} & - & \multicolumn{1}{c|}{150} & 141 \\ \cline{2-14} 
 & grade IV & \multicolumn{1}{c|}{24} & 29 & \multicolumn{1}{c|}{24} & 24 & \multicolumn{1}{c|}{-} & - & \multicolumn{1}{c|}{-} & - & \multicolumn{1}{c|}{453} & 273 & \multicolumn{1}{c|}{-} & - \\ \hline
\multirow{4}{*}{time bin} & bin 1 & \multicolumn{1}{c|}{80} & 82 & \multicolumn{1}{c|}{72} & 72 & \multicolumn{1}{c|}{66} & 63 & \multicolumn{1}{c|}{255} & 249 & \multicolumn{1}{c|}{453} & 273 & \multicolumn{1}{c|}{474} & 423 \\ \cline{2-14} 
 & bin 2 & \multicolumn{1}{c|}{4} & 13 & \multicolumn{1}{c|}{39} & 60 & \multicolumn{1}{c|}{27} & 27 & \multicolumn{1}{c|}{45} & 45 & \multicolumn{1}{c|}{-} & - & \multicolumn{1}{c|}{108} & 87 \\ \cline{2-14} 
 & bin 3 & \multicolumn{1}{c|}{2} & 9 & \multicolumn{1}{c|}{21} & 18 & \multicolumn{1}{c|}{-} & - & \multicolumn{1}{c|}{3} & 3 & \multicolumn{1}{c|}{-} & - & \multicolumn{1}{c|}{-} & - \\ \cline{2-14} 
 & bin 4 & \multicolumn{1}{c|}{-} & - & \multicolumn{1}{c|}{12} & 3 & \multicolumn{1}{c|}{-} & - & \multicolumn{1}{c|}{-} & - & \multicolumn{1}{c|}{-} & - & \multicolumn{1}{c|}{-} & - \\ \hline
\end{tabular}
}
\end{table}

\begin{table}[!ht]
\centering
\captionsetup{labelformat=default,labelsep=colon,name=Supplementary Table}
\caption{\label{result_gen_use} Table of results for evaluation of the use of synthesized transcriptomic data for grade and survival prediction in four different settings- 1) We do not use transcriptomic data at all and attend to the WSI patches randomly; 2) We use the gene expression levels only to co-attend to the WSI patch embeddings; 3) We use the transcriptomic embeddings collectively with co-attended WSI features for survival risk estimation only; 4) We use transcriptomic features for co-attending to WSI patches, and then use the transcriptomic embeddings along with the co-attended features for performing both survival risk estimation and gradation. Setting 3 gives the best mean AUC and C Index for gradation and survival risk estimation across both data cohorts. Further, ANOSIM test for combined risk estimation and gradation between the two cases suggested that the difference in performance between all the cases is significant (p-value $< 0.05$). Thus, the architecture of our model is aligned to setting 3. \textuparrow \ indicates the larger the better. Best mean scores are marked in \textbf{bold}. \\}

\begin{tabular}{|c||c|c||c|c||c|c||c|c|}
\hline
~  & \multicolumn{2}{ c ||}{Setting 1} & \multicolumn{2}{ c ||}{Setting 2} & \multicolumn{2}{ c ||}{Setting 3} & \multicolumn{2}{ c |}{Setting 4} \\ 
~  & \multicolumn{2}{ c ||}{random (not used)} & \multicolumn{2}{ c ||}{only co-attention} & \multicolumn{2}{ c ||}{co-attention, risk} & \multicolumn{2}{ c |}{co-attention, risk, grade} \\ \hline
                & Risk      & Grade & Risk         & Grade & Risk               & Grade & Risk                      & Grade \\
 $\lambda$ & C Index \textuparrow & AUC \textuparrow   & C Index \textuparrow      & AUC \textuparrow   & C Index \textuparrow            & AUC \textuparrow   & C Index \textuparrow                   & AUC \textuparrow   \\ \hline \hline
\multicolumn{9}{| c |}{TCGA - GBMLGG}  \\ \hline
0.1 & 0.835 & 0.829 & 0.843 & 0.878 & 0.863 & 0.884 & 0.848 & 0.868 \\
0.2 & 0.832 & 0.863 & 0.872 & 0.869 & 0.852 & 0.895 & 0.864 & 0.809 \\
0.3 & 0.842 & 0.823 & 0.849 & 0.886 & 0.861 & 0.890 & 0.859 & 0.849 \\
0.4 & 0.844 & 0.828 & 0.857 & 0.885 & 0.864 & 0.886 & 0.858 & 0.829 \\
0.5 & 0.856 & 0.847 & 0.861 & 0.890 & 0.846 & 0.864 & 0.830 & 0.844  \\
0.6 & 0.847 & 0.873 & 0.865 & 0.877 & 0.847 & 0.883 & 0.820 & 0.733 \\
0.7 & 0.861 & 0.836 & 0.865 & 0.867 & 0.855 & 0.888 & 0.879 & 0.843 \\
0.8 & 0.862 & 0.745 & 0.857 & 0.856 & 0.836 & 0.811 & 0.808 & 0.731 \\
0.9 & 0.850 & 0.810 & 0.852 & 0.801 & 0.845 & 0.809 & 0.801 & 0.724 \\ \hline
mean & 0.843 & 0.820 & 0.848 & 0.839 & \textbf{0.854} & \textbf{0.847} & 0.825 & 0.796 \\ \hline
\hline
\multicolumn{9}{| c |}{TCGA - KIRC} \\ \hline 
0.1 & 0.626 & 0.729 & 0.666 & 0.713 & 0.703 & 0.773 & 0.576 & 0.741 \\
0.2 & 0.709 & 0.713 & 0.658 & 0.768 & 0.650 & 0.778 & 0.516 & 0.735 \\
0.3 & 0.671 & 0.714 & 0.688 & 0.763 & 0.681 & 0.773 & 0.548 & 0.746 \\
0.4 & 0.701 & 0.712 & 0.701 & 0.770 & 0.675 & 0.772 & 0.513 & 0.740  \\
0.5 & 0.662 & 0.770 & 0.605 & 0.766 & 0.658 & 0.769 & 0.509 & 0.747  \\
0.6 & 0.672 & 0.763 & 0.654 & 0.767 & 0.667 & 0.763 & 0.516 & 0.743  \\
0.7 & 0.677 & 0.739 & 0.673 & 0.768 & 0.643 & 0.760 & 0.512 & 0.750 \\
0.8 & 0.708 & 0.763 & 0.650 & 0.774 & 0.613 & 0.765 & 0.551 & 0.753 \\
0.9 & 0.607 & 0.754 & 0.592 & 0.760 & 0.575 & 0.757 & 0.573 & 0.742 \\ \hline
mean & 0.617 & 0.742 & 0.629 & 0.737 & \textbf{0.639} & \textbf{0.765} & 0.575 & 0.742 \\ \hline
\end{tabular}
\end{table}

\begin{figure}[!ht]
    \centerline{\includegraphics[scale=0.217]{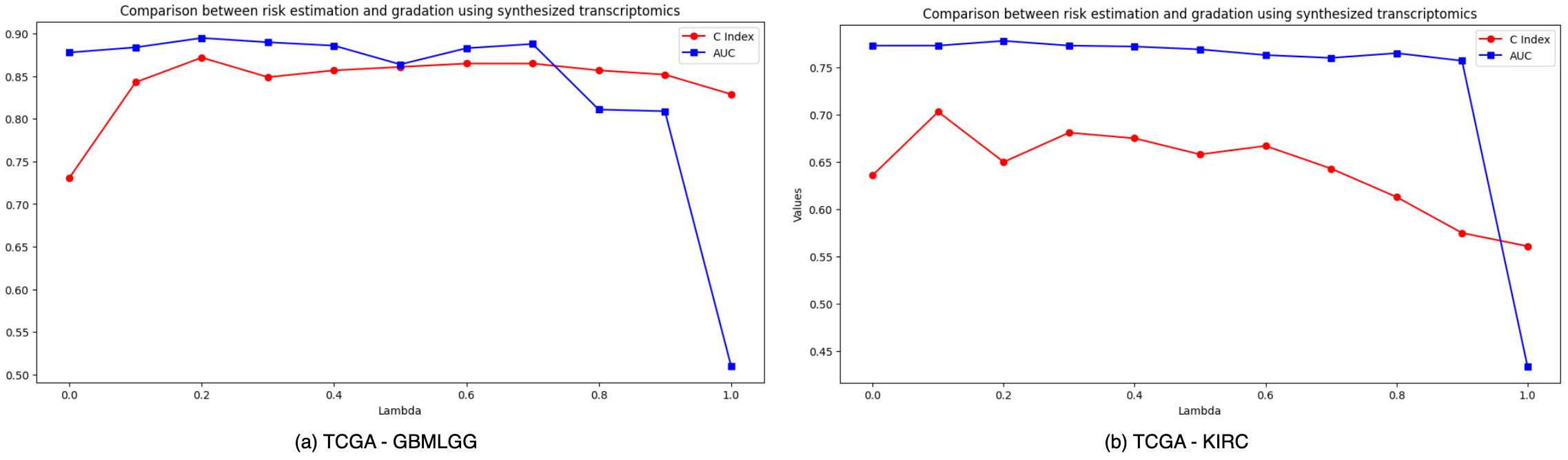}}
    \captionsetup{labelformat=default,labelsep=colon,name=Supplementary Figure}
    \caption{ \label{fig_lambda} \textbf{Ablation for $\lambda$.} Plot for AUC and C Index across $\lambda$ for (a) TCGA-GBMLGG, (b) TCGA-KIRC. $\lambda=0$ and $\lambda=1$ correspond to the extreme cases of equation $L = \lambda \times L_{survival} + (1-\lambda) \times L_{grade}$ where the model is trained only using gradation loss and only using survival loss respectively. Using only gradation loss leads to poor performance of risk estimation when $\lambda=0$, and using only risk estimation loss leads to poor gradation performance. Learning both grade and survival risk together helps the model correlate between the two and learn better. This is expected as one of the losses essentially becomes unused, so we decide to ignore the extreme cases of $\lambda=0$ and $\lambda=1$. Also, we observe that for $\lambda=0.3$ the model reaches an optimal performance where both AUC and C Index scores are close and high together for both cohorts.}
\end{figure}

\begin{table}[!ht]
\centering
\captionsetup{labelformat=default,labelsep=colon,name=Supplementary Table}
\caption{\label{result_trans}Table of results for comparison of synthesized and real transcriptomic features. Normalized MAE results are denoted with nMAE. The correlation coefficient is calculated using Spearman correlation, and all corresponding correlation p-values are found to be $ < 0.05 $, denoting significant correlation. \textuparrow \ indicates the larger the better, and \textdownarrow \ indicates the smaller the better. A higher correlation with a high corresponding MAE would denote that the range of gene expression levels is large for the particular group.}
\begin{tabular}{|l|cccccc|}
\hline
\multicolumn{1}{|c|}{\multirow{2}{*}{\textbf{Gene type}}} & \multicolumn{2}{c|}{\textbf{TCGA-GBMLGG}} & \multicolumn{2}{c|}{\textbf{TCGA-KIRC}} & \multicolumn{2}{c|}{\textbf{TCGA-UCEC}} \\ \cline{2-7} 
\multicolumn{1}{|c|}{} & nMAE & \multicolumn{1}{c|}{Correlation} & nMAE & \multicolumn{1}{c|}{Correlation} & nMAE & Correlation \\ \hline
Tumor Suppressor Genes & 0.184 & \multicolumn{1}{c|}{0.472} & 0.166 & \multicolumn{1}{c|}{0.629} & 0.177 & 0.372 \\
Oncogenes & 0.160 & \multicolumn{1}{c|}{0.770} & 0.136 & \multicolumn{1}{c|}{0.817} & 0.182 & 0.397 \\
Protein Kinases & 0.148 & \multicolumn{1}{c|}{0.749} & 0.183 & \multicolumn{1}{c|}{0.785} & 0.183 & 0.306 \\
Cell Differentiation Markers & 0.137 & \multicolumn{1}{c|}{0.724} & 0.130 & \multicolumn{1}{c|}{0.777} & 0.194 & 0.339 \\
Transcription Factors & 0.143 & \multicolumn{1}{c|}{0.615} & 0.197 & \multicolumn{1}{c|}{0.666} & 0.139 & 0.623 \\
Cytokines and Growth Factors & 0.107 & \multicolumn{1}{c|}{0.771} & 0.091 & \multicolumn{1}{c|}{0.610} & 0.236 & 0.270 \\ \hline
All gene groups & 0.141 & \multicolumn{1}{c|}{0.713} & 0.160 & \multicolumn{1}{c|}{0.717} & 0.173 & 0.436 \\ \hline \hline
\multicolumn{1}{|c|}{\multirow{2}{*}{\textbf{Gene type}}} & \multicolumn{2}{c|}{\textbf{TCGA-BRCA}} & \multicolumn{2}{c|}{\textbf{CPTAC-GBM}} & \multicolumn{2}{c|}{\textbf{CPTAC-UCEC}} \\ \cline{2-7} 
\multicolumn{1}{|c|}{} & nMAE & \multicolumn{1}{c|}{Correlation} & nMAE & \multicolumn{1}{c|}{Correlation} & nMAE & Correlation \\ \hline
Tumor Suppressor Genes & 0.194 & \multicolumn{1}{c|}{0.568} & 0.259 & \multicolumn{1}{c|}{0.445} & 0.228 & 0.540 \\
Oncogenes & 0.136 & \multicolumn{1}{c|}{0.705} & 0.043 & \multicolumn{1}{c|}{0.874} & 0.197 & 0.582 \\
Protein Kinases & 0.131 & \multicolumn{1}{c|}{0.766} & 0.036 & \multicolumn{1}{c|}{0.923} & 0.209 & 0.516 \\
Cell Differentiation Markers & 0.166 & \multicolumn{1}{c|}{0.617} & 0.036 & \multicolumn{1}{c|}{0.941} & 0.206 & 0.541 \\
Transcription Factors & 0.167 & \multicolumn{1}{c|}{0.629} & 0.316 & \multicolumn{1}{c|}{0.490} & 0.146 & 0.827 \\
Cytokines and Growth Factors & 0.136 & \multicolumn{1}{c|}{0.628} & 0.035 & \multicolumn{1}{c|}{0.965} & 0.197 & 0.493 \\ \hline
All gene groups & 0.155 & \multicolumn{1}{c|}{0.642} & 0.167 & \multicolumn{1}{c|}{0.662} & 0.178 & 0.669 \\ \hline
\end{tabular}
\end{table}

\begin{table}[!ht]
\centering
\captionsetup{labelformat=default,labelsep=colon,name=Supplementary Table}
\caption{\label{result_main}Table of results for comparing predictive performance on using only WSIs (without transcriptomic data), using synthesized transcriptomic data and WSIs, and using real transcriptomic data and WSIs. The separate p-values for gradation and risk estimation are computed by performing two-sided Wilcoxon rank-sum test, whereas the p-values over gradation and risk estimation together are computed using two-sided ANOSIM test. ANOSIM test $p$-values $= 0.01$, for comparison between without vs. synthesized, for TCGA-GBMLGG, TCGA-KIRC, TCGA-UCEC, CPTAC-GBM, and CPTAC-UCEC. However, for real vs. synthesized, ANOSIM test $p$-values evaluate to 0.929 for TCGA-GBMLGG, 0.327 for TCGA-KIRC, 1.000 for TCGA-UCEC, 0.993 for CPTAC-GBM and 1.000 for CPTAC-UCEC. \textuparrow indicates the larger the better, and \textdownarrow indicates the smaller the better.}
\begin{tabular}{|l|ccccc|}
\hline
\multicolumn{1}{|c|}{} & \multicolumn{5}{c|}{Gradation} \\ \cline{2-6} 
\multicolumn{1}{|c|}{} & \multicolumn{3}{c|}{AUC\textuparrow} & \multicolumn{2}{c|}{Wilcoxon test p-value} \\ \cline{2-6} 
\multicolumn{1}{|c|}{\multirow{-3}{*}{Datasets}} & \multicolumn{1}{c|}{without} & \multicolumn{1}{c|}{synthesized} & \multicolumn{1}{c|}{real} & \multicolumn{1}{c|}{without vs. synthesized \textdownarrow} & real vs. synthesized \textuparrow \\ \hline
TCGA-GBMLGG & \multicolumn{1}{c|}{0.823} & \multicolumn{1}{c|}{0.890} & \multicolumn{1}{c|}{0.907} & \multicolumn{1}{c|}{4.23E-02} & 9.95E-01 \\ \hline
TCGA-KIRC & \multicolumn{1}{c|}{{\color[HTML]{212121} 0.714}} & \multicolumn{1}{c|}{\textit{0.773}} & \multicolumn{1}{c|}{0.778} & \multicolumn{1}{c|}{2.21E-04} & 8.50E-01 \\ \hline
TCGA-UCEC & \multicolumn{1}{c|}{0.796} & \multicolumn{1}{c|}{\textit{0.821}} & \multicolumn{1}{c|}{0.828} & \multicolumn{1}{c|}{1.28E-02} & 9.41E-01 \\ \hline
CPTAC-GBM & \multicolumn{1}{c|}{0.736} & \multicolumn{1}{c|}{\textit{0.865}} & \multicolumn{1}{c|}{0.863} & \multicolumn{1}{c|}{8.40E-03} & 9.85E-01 \\ \hline
CPTAC-UCEC & \multicolumn{1}{c|}{0.508} & \multicolumn{1}{c|}{\textit{0.593}} & \multicolumn{1}{c|}{0.593} & \multicolumn{1}{c|}{7.76E-42} & 1.00E+00 \\ \hline
\multicolumn{1}{|c|}{} & \multicolumn{5}{c|}{Risk Estimation} \\ \cline{2-6} 
\multicolumn{1}{|c|}{} & \multicolumn{3}{c|}{C Index \textuparrow} & \multicolumn{2}{c|}{Wilcoxon test p-value} \\ \cline{2-6} 
\multicolumn{1}{|c|}{\multirow{-3}{*}{Datasets}} & \multicolumn{1}{c|}{without} & \multicolumn{1}{c|}{synthesized} & \multicolumn{1}{c|}{real} & \multicolumn{1}{c|}{without vs. synthesized \textdownarrow} & real vs. synthesized \textuparrow \\ \hline
TCGA-GBMLGG & \multicolumn{1}{c|}{0.842} & \multicolumn{1}{c|}{\textit{0.861}} & \multicolumn{1}{c|}{0.866} & \multicolumn{1}{c|}{8.20E-07} & 6.04E-01 \\ \hline
TCGA-KIRC & \multicolumn{1}{c|}{{\color[HTML]{212121} 0.671}} & \multicolumn{1}{c|}{0.681} & \multicolumn{1}{c|}{0.697} & \multicolumn{1}{c|}{1.23E-11} & 3.00E-01 \\ \hline
TCGA-UCEC & \multicolumn{1}{c|}{0.663} & \multicolumn{1}{c|}{\textit{0.673}} & \multicolumn{1}{c|}{0.680} & \multicolumn{1}{c|}{4.81E-31} & 8.87E-01 \\ \hline
TCGA-BRCA & \multicolumn{1}{c|}{0.603} & \multicolumn{1}{c|}{\textit{0.720}} & \multicolumn{1}{c|}{0.720} & \multicolumn{1}{c|}{2.35E-03} & 9.30E-01 \\ \hline
CPTAC-GBM & \multicolumn{1}{c|}{0.547} & \multicolumn{1}{c|}{\textit{0.565}} & \multicolumn{1}{c|}{0.564} & \multicolumn{1}{c|}{3.90E-33} & 8.15E-01 \\ \hline
CPTAC-UCEC & \multicolumn{1}{c|}{0.518} & \multicolumn{1}{c|}{\textit{0.530}} & \multicolumn{1}{c|}{0.533} & \multicolumn{1}{c|}{2.26E-164} & 9.32E-01 \\ \hline
\end{tabular}

\end{table}

\begin{table}[!ht]
\centering
\captionsetup{labelformat=default,labelsep=colon,name=Supplementary Table}
\caption{\label{result_gbm_lgg}Table for results for gradation and survival risk estimation using real and synthesized transcriptomic data, separately for the TCGA-GBMLGG cohorts. The $p$-values are computed using two-sided Wilcoxon rank-sum test.\\}
\begin{tabular}{|l|ccc|ccc|}
\hline
\multicolumn{1}{|c|}{TCGA-GBMLGG} & \multicolumn{3}{c|}{Gradation (AUC\textuparrow)} & \multicolumn{3}{c|}{Risk Estimation (C Index \textuparrow)} \\ \cline{2-7} 
\multicolumn{1}{|c|}{cohorts} & \multicolumn{1}{c|}{real} & \multicolumn{1}{c|}{synthesized} & Wilcoxon p-value & \multicolumn{1}{c|}{real} & \multicolumn{1}{c|}{synthesized} & Wilcoxon p-value \\ \hline
TCGA-LGG & \multicolumn{1}{c|}{0.835} & \multicolumn{1}{c|}{0.81} & 0.775 & \multicolumn{1}{c|}{0.707} & \multicolumn{1}{c|}{0.675} & 0.472 \\ \hline
TCGA-GBM & \multicolumn{1}{c|}{0.995} & \multicolumn{1}{c|}{0.993} & 0.822 & \multicolumn{1}{c|}{0.509} & \multicolumn{1}{c|}{0.539} & 0.489 \\ \hline
\end{tabular}
\end{table}

\begin{table}[!ht]
\captionsetup{labelformat=default,labelsep=colon,name=Supplementary Table}
\caption{\label{result_coattn_perc}Table for ablation study on percentage contribution of the different gene groups in co-attention with WSI features for different cohorts. \\}
\resizebox{\textwidth}{!}{
\begin{tabular}{|l|cccccc|}
\hline
\multicolumn{1}{|c|}{\multirow{2}{*}{\textbf{Gene type}}} & \multicolumn{6}{c|}{\textbf{Percentage Contribution (in \%)}}                           \\ \cline{2-7} 
\multicolumn{1}{|c|}{}                                    & TCGA-GBMLGG & TCGA-KIRC & TCGA-UCEC & TCGA-BRCA & CPTAC-GBM & CPTAC-UCEC \\ \hline \hline
Tumor Suppressor Genes                                    & 12.51       & 6.05      & 5.25      & 11.56     & 12.15     & 4.74       \\
Oncogenes                                                 & 18.02       & 15.59     & 4.33      & 12.90     & 13.16     & 6.35       \\
Protein Kinases                                           & 10.40       & 11.86     & 38.41     & 21.12     & 15.30     & 29.09      \\
Cell Differentiation Markers                              & 10.00       & 8.10      & 10.82     & 18.14     & 23.55     & 18.00      \\
Transcription Factors                                     & 42.21       & 48.80     & 17.74     & 23.49     & 29.06     & 25.89      \\
Cytokines and Growth Factors                              & 6.85        & 9.61      & 23.44     & 12.80     & 6.77      & 15.93      \\ \hline
\end{tabular}
}
\end{table}

\begin{figure} [!ht]
    \centerline{\includegraphics[scale=0.21]{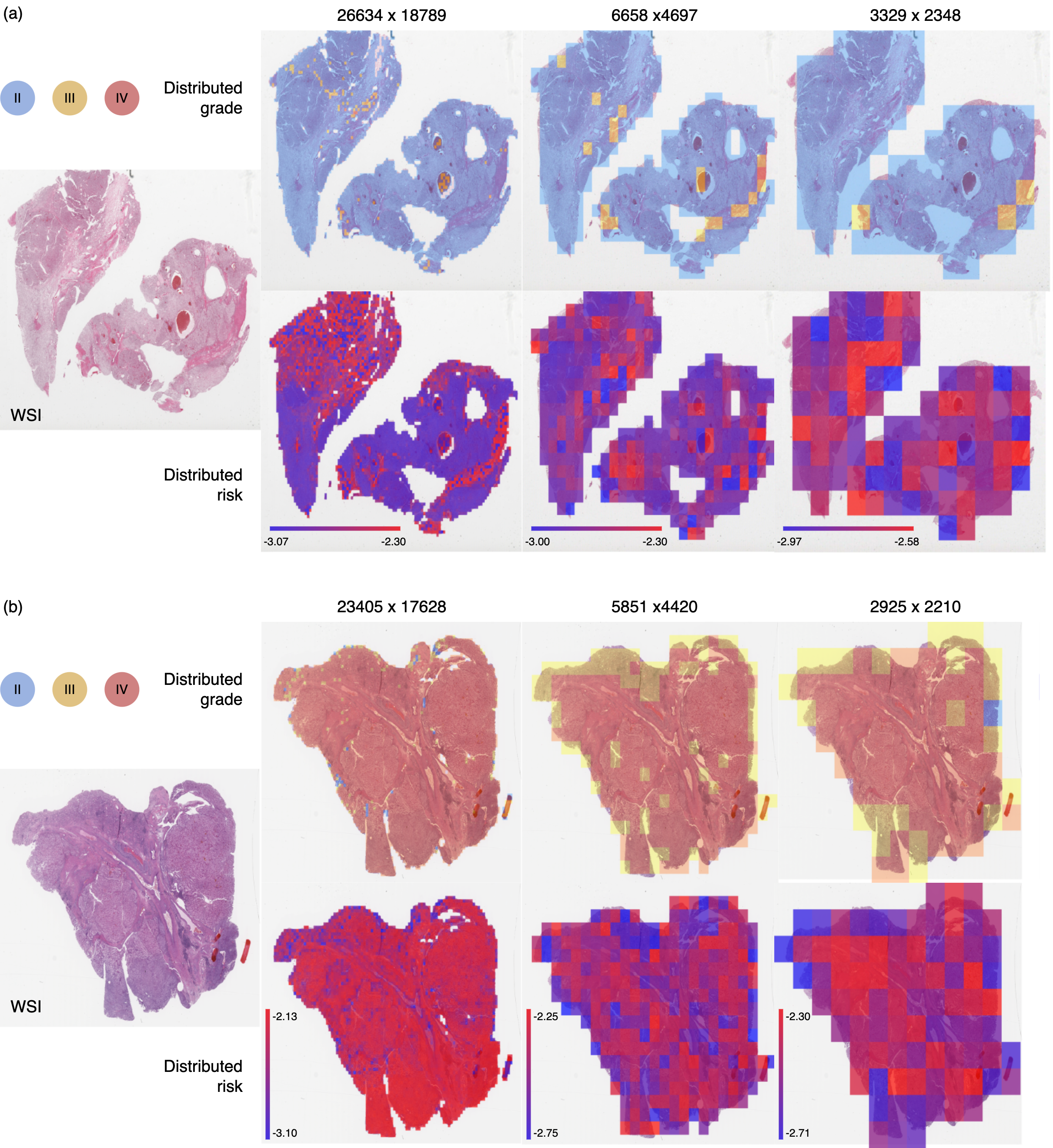}}
    \captionsetup{labelformat=default,labelsep=colon,name=Supplementary Figure}
    \caption{ \label{fig_mag} \textbf{Predicted intra-tumour heterogeneity at different magnification levels} (a) A whole slide image (WSI) of ground truth grade II, survival time 62.62 months and alive survival status belonging to a male patient of 47 years. (b) A WSI of ground truth grade IV, survival time 54.61 months and dead survival status belonging to a 55 years old female patient. The WSIs belong to TCGA-KIRC. It is observed that for higher resolution images, the distributed predictions are more accurate.}
\end{figure}

\begin{figure} [!t]
    \centerline{\includegraphics[scale=0.155]{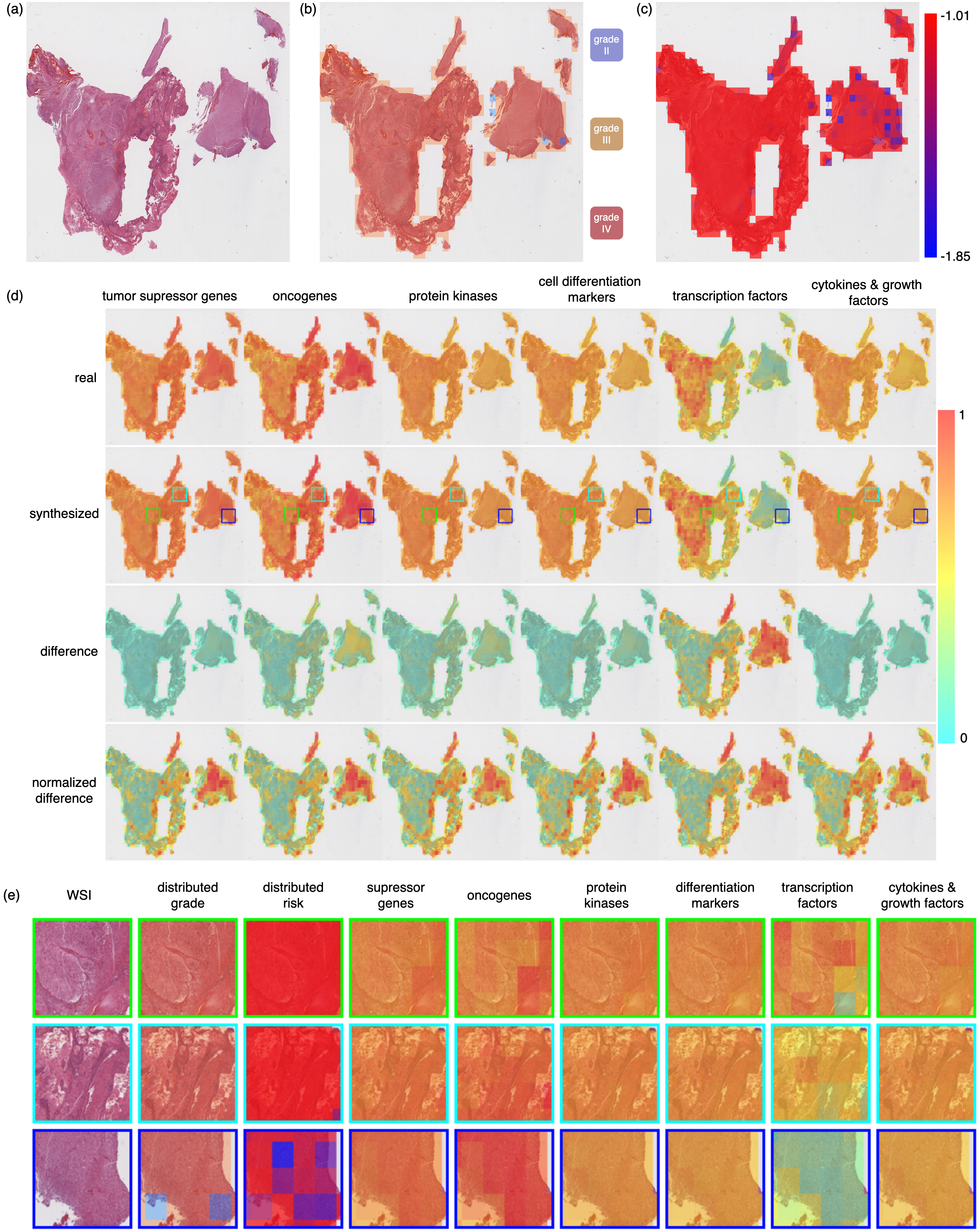}}
    \captionsetup{labelformat=default,labelsep=colon,name=Supplementary Figure}
    \caption{\textbf{Explainability maps for a male patient of age 62, ground truth grade IV, survival time 10.28 months, and survival status deceased from the TCGA-GBM cohort}. (a) whole slide image (WSI) (b) intra-tumour gradation heterogeneity (c) intra-tumour survival heterogeneity (d) co-attention maps for real and synthesized transcriptomic data and WSI patches for different gene groups (e) comparative study of intra-tumour heterogeneity and corresponding co-attention maps obtained for prediction using synthesized transcriptomes for chosen regions marked with different coloured rectangles. }
    \label{fig_gbmlgg1}
\end{figure}

\begin{figure} [!t]
    \centerline{\includegraphics[scale=0.13]{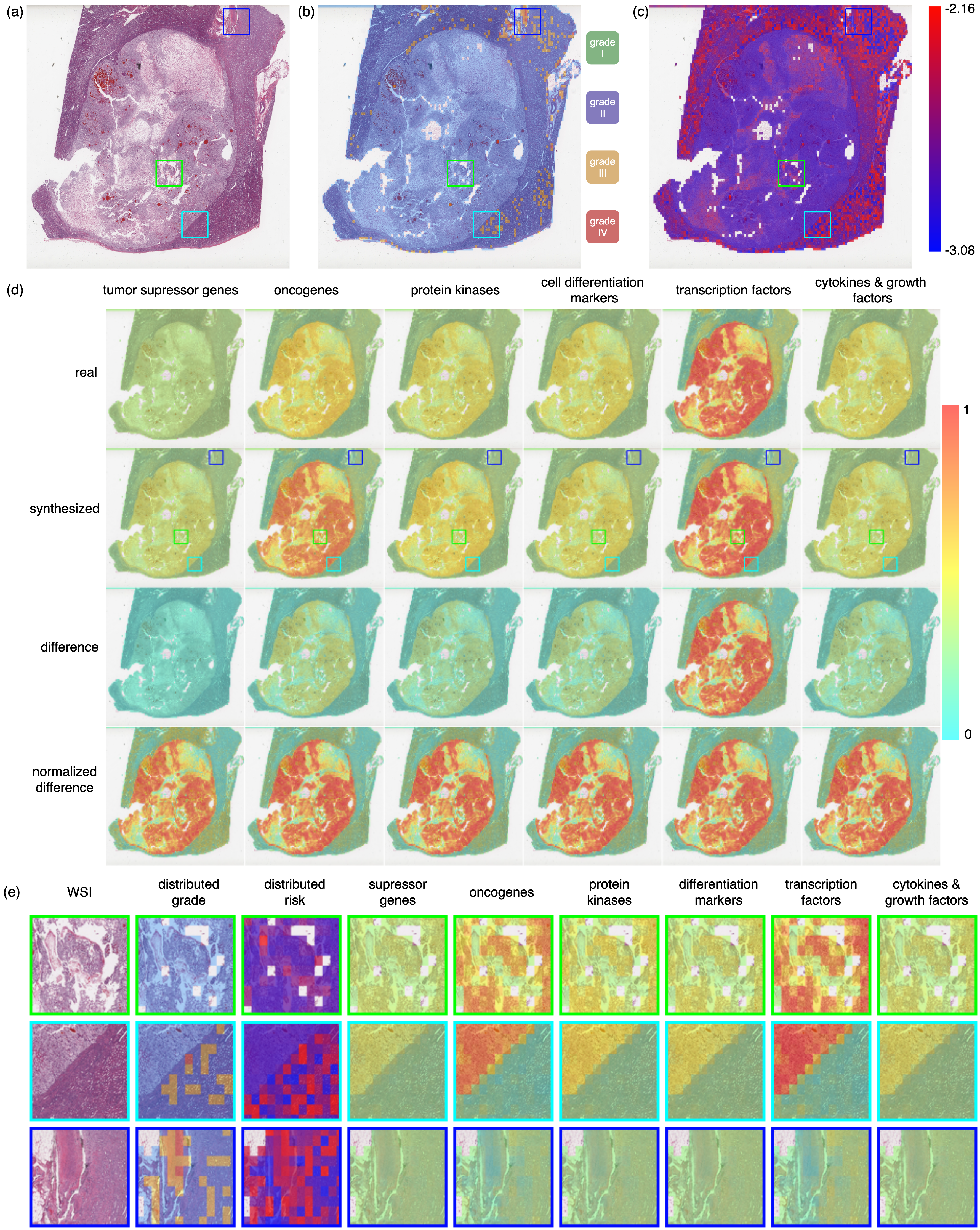}}
    \captionsetup{labelformat=default,labelsep=colon,name=Supplementary Figure}
    \caption{\textbf{Explainability maps for a female patient of age 50, ground truth grade II, survival time 22.78 months, and survival status alive from the TCGA-KIRC cohort}. (a) whole slide image (WSI) (b) intra-tumour gradation heterogeneity (c) intra-tumour survival heterogeneity (d) co-attention maps for real and synthesized transcriptomic data and WSI patches for different gene groups (e) comparative study of intra-tumour heterogeneity and corresponding co-attention maps obtained for prediction using synthesized transcriptomes for chosen regions marked with different coloured rectangles.  }
    \label{fig_kirc1}
\end{figure}

\begin{figure} [!t]
    \centerline{\includegraphics[scale=0.13]{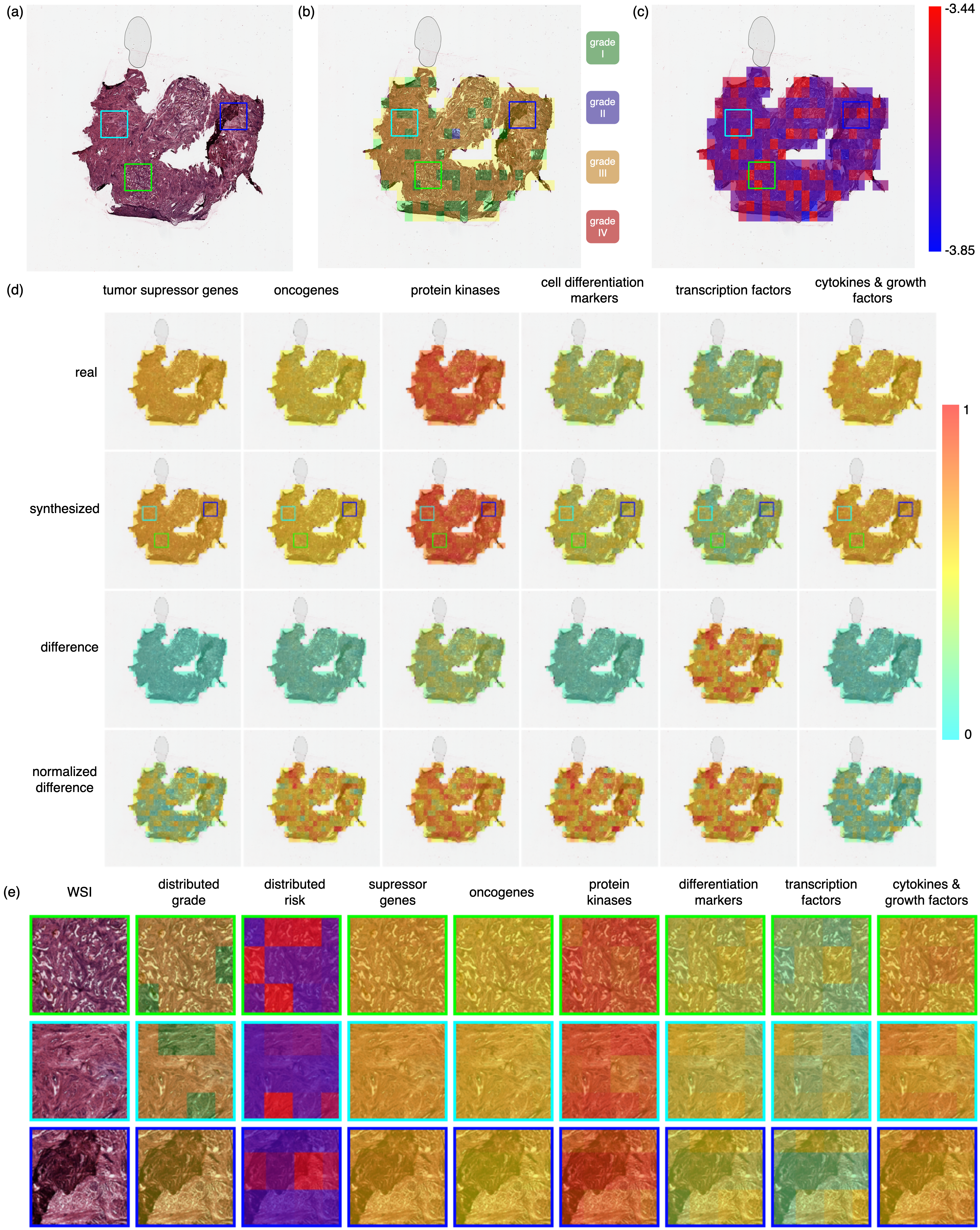}}
    \captionsetup{labelformat=default,labelsep=colon,name=Supplementary Figure}
    \caption{\textbf{Explainability maps for a female patient of ground truth grade III, survival time 58.08 months, and survival status alive from the CPTAC-UCEC cohort}. (a) whole slide image (WSI) (b) intra-tumour gradation heterogeneity (c) intra-tumour survival heterogeneity (d) co-attention maps for real and synthesized transcriptomic data and WSI patches for different gene groups (e) comparative study of intra-tumour heterogeneity and corresponding co-attention maps obtained for prediction using synthesized transcriptomes for chosen regions marked with different coloured rectangles.  }
    \label{fig_cptac_ucec}
\end{figure}

\begin{table}[!ht]
    \centering
    \captionsetup{labelformat=default,labelsep=colon,name=Supplementary Table}
    \caption{\label{result_GBMLGG} Table of uncertainty and fairness evaluation for \textbf{TCGA-GBMLGG} dataset. The $p$-values are computed using two-sided Wilcoxon rank-sum test to compare the grades and survival risks predicted using real and synthesized transcriptomic data, in addition to WSIs. \munc, \mcov, \cunc and \ccov refer to marginal uncertainty, marginal coverage, conditional uncertainty and conditional coverage, respectively. \textuparrow \ indicates the larger the better, and \textdownarrow \ indicates the smaller the better. Equivalent or close ($\pm 0.01$) performance using real and synthesized data is marked in \textbf{bold}. \\}
\resizebox{\textwidth}{!}{
\begin{tabular}{|r|l|ccccccccccccc|}
\hline
\multicolumn{1}{|c|}{\multirow{3}{*}{Category}} & \multicolumn{1}{c|}{\multirow{3}{*}{Group}} & \multicolumn{13}{c|}{Gradation} \\ \cline{3-15} 
\multicolumn{1}{|c|}{} & \multicolumn{1}{c|}{} & \multicolumn{5}{c|}{Real} & \multicolumn{5}{c|}{Synthesized} & \multicolumn{3}{c|}{p-value \textuparrow} \\ \cline{3-15} 
\multicolumn{1}{|c|}{} & \multicolumn{1}{c|}{} & \multicolumn{1}{c|}{AUC\textuparrow} & \multicolumn{1}{c|}{\munc\textdownarrow} & \multicolumn{1}{c|}{\mcov\textuparrow} & \multicolumn{1}{c|}{\cunc\textdownarrow} & \multicolumn{1}{c|}{\ccov\textuparrow} & \multicolumn{1}{c|}{AUC\textuparrow} & \multicolumn{1}{c|}{\munc\textdownarrow} & \multicolumn{1}{c|}{\mcov\textuparrow} & \multicolumn{1}{c|}{\cunc\textdownarrow} & \multicolumn{1}{c|}{\ccov\textuparrow} & \multicolumn{1}{c|}{grade} & \multicolumn{1}{c|}{\munc} & \cunc\\ \hline
overall & - & \multicolumn{1}{c|}{0.907} & \multicolumn{1}{c|}{0.407} & \multicolumn{1}{c|}{0.981} & \multicolumn{1}{c|}{-} & \multicolumn{1}{c|}{-} & \multicolumn{1}{c|}{0.890} & \multicolumn{1}{c|}{0.413} & \multicolumn{1}{c|}{0.981} & \multicolumn{1}{c|}{-} & \multicolumn{1}{c|}{-} & \multicolumn{1}{c|}{0.995} & \multicolumn{1}{c|}{0.935} & - \\ \hline
\multirow{2}{*}{gender} & male & \multicolumn{1}{c|}{0.915} & \multicolumn{1}{c|}{0.439} & \multicolumn{1}{c|}{0.985} & \multicolumn{1}{c|}{\multirow{2}{*}{0.312}} & \multicolumn{1}{c|}{\multirow{2}{*}{0.953}} & \multicolumn{1}{c|}{0.896} & \multicolumn{1}{c|}{\textbf{0.444}} & \multicolumn{1}{c|}{0.985} & \multicolumn{1}{c|}{\multirow{2}{*}{0.323}} & \multicolumn{1}{c|}{\multirow{2}{*}{0.953}} & \multicolumn{1}{c|}{0.975} & \multicolumn{1}{c|}{0.920} & \multirow{2}{*}{0.921} \\ \cline{2-5} \cline{8-10} \cline{13-14}
 & female & \multicolumn{1}{c|}{0.896} & \multicolumn{1}{c|}{0.184} & \multicolumn{1}{c|}{0.921} & \multicolumn{1}{c|}{} & \multicolumn{1}{c|}{} & \multicolumn{1}{c|}{\textbf{0.886}} & \multicolumn{1}{c|}{0.202} & \multicolumn{1}{c|}{0.921} & \multicolumn{1}{c|}{} & \multicolumn{1}{c|}{} & \multicolumn{1}{c|}{1.000} & \multicolumn{1}{c|}{0.921} &  \\ \hline
\multirow{3}{*}{age} & \textless 40 & \multicolumn{1}{c|}{0.883} & \multicolumn{1}{c|}{1.000} & \multicolumn{1}{c|}{1.000} & \multicolumn{1}{c|}{\multirow{3}{*}{0.727}} & \multicolumn{1}{c|}{\multirow{3}{*}{0.986}} & \multicolumn{1}{c|}{0.849} & \multicolumn{1}{c|}{\textbf{1.000}} & \multicolumn{1}{c|}{1.000} & \multicolumn{1}{c|}{\multirow{3}{*}{\textbf{0.720}}} & \multicolumn{1}{c|}{\multirow{3}{*}{0.986}} & \multicolumn{1}{c|}{0.736} & \multicolumn{1}{c|}{1.000} & \multirow{3}{*}{0.977} \\ \cline{2-5} \cline{8-10} \cline{13-14}
 & 40 - 60 & \multicolumn{1}{c|}{0.928} & \multicolumn{1}{c|}{0.459} & \multicolumn{1}{c|}{1.000} & \multicolumn{1}{c|}{} & \multicolumn{1}{c|}{} & \multicolumn{1}{c|}{\textbf{0.921}} & \multicolumn{1}{c|}{\textbf{0.450}} & \multicolumn{1}{c|}{1.000} & \multicolumn{1}{c|}{} & \multicolumn{1}{c|}{} & \multicolumn{1}{c|}{0.804} & \multicolumn{1}{c|}{0.996} &  \\ \cline{2-5} \cline{8-10} \cline{13-14}
 & \textgreater 60 & \multicolumn{1}{c|}{0.909} & \multicolumn{1}{c|}{0.722} & \multicolumn{1}{c|}{0.958} & \multicolumn{1}{c|}{} & \multicolumn{1}{c|}{} & \multicolumn{1}{c|}{\textbf{0.906}} & \multicolumn{1}{c|}{\textbf{0.708}} & \multicolumn{1}{c|}{0.958} & \multicolumn{1}{c|}{} & \multicolumn{1}{c|}{} & \multicolumn{1}{c|}{1.000} & \multicolumn{1}{c|}{0.934} &  \\ \hline
magni- & level1 & \multicolumn{1}{c|}{0.902} & \multicolumn{1}{c|}{0.195} & \multicolumn{1}{c|}{0.966} & \multicolumn{1}{c|}{\multirow{3}{*}{0.617}} & \multicolumn{1}{c|}{\multirow{3}{*}{0.989}} & \multicolumn{1}{c|}{0.885} & \multicolumn{1}{c|}{\textbf{0.172}} & \multicolumn{1}{c|}{0.931} & \multicolumn{1}{c|}{\multirow{3}{*}{\textbf{0.609}}} & \multicolumn{1}{c|}{\multirow{3}{*}{0.977}} & \multicolumn{1}{c|}{1.000} & \multicolumn{1}{c|}{0.652} & \multirow{3}{*}{0.884} \\ \cline{2-5} \cline{8-10} \cline{13-14}
fication & level2 & \multicolumn{1}{c|}{0.952} & \multicolumn{1}{c|}{0.754} & \multicolumn{1}{c|}{1.000} & \multicolumn{1}{c|}{} & \multicolumn{1}{c|}{} & \multicolumn{1}{c|}{0.936} & \multicolumn{1}{c|}{\textbf{0.754}} & \multicolumn{1}{c|}{1.000} & \multicolumn{1}{c|}{} & \multicolumn{1}{c|}{} & \multicolumn{1}{c|}{0.811} & \multicolumn{1}{c|}{1.000} &  \\ \cline{2-5} \cline{8-10} \cline{13-14}
\multicolumn{1}{|l|}{} & level3 & \multicolumn{1}{c|}{0.852} & \multicolumn{1}{c|}{0.901} & \multicolumn{1}{c|}{1.000} & \multicolumn{1}{c|}{} & \multicolumn{1}{c|}{} & \multicolumn{1}{c|}{0.836} & \multicolumn{1}{c|}{\textbf{0.901}} & \multicolumn{1}{c|}{1.000} & \multicolumn{1}{c|}{} & \multicolumn{1}{c|}{} & \multicolumn{1}{c|}{0.795} & \multicolumn{1}{c|}{1.000} &  \\ \hline
vital & alive & \multicolumn{1}{c|}{0.865} & \multicolumn{1}{c|}{0.990} & \multicolumn{1}{c|}{1.000} & \multicolumn{1}{c|}{\multirow{2}{*}{0.612}} & \multicolumn{1}{c|}{\multirow{2}{*}{0.986}} & \multicolumn{1}{c|}{0.832} & \multicolumn{1}{c|}{\textbf{1.000}} & \multicolumn{1}{c|}{1.000} & \multicolumn{1}{c|}{\multirow{2}{*}{\textbf{0.617}}} & \multicolumn{1}{c|}{\multirow{2}{*}{0.986}} & \multicolumn{1}{c|}{0.894} & \multicolumn{1}{c|}{0.881} & \multirow{2}{*}{0.941} \\ \cline{2-5} \cline{8-10} \cline{13-14}
status & dead & \multicolumn{1}{c|}{0.951} & \multicolumn{1}{c|}{0.234} & \multicolumn{1}{c|}{0.973} & \multicolumn{1}{c|}{} & \multicolumn{1}{c|}{} & \multicolumn{1}{c|}{\textbf{0.951}} & \multicolumn{1}{c|}{\textbf{0.234}} & \multicolumn{1}{c|}{0.973} & \multicolumn{1}{c|}{} & \multicolumn{1}{c|}{} & \multicolumn{1}{c|}{0.867} & \multicolumn{1}{c|}{1.000} &  \\ \hline
\multirow{3}{*}{grade} & grade II & \multicolumn{1}{c|}{0.914} & \multicolumn{1}{c|}{0.547} & \multicolumn{1}{c|}{0.974} & \multicolumn{1}{c|}{\multirow{3}{*}{0.527}} & \multicolumn{1}{c|}{\multirow{3}{*}{0.980}} & \multicolumn{1}{c|}{0.891} & \multicolumn{1}{c|}{0.598} & \multicolumn{1}{c|}{0.974} & \multicolumn{1}{c|}{\multirow{3}{*}{0.544}} & \multicolumn{1}{c|}{\multirow{3}{*}{0.980}} & \multicolumn{1}{c|}{0.704} & \multicolumn{1}{c|}{0.569} & \multirow{3}{*}{0.856} \\ \cline{2-5} \cline{8-10} \cline{13-14}
 & grade III & \multicolumn{1}{c|}{0.755} & \multicolumn{1}{c|}{1.000} & \multicolumn{1}{c|}{1.000} & \multicolumn{1}{c|}{} & \multicolumn{1}{c|}{} & \multicolumn{1}{c|}{0.728} & \multicolumn{1}{c|}{\textbf{1.000}} & \multicolumn{1}{c|}{1.000} & \multicolumn{1}{c|}{} & \multicolumn{1}{c|}{} & \multicolumn{1}{c|}{0.866} & \multicolumn{1}{c|}{1.000} &  \\ \cline{2-5} \cline{8-10} \cline{13-14}
 & grade IV & \multicolumn{1}{c|}{0.995} & \multicolumn{1}{c|}{0.034} & \multicolumn{1}{c|}{0.966} & \multicolumn{1}{c|}{} & \multicolumn{1}{c|}{} & \multicolumn{1}{c|}{\textbf{0.993}} & \multicolumn{1}{c|}{\textbf{0.034}} & \multicolumn{1}{c|}{0.966} & \multicolumn{1}{c|}{} & \multicolumn{1}{c|}{} & \multicolumn{1}{c|}{0.822} & \multicolumn{1}{c|}{1.000} &  \\ \hline
\multirow{3}{*}{time bin} & bin 1 & \multicolumn{1}{c|}{0.898} & \multicolumn{1}{c|}{0.431} & \multicolumn{1}{c|}{0.976} & \multicolumn{1}{c|}{\multirow{3}{*}{0.335}} & \multicolumn{1}{c|}{\multirow{3}{*}{0.992}} & \multicolumn{1}{c|}{0.882} & \multicolumn{1}{c|}{\textbf{0.407}} & \multicolumn{1}{c|}{0.963} & \multicolumn{1}{c|}{\multirow{3}{*}{\textbf{0.331}}} & \multicolumn{1}{c|}{\multirow{3}{*}{0.988}} & \multicolumn{1}{c|}{0.984} & \multicolumn{1}{c|}{0.790} & \multirow{3}{*}{0.766} \\ \cline{2-5} \cline{8-10} \cline{13-14}
 & bin 2 & \multicolumn{1}{c|}{0.786} & \multicolumn{1}{c|}{0.538} & \multicolumn{1}{c|}{1.000} & \multicolumn{1}{c|}{} & \multicolumn{1}{c|}{} & \multicolumn{1}{c|}{\textbf{0.784}} & \multicolumn{1}{c|}{\textbf{0.513}} & \multicolumn{1}{c|}{1.000} & \multicolumn{1}{c|}{} & \multicolumn{1}{c|}{} & \multicolumn{1}{c|}{1.000} & \multicolumn{1}{c|}{0.817} &  \\ \cline{2-5} \cline{8-10} \cline{13-14}
 & bin 3 & \multicolumn{1}{c|}{1.000} & \multicolumn{1}{c|}{0.037} & \multicolumn{1}{c|}{1.000} & \multicolumn{1}{c|}{} & \multicolumn{1}{c|}{} & \multicolumn{1}{c|}{\textbf{1.000}} & \multicolumn{1}{c|}{0.074} & \multicolumn{1}{c|}{1.000} & \multicolumn{1}{c|}{} & \multicolumn{1}{c|}{} & \multicolumn{1}{c|}{1.000} & \multicolumn{1}{c|}{0.691} &  \\ \hline \hline
\multicolumn{1}{|c|}{\multirow{3}{*}{Category}} & \multicolumn{1}{c|}{\multirow{3}{*}{Group}} & \multicolumn{13}{c|}{Risk Estimation} \\ \cline{3-15} 
\multicolumn{1}{|c|}{} & \multicolumn{1}{c|}{} & \multicolumn{5}{c|}{Real} & \multicolumn{5}{c|}{Synthesized} & \multicolumn{3}{c|}{p-value \textuparrow} \\ \cline{3-15} 
\multicolumn{1}{|c|}{} & \multicolumn{1}{c|}{} & \multicolumn{1}{c|}{C Index\textuparrow} & \multicolumn{1}{c|}{\munc\textdownarrow} & \multicolumn{1}{c|}{\mcov\textuparrow} & \multicolumn{1}{c|}{\cunc\textdownarrow} & \multicolumn{1}{c|}{\ccov\textuparrow} & \multicolumn{1}{c|}{C Index\textuparrow} & \multicolumn{1}{c|}{\munc\textdownarrow} & \multicolumn{1}{c|}{\mcov\textuparrow} & \multicolumn{1}{c|}{\cunc\textdownarrow} & \multicolumn{1}{c|}{\ccov\textuparrow} & \multicolumn{1}{c|}{grade} & \multicolumn{1}{c|}{\munc} & \cunc  \\ \hline
overall & - & \multicolumn{1}{c|}{0.866} & \multicolumn{1}{c|}{0.446} & \multicolumn{1}{c|}{0.933} & \multicolumn{1}{c|}{-} & \multicolumn{1}{c|}{-} & \multicolumn{1}{c|}{\textbf{0.861}} & \multicolumn{1}{c|}{\textbf{0.438}} & \multicolumn{1}{c|}{0.923} & \multicolumn{1}{c|}{-} & \multicolumn{1}{c|}{-} & \multicolumn{1}{c|}{0.604} & \multicolumn{1}{c|}{0.652} & - \\ \hline
\multirow{2}{*}{gender} & male & \multicolumn{1}{c|}{0.808} & \multicolumn{1}{c|}{0.375} & \multicolumn{1}{c|}{0.985} & \multicolumn{1}{c|}{\multirow{2}{*}{0.514}} & \multicolumn{1}{c|}{\multirow{2}{*}{0.953}} & \multicolumn{1}{c|}{\textbf{0.813}} & \multicolumn{1}{c|}{\textbf{0.376}} & \multicolumn{1}{c|}{0.985} & \multicolumn{1}{c|}{\multirow{2}{*}{\textbf{0.507}}} & \multicolumn{1}{c|}{\multirow{2}{*}{0.953}} & \multicolumn{1}{c|}{0.956} & \multicolumn{1}{c|}{0.956} & \multirow{2}{*}{0.678} \\ \cline{2-5} \cline{8-10} \cline{13-14}
 & female & \multicolumn{1}{c|}{0.795} & \multicolumn{1}{c|}{0.653} & \multicolumn{1}{c|}{0.921} & \multicolumn{1}{c|}{} & \multicolumn{1}{c|}{} & \multicolumn{1}{c|}{\textbf{0.795}} & \multicolumn{1}{c|}{\textbf{0.639}} & \multicolumn{1}{c|}{0.921} & \multicolumn{1}{c|}{} & \multicolumn{1}{c|}{} & \multicolumn{1}{c|}{0.350} & \multicolumn{1}{c|}{0.400} &  \\ \hline
\multirow{3}{*}{age} & \textless 40 & \multicolumn{1}{c|}{0.524} & \multicolumn{1}{c|}{0.767} & \multicolumn{1}{c|}{0.977} & \multicolumn{1}{c|}{\multirow{3}{*}{0.541}} & \multicolumn{1}{c|}{\multirow{3}{*}{0.983}} & \multicolumn{1}{c|}{\textbf{0.540}} & \multicolumn{1}{c|}{\textbf{0.727}} & \multicolumn{1}{c|}{1.000} & \multicolumn{1}{c|}{\multirow{3}{*}{\textbf{0.535}}} & \multicolumn{1}{c|}{\multirow{3}{*}{0.991}} & \multicolumn{1}{c|}{0.119} & \multicolumn{1}{c|}{0.135} & \multirow{3}{*}{0.546} \\ \cline{2-5} \cline{8-10} \cline{13-14}
 & 40 - 60 & \multicolumn{1}{c|}{0.848} & \multicolumn{1}{c|}{0.480} & \multicolumn{1}{c|}{0.973} & \multicolumn{1}{c|}{} & \multicolumn{1}{c|}{} & \multicolumn{1}{c|}{\textbf{0.845}} & \multicolumn{1}{c|}{0.492} & \multicolumn{1}{c|}{0.973} & \multicolumn{1}{c|}{} & \multicolumn{1}{c|}{} & \multicolumn{1}{c|}{0.775} & \multicolumn{1}{c|}{0.770} &  \\ \cline{2-5} \cline{8-10} \cline{13-14}
 & \textgreater 60 & \multicolumn{1}{c|}{0.738} & \multicolumn{1}{c|}{0.377} & \multicolumn{1}{c|}{1.000} & \multicolumn{1}{c|}{} & \multicolumn{1}{c|}{} & \multicolumn{1}{c|}{\textbf{0.742}} & \multicolumn{1}{c|}{\textbf{0.387}} & \multicolumn{1}{c|}{1.000} & \multicolumn{1}{c|}{} & \multicolumn{1}{c|}{} & \multicolumn{1}{c|}{0.741} & \multicolumn{1}{c|}{0.734} &  \\ \hline
magni- & level1 & \multicolumn{1}{c|}{0.855} & \multicolumn{1}{c|}{0.735} & \multicolumn{1}{c|}{0.966} & \multicolumn{1}{c|}{\multirow{3}{*}{0.586}} & \multicolumn{1}{c|}{\multirow{3}{*}{0.971}} & \multicolumn{1}{c|}{0.840} & \multicolumn{1}{c|}{\textbf{0.725}} & \multicolumn{1}{c|}{1.000} & \multicolumn{1}{c|}{\multirow{3}{*}{\textbf{0.576}}} & \multicolumn{1}{c|}{\multirow{3}{*}{0.982}} & \multicolumn{1}{c|}{0.646} & \multicolumn{1}{c|}{0.810} & \multirow{3}{*}{0.787} \\ \cline{2-5} \cline{8-10} \cline{13-14}
fication & level2 & \multicolumn{1}{c|}{0.862} & \multicolumn{1}{c|}{0.484} & \multicolumn{1}{c|}{0.974} & \multicolumn{1}{c|}{} & \multicolumn{1}{c|}{} & \multicolumn{1}{c|}{\textbf{0.856}} & \multicolumn{1}{c|}{\textbf{0.486}} & \multicolumn{1}{c|}{0.974} & \multicolumn{1}{c|}{} & \multicolumn{1}{c|}{} & \multicolumn{1}{c|}{0.950} & \multicolumn{1}{c|}{0.975} &  \\ \cline{2-5} \cline{8-10} \cline{13-14}
\multicolumn{1}{|l|}{} & level3 & \multicolumn{1}{c|}{0.873} & \multicolumn{1}{c|}{0.540} & \multicolumn{1}{c|}{0.973} & \multicolumn{1}{c|}{} & \multicolumn{1}{c|}{} & \multicolumn{1}{c|}{0.858} & \multicolumn{1}{c|}{\textbf{0.517}} & \multicolumn{1}{c|}{0.973} & \multicolumn{1}{c|}{} & \multicolumn{1}{c|}{} & \multicolumn{1}{c|}{0.578} & \multicolumn{1}{c|}{0.578} &  \\ \hline
vital & alive & \multicolumn{1}{c|}{-} & \multicolumn{1}{c|}{0.368} & \multicolumn{1}{c|}{0.851} & \multicolumn{1}{c|}{\multirow{2}{*}{0.336}} & \multicolumn{1}{c|}{\multirow{2}{*}{0.912}} & \multicolumn{1}{c|}{-} & \multicolumn{1}{c|}{\textbf{0.359}} & \multicolumn{1}{c|}{0.851} & \multicolumn{1}{c|}{\multirow{2}{*}{\textbf{0.337}}} & \multicolumn{1}{c|}{\multirow{2}{*}{0.925}} & \multicolumn{1}{c|}{0.237} & \multicolumn{1}{c|}{0.441} & \multirow{2}{*}{0.502} \\ \cline{2-5} \cline{8-10} \cline{13-14}
status & dead & \multicolumn{1}{c|}{0.614} & \multicolumn{1}{c|}{0.304} & \multicolumn{1}{c|}{0.973} & \multicolumn{1}{c|}{} & \multicolumn{1}{c|}{} & \multicolumn{1}{c|}{0.633} & \multicolumn{1}{c|}{\textbf{0.314}} & \multicolumn{1}{c|}{1.000} & \multicolumn{1}{c|}{} & \multicolumn{1}{c|}{} & \multicolumn{1}{c|}{0.563} & \multicolumn{1}{c|}{0.563} &  \\ \hline
\multirow{3}{*}{grade} & grade II & \multicolumn{1}{c|}{0.823} & \multicolumn{1}{c|}{0.582} & \multicolumn{1}{c|}{0.872} & \multicolumn{1}{c|}{\multirow{3}{*}{0.431}} & \multicolumn{1}{c|}{\multirow{3}{*}{0.937}} & \multicolumn{1}{c|}{\textbf{0.823}} & \multicolumn{1}{c|}{\textbf{0.544}} & \multicolumn{1}{c|}{0.824} & \multicolumn{1}{c|}{\multirow{3}{*}{0.420}} & \multicolumn{1}{c|}{\multirow{3}{*}{0.920}} & \multicolumn{1}{c|}{0.058} & \multicolumn{1}{c|}{0.054} & \multirow{3}{*}{0.471} \\ \cline{2-5} \cline{8-10} \cline{13-14}
 & grade III & \multicolumn{1}{c|}{0.590} & \multicolumn{1}{c|}{0.655} & \multicolumn{1}{c|}{0.972} & \multicolumn{1}{c|}{} & \multicolumn{1}{c|}{} & \multicolumn{1}{c|}{0.526} & \multicolumn{1}{c|}{\textbf{0.659}} & \multicolumn{1}{c|}{0.972} & \multicolumn{1}{c|}{} & \multicolumn{1}{c|}{} & \multicolumn{1}{c|}{0.901} & \multicolumn{1}{c|}{0.866} &  \\ \cline{2-5} \cline{8-10} \cline{13-14}
 & grade IV & \multicolumn{1}{c|}{0.509} & \multicolumn{1}{c|}{0.056} & \multicolumn{1}{c|}{0.966} & \multicolumn{1}{c|}{} & \multicolumn{1}{c|}{} & \multicolumn{1}{c|}{\textbf{0.539}} & \multicolumn{1}{c|}{\textbf{0.059}} & \multicolumn{1}{c|}{0.966} & \multicolumn{1}{c|}{} & \multicolumn{1}{c|}{} & \multicolumn{1}{c|}{0.489} & \multicolumn{1}{c|}{0.494} &  \\ \hline
\multirow{3}{*}{time bin} & bin 1 & \multicolumn{1}{c|}{0.827} & \multicolumn{1}{c|}{0.252} & \multicolumn{1}{c|}{0.951} & \multicolumn{1}{c|}{\multirow{3}{*}{0.684}} & \multicolumn{1}{c|}{\multirow{3}{*}{0.958}} & \multicolumn{1}{c|}{\textbf{0.822}} & \multicolumn{1}{c|}{\textbf{0.259}} & \multicolumn{1}{c|}{0.939} & \multicolumn{1}{c|}{\multirow{3}{*}{\textbf{0.675}}} & \multicolumn{1}{c|}{\multirow{3}{*}{0.954}} & \multicolumn{1}{c|}{0.942} & \multicolumn{1}{c|}{0.893} & \multirow{3}{*}{0.465} \\ \cline{2-5} \cline{8-10} \cline{13-14}
 & bin 2 & \multicolumn{1}{c|}{-} & \multicolumn{1}{c|}{0.804} & \multicolumn{1}{c|}{0.923} & \multicolumn{1}{c|}{} & \multicolumn{1}{c|}{} & \multicolumn{1}{c|}{-} & \multicolumn{1}{c|}{\textbf{0.782}} & \multicolumn{1}{c|}{0.923} & \multicolumn{1}{c|}{} & \multicolumn{1}{c|}{} & \multicolumn{1}{c|}{0.209} & \multicolumn{1}{c|}{0.270} &  \\ \cline{2-5} \cline{8-10} \cline{13-14}
 & bin 3 & \multicolumn{1}{c|}{-} & \multicolumn{1}{c|}{0.996} & \multicolumn{1}{c|}{1.000} & \multicolumn{1}{c|}{} & \multicolumn{1}{c|}{} & \multicolumn{1}{c|}{-} & \multicolumn{1}{c|}{\textbf{0.984}} & \multicolumn{1}{c|}{1.000} & \multicolumn{1}{c|}{} & \multicolumn{1}{c|}{} & \multicolumn{1}{c|}{0.200} & \multicolumn{1}{c|}{0.233} &  \\ \hline
\end{tabular}
}    
\end{table}

\begin{table}[!ht]
    \centering
    \captionsetup{labelformat=default,labelsep=colon,name=Supplementary Table}
    \caption{\label{result_GBM} Table of uncertainty and fairness evaluation for \textbf{CPTAC-GBM} dataset. p-values are computed using the two-sided Wilcoxon rank-sum test to compare the grades and survival risks predicted using real and synthesized transcriptomic data, in addition to WSIs. \munc, \mcov, \cunc and \ccov refer to marginal uncertainty, marginal coverage, conditional uncertainty and conditional coverage, respectively. \textuparrow \ indicates the larger the better, and \textdownarrow \ indicates the smaller the better. Equivalent or close ($\pm 0.01$) performance using real and synthesized data is marked in \textbf{bold}. \\}
    \resizebox{\textwidth}{!}{
\begin{tabular}{|r|l|ccccccccccccc|}
\hline
\multicolumn{1}{|c|}{\multirow{3}{*}{Category}} & \multicolumn{1}{c|}{\multirow{3}{*}{Group}} & \multicolumn{13}{c|}{Gradation} \\ \cline{3-15} 
\multicolumn{1}{|c|}{} & \multicolumn{1}{c|}{} & \multicolumn{5}{c|}{Real} & \multicolumn{5}{c|}{Synthesized} & \multicolumn{3}{c|}{p-value \textuparrow} \\ \cline{3-15} 
\multicolumn{1}{|c|}{} & \multicolumn{1}{c|}{} & \multicolumn{1}{c|}{AUC \textuparrow} & \multicolumn{1}{c|}{\munc \textdownarrow} & \multicolumn{1}{c|}{\mcov \textuparrow} & \multicolumn{1}{c|}{\cunc \textdownarrow} & \multicolumn{1}{c|}{\ccov \textuparrow} & \multicolumn{1}{c|}{AUC \textuparrow} & \multicolumn{1}{c|}{\munc \textdownarrow} & \multicolumn{1}{c|}{\mcov \textuparrow} & \multicolumn{1}{c|}{\cunc \textdownarrow} & \multicolumn{1}{c|}{\ccov \textuparrow} & \multicolumn{1}{c|}{grade} & \multicolumn{1}{c|}{\munc} & \cunc \\ \hline
overall & - & \multicolumn{1}{c|}{0.863} & \multicolumn{1}{c|}{0.465} & \multicolumn{1}{c|}{0.932} & \multicolumn{1}{l|}{} & \multicolumn{1}{c|}{-} & \multicolumn{1}{c|}{\textbf{0.865}} & \multicolumn{1}{c|}{\textbf{0.465}} & \multicolumn{1}{c|}{0.936} & \multicolumn{1}{c|}{-} & \multicolumn{1}{c|}{-} & \multicolumn{1}{c|}{0.985} & \multicolumn{1}{c|}{0.991} & - \\ \hline
\multirow{2}{*}{gender} & male & \multicolumn{1}{c|}{0.847} & \multicolumn{1}{c|}{0.514} & \multicolumn{1}{c|}{0.972} & \multicolumn{1}{c|}{\multirow{2}{*}{0.495}} & \multicolumn{1}{c|}{\multirow{2}{*}{0.962}} & \multicolumn{1}{c|}{\textbf{0.847}} & \multicolumn{1}{c|}{\textbf{0.515}} & \multicolumn{1}{c|}{0.975} & \multicolumn{1}{c|}{\multirow{2}{*}{\textbf{0.494}}} & \multicolumn{1}{c|}{\multirow{2}{*}{0.964}} & \multicolumn{1}{c|}{0.975} & \multicolumn{1}{c|}{0.968} & \multirow{2}{*}{0.961} \\ \cline{2-5} \cline{8-10} \cline{13-14}
 & female & \multicolumn{1}{c|}{0.894} & \multicolumn{1}{c|}{0.476} & \multicolumn{1}{c|}{0.953} & \multicolumn{1}{c|}{} & \multicolumn{1}{c|}{} & \multicolumn{1}{c|}{\textbf{0.898}} & \multicolumn{1}{c|}{\textbf{0.472}} & \multicolumn{1}{c|}{0.953} & \multicolumn{1}{c|}{} & \multicolumn{1}{c|}{} & \multicolumn{1}{c|}{0.935} & \multicolumn{1}{c|}{0.955} &  \\ \hline
magni- & level1 & \multicolumn{1}{c|}{0.861} & \multicolumn{1}{c|}{0.430} & \multicolumn{1}{c|}{0.940} & \multicolumn{1}{c|}{\multirow{3}{*}{0.478}} & \multicolumn{1}{c|}{\multirow{3}{*}{0.949}} & \multicolumn{1}{c|}{\textbf{0.864}} & \multicolumn{1}{c|}{\textbf{0.435}} & \multicolumn{1}{c|}{0.947} & \multicolumn{1}{c|}{\multirow{3}{*}{\textbf{0.483}}} & \multicolumn{1}{c|}{\multirow{3}{*}{0.956}} & \multicolumn{1}{c|}{0.918} & \multicolumn{1}{c|}{0.958} & \multirow{3}{*}{0.938} \\ \cline{2-5} \cline{8-10} \cline{13-14}
fication & level2 & \multicolumn{1}{c|}{0.865} & \multicolumn{1}{c|}{0.506} & \multicolumn{1}{c|}{0.954} & \multicolumn{1}{c|}{} & \multicolumn{1}{c|}{} & \multicolumn{1}{c|}{\textbf{0.867}} & \multicolumn{1}{c|}{\textbf{0.510}} & \multicolumn{1}{c|}{0.960} & \multicolumn{1}{c|}{} & \multicolumn{1}{c|}{} & \multicolumn{1}{c|}{0.931} & \multicolumn{1}{c|}{0.949} &  \\ \cline{2-5} \cline{8-10} \cline{13-14}
\multicolumn{1}{|l|}{} & level3 & \multicolumn{1}{c|}{0.869} & \multicolumn{1}{c|}{0.497} & \multicolumn{1}{c|}{0.954} & \multicolumn{1}{c|}{} & \multicolumn{1}{c|}{} & \multicolumn{1}{c|}{\textbf{0.870}} & \multicolumn{1}{c|}{\textbf{0.503}} & \multicolumn{1}{c|}{0.960} & \multicolumn{1}{c|}{} & \multicolumn{1}{c|}{} & \multicolumn{1}{c|}{1.000} & \multicolumn{1}{c|}{0.905} &  \\ \hline
vital & alive & \multicolumn{1}{c|}{0.835} & \multicolumn{1}{c|}{0.463} & \multicolumn{1}{c|}{0.903} & \multicolumn{1}{c|}{\multirow{2}{*}{0.478}} & \multicolumn{1}{c|}{\multirow{2}{*}{0.932}} & \multicolumn{1}{c|}{\textbf{0.833}} & \multicolumn{1}{c|}{\textbf{0.465}} & \multicolumn{1}{c|}{0.903} & \multicolumn{1}{c|}{\multirow{2}{*}{\textbf{0.482}}} & \multicolumn{1}{c|}{\multirow{2}{*}{0.938}} & \multicolumn{1}{c|}{0.970} & \multicolumn{1}{c|}{1.000} & \multirow{2}{*}{0.947} \\ \cline{2-5} \cline{8-10} \cline{13-14}
status & dead & \multicolumn{1}{c|}{0.880} & \multicolumn{1}{c|}{0.493} & \multicolumn{1}{c|}{0.962} & \multicolumn{1}{c|}{} & \multicolumn{1}{c|}{} & \multicolumn{1}{c|}{\textbf{0.883}} & \multicolumn{1}{c|}{\textbf{0.499}} & \multicolumn{1}{c|}{0.972} & \multicolumn{1}{c|}{} & \multicolumn{1}{c|}{} & \multicolumn{1}{c|}{0.952} & \multicolumn{1}{c|}{0.894} &  \\ \hline
grade & grade IV & \multicolumn{1}{c|}{0.863} & \multicolumn{1}{c|}{0.465} & \multicolumn{1}{c|}{0.932} & \multicolumn{1}{c|}{-} & \multicolumn{1}{c|}{-} & \multicolumn{1}{c|}{\textbf{0.865}} & \multicolumn{1}{c|}{\textbf{0.465}} & \multicolumn{1}{c|}{0.936} & \multicolumn{1}{c|}{-} & \multicolumn{1}{c|}{-} & \multicolumn{1}{c|}{0.985} & \multicolumn{1}{c|}{0.991} & - \\ \hline
time bin & bin 1 & \multicolumn{1}{c|}{0.863} & \multicolumn{1}{c|}{0.465} & \multicolumn{1}{c|}{0.932} & \multicolumn{1}{c|}{-} & \multicolumn{1}{c|}{-} & \multicolumn{1}{c|}{\textbf{0.865}} & \multicolumn{1}{c|}{\textbf{0.465}} & \multicolumn{1}{c|}{0.936} & \multicolumn{1}{c|}{-} & \multicolumn{1}{c|}{-} & \multicolumn{1}{c|}{0.985} & \multicolumn{1}{c|}{0.991} & - \\ \hline \hline
\multicolumn{1}{|c|}{\multirow{3}{*}{Category}} & \multicolumn{1}{c|}{\multirow{3}{*}{Group}} & \multicolumn{13}{c|}{Risk Estimation} \\ \cline{3-15} 
\multicolumn{1}{|c|}{} & \multicolumn{1}{c|}{} & \multicolumn{5}{c|}{Real} & \multicolumn{5}{c|}{Synthesized} & \multicolumn{3}{c|}{p-value \textuparrow} \\ \cline{3-15} 
\multicolumn{1}{|c|}{} & \multicolumn{1}{c|}{} & \multicolumn{1}{c|}{C Index \textuparrow} & \multicolumn{1}{c|}{\munc \textdownarrow} & \multicolumn{1}{c|}{\mcov \textuparrow} & \multicolumn{1}{c|}{\cunc \textdownarrow} & \multicolumn{1}{c|}{\ccov \textuparrow} & \multicolumn{1}{c|}{C Index \textuparrow} & \multicolumn{1}{c|}{\munc \textdownarrow} & \multicolumn{1}{c|}{\mcov \textuparrow} & \multicolumn{1}{c|}{\cunc \textdownarrow} & \multicolumn{1}{c|}{\ccov \textuparrow} & \multicolumn{1}{c|}{risk} & \multicolumn{1}{c|}{\munc} & \cunc \\ \hline
overall & - & \multicolumn{1}{c|}{0.564} & \multicolumn{1}{c|}{0.206} & \multicolumn{1}{c|}{0.971} & \multicolumn{1}{c|}{-} & \multicolumn{1}{c|}{-} & \multicolumn{1}{c|}{\textbf{0.565}} & \multicolumn{1}{c|}{\textbf{0.206}} & \multicolumn{1}{c|}{0.971} & \multicolumn{1}{c|}{-} & \multicolumn{1}{c|}{-} & \multicolumn{1}{c|}{0.815} & \multicolumn{1}{c|}{0.811} & - \\ \hline
\multirow{2}{*}{gender} & male & \multicolumn{1}{c|}{0.653} & \multicolumn{1}{c|}{0.223} & \multicolumn{1}{c|}{0.972} & \multicolumn{1}{c|}{\multirow{2}{*}{0.205}} & \multicolumn{1}{c|}{\multirow{2}{*}{0.971}} & \multicolumn{1}{c|}{\textbf{0.651}} & \multicolumn{1}{c|}{\textbf{0.222}} & \multicolumn{1}{c|}{0.975} & \multicolumn{1}{c|}{\multirow{2}{*}{\textbf{0.204}}} & \multicolumn{1}{c|}{\multirow{2}{*}{0.973}} & \multicolumn{1}{c|}{0.809} & \multicolumn{1}{c|}{0.811} & \multirow{2}{*}{0.870} \\ \cline{2-5} \cline{8-10} \cline{13-14}
 & female & \multicolumn{1}{c|}{0.430} & \multicolumn{1}{c|}{0.186} & \multicolumn{1}{c|}{0.971} & \multicolumn{1}{c|}{} & \multicolumn{1}{c|}{} & \multicolumn{1}{c|}{\textbf{0.432}} & \multicolumn{1}{c|}{\textbf{0.185}} & \multicolumn{1}{c|}{0.971} & \multicolumn{1}{c|}{} & \multicolumn{1}{c|}{} & \multicolumn{1}{c|}{0.931} & \multicolumn{1}{c|}{0.929} &  \\ \hline
magni- & level1 & \multicolumn{1}{c|}{0.544} & \multicolumn{1}{c|}{0.208} & \multicolumn{1}{c|}{0.980} & \multicolumn{1}{c|}{\multirow{3}{*}{0.208}} & \multicolumn{1}{c|}{\multirow{3}{*}{0.971}} & \multicolumn{1}{c|}{\textbf{0.543}} & \multicolumn{1}{c|}{\textbf{0.207}} & \multicolumn{1}{c|}{0.987} & \multicolumn{1}{c|}{\multirow{3}{*}{\textbf{0.208}}} & \multicolumn{1}{c|}{\multirow{3}{*}{0.976}} & \multicolumn{1}{c|}{0.718} & \multicolumn{1}{c|}{0.719} & \multirow{3}{*}{0.852} \\ \cline{2-5} \cline{8-10} \cline{13-14}
fication & level2 & \multicolumn{1}{c|}{0.574} & \multicolumn{1}{c|}{0.209} & \multicolumn{1}{c|}{0.967} & \multicolumn{1}{c|}{} & \multicolumn{1}{c|}{} & \multicolumn{1}{c|}{\textbf{0.578}} & \multicolumn{1}{c|}{\textbf{0.209}} & \multicolumn{1}{c|}{0.974} & \multicolumn{1}{c|}{} & \multicolumn{1}{c|}{} & \multicolumn{1}{c|}{0.931} & \multicolumn{1}{c|}{0.937} &  \\ \cline{2-5} \cline{8-10} \cline{13-14}
\multicolumn{1}{|l|}{} & level3 & \multicolumn{1}{c|}{0.570} & \multicolumn{1}{c|}{0.208} & \multicolumn{1}{c|}{0.967} & \multicolumn{1}{c|}{} & \multicolumn{1}{c|}{} & \multicolumn{1}{c|}{\textbf{0.572}} & \multicolumn{1}{c|}{\textbf{0.209}} & \multicolumn{1}{c|}{0.967} & \multicolumn{1}{c|}{} & \multicolumn{1}{c|}{} & \multicolumn{1}{c|}{0.900} & \multicolumn{1}{c|}{0.899} &  \\ \hline
vital & alive & \multicolumn{1}{c|}{-} & \multicolumn{1}{c|}{0.183} & \multicolumn{1}{c|}{0.970} & \multicolumn{1}{c|}{\multirow{2}{*}{0.195}} & \multicolumn{1}{c|}{\multirow{2}{*}{0.964}} & \multicolumn{1}{c|}{\textbf{-}} & \multicolumn{1}{c|}{\textbf{0.184}} & \multicolumn{1}{c|}{0.982} & \multicolumn{1}{c|}{\multirow{2}{*}{\textbf{0.195}}} & \multicolumn{1}{c|}{\multirow{2}{*}{0.974}} & \multicolumn{1}{c|}{0.821} & \multicolumn{1}{c|}{0.834} & \multirow{2}{*}{0.853} \\ \cline{2-5} \cline{8-10} \cline{13-14}
status & dead & \multicolumn{1}{c|}{0.582} & \multicolumn{1}{c|}{0.207} & \multicolumn{1}{c|}{0.958} & \multicolumn{1}{c|}{} & \multicolumn{1}{c|}{} & \multicolumn{1}{c|}{\textbf{0.581}} & \multicolumn{1}{c|}{\textbf{0.206}} & \multicolumn{1}{c|}{0.965} & \multicolumn{1}{c|}{} & \multicolumn{1}{c|}{} & \multicolumn{1}{c|}{0.870} & \multicolumn{1}{c|}{0.873} &  \\ \hline
grade & grade IV & \multicolumn{1}{c|}{0.564} & \multicolumn{1}{c|}{0.206} & \multicolumn{1}{c|}{0.971} & \multicolumn{1}{c|}{-} & \multicolumn{1}{c|}{-} & \multicolumn{1}{c|}{\textbf{0.565}} & \multicolumn{1}{c|}{\textbf{0.206}} & \multicolumn{1}{c|}{0.971} & \multicolumn{1}{c|}{-} & \multicolumn{1}{c|}{-} & \multicolumn{1}{c|}{0.815} & \multicolumn{1}{c|}{0.811} & - \\ \hline
time bin & bin 1 & \multicolumn{1}{c|}{0.564} & \multicolumn{1}{c|}{0.206} & \multicolumn{1}{c|}{0.971} & \multicolumn{1}{c|}{-} & \multicolumn{1}{c|}{-} & \multicolumn{1}{c|}{\textbf{0.565}} & \multicolumn{1}{c|}{\textbf{0.206}} & \multicolumn{1}{c|}{0.971} & \multicolumn{1}{c|}{-} & \multicolumn{1}{c|}{-} & \multicolumn{1}{c|}{0.815} & \multicolumn{1}{c|}{0.811} & - \\ \hline
\end{tabular}
    }
\end{table}

\begin{table}[!ht]
    \centering
    \captionsetup{labelformat=default,labelsep=colon,name=Supplementary Table}
    \caption{\label{result_KIRC} Table of uncertainty and fairness evaluation for \textbf{TCGA-KIRC} dataset. p-values are computed using the two-sided Wilcoxon rank-sum test to compare the grades and survival risks predicted using real and synthesized transcriptomic data, in addition to WSIs. \munc, \mcov, \cunc and \ccov refer to marginal uncertainty, marginal coverage, conditional uncertainty and conditional coverage, respectively. \textuparrow \ indicates the larger the better, and \textdownarrow \ indicates the smaller the better. Equivalent or close ($\pm 0.01$) performance using real and synthesized data is marked in \textbf{bold}. \\}
    \resizebox{\textwidth}{!}{
\begin{tabular}{|r|l|ccccccccccccc|}
\hline
\multicolumn{1}{|c|}{\multirow{3}{*}{Category}} & \multicolumn{1}{c|}{\multirow{3}{*}{Group}} & \multicolumn{13}{c|}{Gradation} \\ \cline{3-15} 
\multicolumn{1}{|c|}{} & \multicolumn{1}{c|}{} & \multicolumn{5}{c|}{Real} & \multicolumn{5}{c|}{Synthesized} & \multicolumn{3}{c|}{p-value \textuparrow} \\ \cline{3-15} 
\multicolumn{1}{|c|}{} & \multicolumn{1}{c|}{} & \multicolumn{1}{c|}{AUC \textuparrow} & \multicolumn{1}{c|}{\munc \textdownarrow} & \multicolumn{1}{c|}{\mcov \textuparrow} & \multicolumn{1}{c|}{\cunc \textdownarrow} & \multicolumn{1}{c|}{\ccov \textuparrow} & \multicolumn{1}{c|}{AUC \textuparrow} & \multicolumn{1}{c|}{\munc \textdownarrow} & \multicolumn{1}{c|}{\mcov \textuparrow} & \multicolumn{1}{c|}{\cunc \textdownarrow} & \multicolumn{1}{c|}{\ccov \textuparrow} & \multicolumn{1}{c|}{grade} & \multicolumn{1}{c|}{\munc} & \cunc \\ \hline
overall & - & \multicolumn{1}{c|}{0.778} & \multicolumn{1}{c|}{0.292} & \multicolumn{1}{c|}{0.922} & \multicolumn{1}{c|}{-} & \multicolumn{1}{c|}{-} & \multicolumn{1}{c|}{\textbf{0.773}} & \multicolumn{1}{c|}{0.334} & \multicolumn{1}{c|}{0.948} & \multicolumn{1}{c|}{-} & \multicolumn{1}{c|}{-} & \multicolumn{1}{c|}{0.850} & \multicolumn{1}{c|}{0.068} & - \\ \hline
\multirow{2}{*}{gender} & male & \multicolumn{1}{c|}{0.726} & \multicolumn{1}{c|}{0.303} & \multicolumn{1}{c|}{0.909} & \multicolumn{1}{c|}{\multirow{2}{*}{0.290}} & \multicolumn{1}{c|}{\multirow{2}{*}{0.936}} & \multicolumn{1}{c|}{\textbf{0.721}} & \multicolumn{1}{c|}{0.337} & \multicolumn{1}{c|}{0.929} & \multicolumn{1}{c|}{\multirow{2}{*}{0.337}} & \multicolumn{1}{c|}{\multirow{2}{*}{0.955}} & \multicolumn{1}{c|}{0.905} & \multicolumn{1}{c|}{0.246} & \multirow{2}{*}{0.184} \\ \cline{2-5} \cline{8-10} \cline{13-14}
 & female & \multicolumn{1}{c|}{0.864} & \multicolumn{1}{c|}{0.278} & \multicolumn{1}{c|}{0.963} & \multicolumn{1}{c|}{} & \multicolumn{1}{c|}{} & \multicolumn{1}{c|}{\textbf{0.864}} & \multicolumn{1}{c|}{0.336} & \multicolumn{1}{c|}{0.981} & \multicolumn{1}{c|}{} & \multicolumn{1}{c|}{} & \multicolumn{1}{c|}{0.878} & \multicolumn{1}{c|}{0.123} &  \\ \hline
\multirow{3}{*}{age} & \textless 40 & \multicolumn{1}{c|}{0.863} & \multicolumn{1}{c|}{0.056} & \multicolumn{1}{c|}{1.000} & \multicolumn{1}{c|}{\multirow{3}{*}{0.248}} & \multicolumn{1}{c|}{\multirow{3}{*}{0.952}} & \multicolumn{1}{c|}{0.852} & \multicolumn{1}{c|}{0.111} & \multicolumn{1}{c|}{1.000} & \multicolumn{1}{c|}{\multirow{3}{*}{0.273}} & \multicolumn{1}{c|}{\multirow{3}{*}{0.953}} & \multicolumn{1}{c|}{0.513} & \multicolumn{1}{c|}{0.513} & \multirow{3}{*}{0.643} \\ \cline{2-5} \cline{8-10} \cline{13-14}
 & 40 - 60 & \multicolumn{1}{c|}{0.780} & \multicolumn{1}{c|}{0.393} & \multicolumn{1}{c|}{0.976} & \multicolumn{1}{c|}{} & \multicolumn{1}{c|}{} & \multicolumn{1}{c|}{\textbf{0.778}} & \multicolumn{1}{c|}{\textbf{0.389}} & \multicolumn{1}{c|}{0.964} & \multicolumn{1}{c|}{} & \multicolumn{1}{c|}{} & \multicolumn{1}{c|}{1.000} & \multicolumn{1}{c|}{0.894} &  \\ \cline{2-5} \cline{8-10} \cline{13-14}
 & \textgreater 60 & \multicolumn{1}{c|}{0.770} & \multicolumn{1}{c|}{0.295} & \multicolumn{1}{c|}{0.879} & \multicolumn{1}{c|}{} & \multicolumn{1}{c|}{} & \multicolumn{1}{c|}{\textbf{0.761}} & \multicolumn{1}{c|}{0.318} & \multicolumn{1}{c|}{0.894} & \multicolumn{1}{c|}{} & \multicolumn{1}{c|}{} & \multicolumn{1}{c|}{0.886} & \multicolumn{1}{c|}{0.521} &  \\ \hline
magni- & level1 & \multicolumn{1}{c|}{0.784} & \multicolumn{1}{c|}{0.369} & \multicolumn{1}{c|}{0.980} & \multicolumn{1}{c|}{\multirow{3}{*}{0.341}} & \multicolumn{1}{c|}{\multirow{3}{*}{0.961}} & \multicolumn{1}{c|}{\textbf{0.780}} & \multicolumn{1}{c|}{\textbf{0.337}} & \multicolumn{1}{c|}{0.961} & \multicolumn{1}{c|}{\multirow{3}{*}{0.358}} & \multicolumn{1}{c|}{\multirow{3}{*}{0.961}} & \multicolumn{1}{c|}{1.000} & \multicolumn{1}{c|}{0.393} & \multirow{3}{*}{0.338} \\ \cline{2-5} \cline{8-10} \cline{13-14}
fication & level2 & \multicolumn{1}{c|}{0.780} & \multicolumn{1}{c|}{0.317} & \multicolumn{1}{c|}{0.961} & \multicolumn{1}{c|}{} & \multicolumn{1}{c|}{} & \multicolumn{1}{c|}{\textbf{0.774}} & \multicolumn{1}{c|}{0.376} & \multicolumn{1}{c|}{0.980} & \multicolumn{1}{c|}{} & \multicolumn{1}{c|}{} & \multicolumn{1}{c|}{0.867} & \multicolumn{1}{c|}{0.125} &  \\ \cline{2-5} \cline{8-10} \cline{13-14}
\multicolumn{1}{|l|}{} & level3 & \multicolumn{1}{c|}{0.771} & \multicolumn{1}{c|}{0.337} & \multicolumn{1}{c|}{0.941} & \multicolumn{1}{c|}{} & \multicolumn{1}{c|}{} & \multicolumn{1}{c|}{\textbf{0.769}} & \multicolumn{1}{c|}{0.363} & \multicolumn{1}{c|}{0.941} & \multicolumn{1}{c|}{} & \multicolumn{1}{c|}{} & \multicolumn{1}{c|}{0.867} & \multicolumn{1}{c|}{0.495} &  \\ \hline
vital & alive & \multicolumn{1}{c|}{0.826} & \multicolumn{1}{c|}{0.239} & \multicolumn{1}{c|}{0.931} & \multicolumn{1}{c|}{\multirow{2}{*}{0.320}} & \multicolumn{1}{c|}{\multirow{2}{*}{0.926}} & \multicolumn{1}{c|}{\textbf{0.822}} & \multicolumn{1}{c|}{0.288} & \multicolumn{1}{c|}{0.951} & \multicolumn{1}{c|}{\multirow{2}{*}{0.358}} & \multicolumn{1}{c|}{\multirow{2}{*}{0.946}} & \multicolumn{1}{c|}{0.905} & \multicolumn{1}{c|}{0.076} & \multirow{2}{*}{0.286} \\ \cline{2-5} \cline{8-10} \cline{13-14}
status & dead & \multicolumn{1}{c|}{0.661} & \multicolumn{1}{c|}{0.402} & \multicolumn{1}{c|}{0.922} & \multicolumn{1}{c|}{} & \multicolumn{1}{c|}{} & \multicolumn{1}{c|}{\textbf{0.656}} & \multicolumn{1}{c|}{0.428} & \multicolumn{1}{c|}{0.941} & \multicolumn{1}{c|}{} & \multicolumn{1}{c|}{} & \multicolumn{1}{c|}{0.880} & \multicolumn{1}{c|}{0.495} &  \\ \hline
\multirow{4}{*}{grade} & grade I & \multicolumn{1}{c|}{0.222} & \multicolumn{1}{c|}{0.500} & \multicolumn{1}{c|}{0.667} & \multicolumn{1}{c|}{\multirow{4}{*}{0.356}} & \multicolumn{1}{c|}{\multirow{4}{*}{0.859}} & \multicolumn{1}{c|}{0.111} & \multicolumn{1}{c|}{\textbf{0.500}} & \multicolumn{1}{c|}{0.333} & \multicolumn{1}{c|}{\multirow{4}{*}{\textbf{0.349}}} & \multicolumn{1}{c|}{\multirow{4}{*}{0.779}} & \multicolumn{1}{c|}{1.000} & \multicolumn{1}{c|}{1.000} & \multirow{4}{*}{0.625} \\ \cline{2-5} \cline{8-10} \cline{13-14}
 & grade II & \multicolumn{1}{c|}{0.987} & \multicolumn{1}{c|}{0.016} & \multicolumn{1}{c|}{0.905} & \multicolumn{1}{c|}{} & \multicolumn{1}{c|}{} & \multicolumn{1}{c|}{\textbf{0.978}} & \multicolumn{1}{c|}{\textbf{0.000}} & \multicolumn{1}{c|}{0.873} & \multicolumn{1}{c|}{} & \multicolumn{1}{c|}{} & \multicolumn{1}{c|}{0.880} & \multicolumn{1}{c|}{0.878} &  \\ \cline{2-5} \cline{8-10} \cline{13-14}
 & grade III & \multicolumn{1}{c|}{0.647} & \multicolumn{1}{c|}{0.283} & \multicolumn{1}{c|}{0.905} & \multicolumn{1}{c|}{} & \multicolumn{1}{c|}{} & \multicolumn{1}{c|}{\textbf{0.658}} & \multicolumn{1}{c|}{0.333} & \multicolumn{1}{c|}{0.952} & \multicolumn{1}{c|}{} & \multicolumn{1}{c|}{} & \multicolumn{1}{c|}{0.645} & \multicolumn{1}{c|}{0.164} &  \\ \cline{2-5} \cline{8-10} \cline{13-14}
 & grade IV & \multicolumn{1}{c|}{0.504} & \multicolumn{1}{c|}{0.625} & \multicolumn{1}{c|}{0.958} & \multicolumn{1}{c|}{} & \multicolumn{1}{c|}{} & \multicolumn{1}{c|}{\textbf{0.499}} & \multicolumn{1}{c|}{\textbf{0.563}} & \multicolumn{1}{c|}{0.958} & \multicolumn{1}{c|}{} & \multicolumn{1}{c|}{} & \multicolumn{1}{c|}{0.711} & \multicolumn{1}{c|}{0.458} &  \\ \hline
\multirow{4}{*}{time bin} & bin 1 & \multicolumn{1}{c|}{0.718} & \multicolumn{1}{c|}{0.366} & \multicolumn{1}{c|}{0.944} & \multicolumn{1}{c|}{\multirow{4}{*}{0.247}} & \multicolumn{1}{c|}{\multirow{4}{*}{0.874}} & \multicolumn{1}{c|}{\textbf{0.713}} & \multicolumn{1}{c|}{0.384} & \multicolumn{1}{c|}{0.944} & \multicolumn{1}{c|}{\multirow{4}{*}{0.279}} & \multicolumn{1}{c|}{\multirow{4}{*}{0.961}} & \multicolumn{1}{c|}{0.889} & \multicolumn{1}{c|}{0.565} & \multirow{4}{*}{0.552} \\ \cline{2-5} \cline{8-10} \cline{13-14}
 & bin 2 & \multicolumn{1}{c|}{0.837} & \multicolumn{1}{c|}{0.214} & \multicolumn{1}{c|}{0.883} & \multicolumn{1}{c|}{} & \multicolumn{1}{c|}{} & \multicolumn{1}{c|}{\textbf{0.828}} & \multicolumn{1}{c|}{\textbf{0.233}} & \multicolumn{1}{c|}{0.900} & \multicolumn{1}{c|}{} & \multicolumn{1}{c|}{} & \multicolumn{1}{c|}{0.885} & \multicolumn{1}{c|}{0.560} &  \\ \cline{2-5} \cline{8-10} \cline{13-14}
 & bin 3 & \multicolumn{1}{c|}{0.848} & \multicolumn{1}{c|}{0.296} & \multicolumn{1}{c|}{1.000} & \multicolumn{1}{c|}{} & \multicolumn{1}{c|}{} & \multicolumn{1}{c|}{0.835} & \multicolumn{1}{c|}{0.333} & \multicolumn{1}{c|}{1.000} & \multicolumn{1}{c|}{} & \multicolumn{1}{c|}{} & \multicolumn{1}{c|}{1.000} & \multicolumn{1}{c|}{0.569} &  \\ \cline{2-5} \cline{8-10} \cline{13-14}
 & bin 4 & \multicolumn{1}{c|}{0.667} & \multicolumn{1}{c|}{0.111} & \multicolumn{1}{c|}{0.667} & \multicolumn{1}{c|}{} & \multicolumn{1}{c|}{} & \multicolumn{1}{c|}{\textbf{0.667}} & \multicolumn{1}{c|}{0.167} & \multicolumn{1}{c|}{1.000} & \multicolumn{1}{c|}{} & \multicolumn{1}{c|}{} & \multicolumn{1}{c|}{1.000} & \multicolumn{1}{c|}{0.513} &  \\ \hline \hline
\multicolumn{1}{|c|}{\multirow{3}{*}{Category}} & \multicolumn{1}{c|}{\multirow{3}{*}{Group}} & \multicolumn{13}{c|}{Risk Estimation} \\ \cline{3-15} 
\multicolumn{1}{|c|}{} & \multicolumn{1}{c|}{} & \multicolumn{5}{c|}{Real} & \multicolumn{5}{c|}{Synthesized} & \multicolumn{3}{c|}{p-value \textuparrow} \\ \cline{3-15} 
\multicolumn{1}{|c|}{} & \multicolumn{1}{c|}{} & \multicolumn{1}{c|}{C Index \textuparrow} & \multicolumn{1}{c|}{\munc \textdownarrow} & \multicolumn{1}{c|}{\mcov \textuparrow} & \multicolumn{1}{c|}{\cunc \textdownarrow} & \multicolumn{1}{c|}{\ccov \textuparrow} & \multicolumn{1}{c|}{C Index \textuparrow} & \multicolumn{1}{c|}{\munc \textdownarrow} & \multicolumn{1}{c|}{\mcov \textuparrow} & \multicolumn{1}{c|}{\cunc \textdownarrow} & \multicolumn{1}{c|}{\ccov \textuparrow} & \multicolumn{1}{c|}{risk} & \multicolumn{1}{c|}{\munc} & \cunc \\ \hline
overall & - & \multicolumn{1}{c|}{0.697} & \multicolumn{1}{c|}{0.898} & \multicolumn{1}{c|}{0.882} & \multicolumn{1}{c|}{-} & \multicolumn{1}{c|}{-} & \multicolumn{1}{c|}{0.681} & \multicolumn{1}{c|}{\textbf{0.909}} & \multicolumn{1}{c|}{0.928} & \multicolumn{1}{c|}{-} & \multicolumn{1}{c|}{-} & \multicolumn{1}{c|}{0.300} & \multicolumn{1}{c|}{0.307} & - \\ \hline
\multirow{2}{*}{gender} & male & \multicolumn{1}{c|}{0.705} & \multicolumn{1}{c|}{0.909} & \multicolumn{1}{c|}{0.909} & \multicolumn{1}{c|}{\multirow{2}{*}{0.882}} & \multicolumn{1}{c|}{\multirow{2}{*}{0.834}} & \multicolumn{1}{c|}{0.652} & \multicolumn{1}{c|}{0.921} & \multicolumn{1}{c|}{0.960} & \multicolumn{1}{c|}{\multirow{2}{*}{0.893}} & \multicolumn{1}{c|}{\multirow{2}{*}{0.878}} & \multicolumn{1}{c|}{0.669} & \multicolumn{1}{c|}{0.654} & \multirow{2}{*}{0.483} \\ \cline{2-5} \cline{8-10} \cline{13-14}
 & female & \multicolumn{1}{c|}{0.710} & \multicolumn{1}{c|}{0.855} & \multicolumn{1}{c|}{0.759} & \multicolumn{1}{c|}{} & \multicolumn{1}{c|}{} & \multicolumn{1}{c|}{\textbf{0.771}} & \multicolumn{1}{c|}{\textbf{0.865}} & \multicolumn{1}{c|}{0.796} & \multicolumn{1}{c|}{} & \multicolumn{1}{c|}{} & \multicolumn{1}{c|}{0.219} & \multicolumn{1}{c|}{0.311} &  \\ \hline
\multirow{3}{*}{age} & \textless 40 & \multicolumn{1}{c|}{-} & \multicolumn{1}{c|}{0.955} & \multicolumn{1}{c|}{1.000} & \multicolumn{1}{c|}{\multirow{3}{*}{0.919}} & \multicolumn{1}{c|}{\multirow{3}{*}{0.934}} & \multicolumn{1}{c|}{-} & \multicolumn{1}{c|}{\textbf{0.942}} & \multicolumn{1}{c|}{1.000} & \multicolumn{1}{c|}{\multirow{3}{*}{\textbf{0.923}}} & \multicolumn{1}{c|}{\multirow{3}{*}{0.971}} & \multicolumn{1}{c|}{0.275} & \multicolumn{1}{c|}{0.275} & \multirow{3}{*}{0.440} \\ \cline{2-5} \cline{8-10} \cline{13-14}
 & 40 - 60 & \multicolumn{1}{c|}{0.612} & \multicolumn{1}{c|}{0.936} & \multicolumn{1}{c|}{0.893} & \multicolumn{1}{c|}{} & \multicolumn{1}{c|}{} & \multicolumn{1}{c|}{0.591} & \multicolumn{1}{c|}{\textbf{0.935}} & \multicolumn{1}{c|}{0.929} & \multicolumn{1}{c|}{} & \multicolumn{1}{c|}{} & \multicolumn{1}{c|}{0.266} & \multicolumn{1}{c|}{0.265} &  \\ \cline{2-5} \cline{8-10} \cline{13-14}
 & \textgreater 60 & \multicolumn{1}{c|}{0.672} & \multicolumn{1}{c|}{0.864} & \multicolumn{1}{c|}{0.909} & \multicolumn{1}{c|}{} & \multicolumn{1}{c|}{} & \multicolumn{1}{c|}{\textbf{0.681}} & \multicolumn{1}{c|}{0.891} & \multicolumn{1}{c|}{0.985} & \multicolumn{1}{c|}{} & \multicolumn{1}{c|}{} & \multicolumn{1}{c|}{0.781} & \multicolumn{1}{c|}{0.781} &  \\ \hline
magni- & level1 & \multicolumn{1}{c|}{0.706} & \multicolumn{1}{c|}{0.932} & \multicolumn{1}{c|}{0.961} & \multicolumn{1}{c|}{\multirow{3}{*}{0.903}} & \multicolumn{1}{c|}{\multirow{3}{*}{0.908}} & \multicolumn{1}{c|}{0.676} & \multicolumn{1}{c|}{0.947} & \multicolumn{1}{c|}{1.000} & \multicolumn{1}{c|}{\multirow{3}{*}{\textbf{0.915}}} & \multicolumn{1}{c|}{\multirow{3}{*}{0.948}} & \multicolumn{1}{c|}{0.563} & \multicolumn{1}{c|}{0.588} & \multirow{3}{*}{0.452} \\ \cline{2-5} \cline{8-10} \cline{13-14}
fication & level2 & \multicolumn{1}{c|}{0.734} & \multicolumn{1}{c|}{0.899} & \multicolumn{1}{c|}{0.902} & \multicolumn{1}{c|}{} & \multicolumn{1}{c|}{} & \multicolumn{1}{c|}{0.702} & \multicolumn{1}{c|}{\textbf{0.908}} & \multicolumn{1}{c|}{0.961} & \multicolumn{1}{c|}{} & \multicolumn{1}{c|}{} & \multicolumn{1}{c|}{0.248} & \multicolumn{1}{c|}{0.277} &  \\ \cline{2-5} \cline{8-10} \cline{13-14}
\multicolumn{1}{|l|}{} & level3 & \multicolumn{1}{c|}{0.694} & \multicolumn{1}{c|}{0.878} & \multicolumn{1}{c|}{0.863} & \multicolumn{1}{c|}{} & \multicolumn{1}{c|}{} & \multicolumn{1}{c|}{0.676} & \multicolumn{1}{c|}{0.890} & \multicolumn{1}{c|}{0.882} & \multicolumn{1}{c|}{} & \multicolumn{1}{c|}{} & \multicolumn{1}{c|}{0.408} & \multicolumn{1}{c|}{0.493} &  \\ \hline
vital & alive & \multicolumn{1}{c|}{-} & \multicolumn{1}{c|}{0.934} & \multicolumn{1}{c|}{0.912} & \multicolumn{1}{c|}{\multirow{2}{*}{0.889}} & \multicolumn{1}{c|}{\multirow{2}{*}{0.907}} & \multicolumn{1}{c|}{-} & \multicolumn{1}{c|}{\textbf{0.934}} & \multicolumn{1}{c|}{0.941} & \multicolumn{1}{c|}{\multirow{2}{*}{0.904}} & \multicolumn{1}{c|}{\multirow{2}{*}{0.961}} & \multicolumn{1}{c|}{0.083} & \multicolumn{1}{c|}{0.089} & \multirow{2}{*}{0.343} \\ \cline{2-5} \cline{8-10} \cline{13-14}
status & dead & \multicolumn{1}{c|}{0.487} & \multicolumn{1}{c|}{0.845} & \multicolumn{1}{c|}{0.902} & \multicolumn{1}{c|}{} & \multicolumn{1}{c|}{} & \multicolumn{1}{c|}{\textbf{0.499}} & \multicolumn{1}{c|}{0.874} & \multicolumn{1}{c|}{0.980} & \multicolumn{1}{c|}{} & \multicolumn{1}{c|}{} & \multicolumn{1}{c|}{0.595} & \multicolumn{1}{c|}{0.597} &  \\ \hline
\multirow{4}{*}{grade} & grade I & \multicolumn{1}{c|}{-} & \multicolumn{1}{c|}{0.976} & \multicolumn{1}{c|}{\textbf{0.667}} & \multicolumn{1}{c|}{\multirow{4}{*}{0.927}} & \multicolumn{1}{c|}{\multirow{4}{*}{0.847}} & \multicolumn{1}{c|}{-} & \multicolumn{1}{c|}{\textbf{0.969}} & \multicolumn{1}{c|}{1.000} & \multicolumn{1}{c|}{\multirow{4}{*}{0.939}} & \multicolumn{1}{c|}{\multirow{4}{*}{0.948}} & \multicolumn{1}{c|}{0.275} & \multicolumn{1}{c|}{0.513} & \multirow{4}{*}{0.358} \\ \cline{2-5} \cline{8-10} \cline{13-14}
 & grade II & \multicolumn{1}{c|}{0.835} & \multicolumn{1}{c|}{0.926} & \multicolumn{1}{c|}{0.905} & \multicolumn{1}{c|}{} & \multicolumn{1}{c|}{} & \multicolumn{1}{c|}{\textbf{0.841}} & \multicolumn{1}{c|}{\textbf{0.919}} & \multicolumn{1}{c|}{0.937} & \multicolumn{1}{c|}{} & \multicolumn{1}{c|}{} & \multicolumn{1}{c|}{0.139} & \multicolumn{1}{c|}{0.149} &  \\ \cline{2-5} \cline{8-10} \cline{13-14}
 & grade III & \multicolumn{1}{c|}{0.612} & \multicolumn{1}{c|}{0.906} & \multicolumn{1}{c|}{0.857} & \multicolumn{1}{c|}{} & \multicolumn{1}{c|}{} & \multicolumn{1}{c|}{0.573} & \multicolumn{1}{c|}{\textbf{0.904}} & \multicolumn{1}{c|}{0.857} & \multicolumn{1}{c|}{} & \multicolumn{1}{c|}{} & \multicolumn{1}{c|}{0.550} & \multicolumn{1}{c|}{0.659} &  \\ \cline{2-5} \cline{8-10} \cline{13-14}
 & grade IV & \multicolumn{1}{c|}{0.481} & \multicolumn{1}{c|}{0.901} & \multicolumn{1}{c|}{0.958} & \multicolumn{1}{c|}{} & \multicolumn{1}{c|}{} & \multicolumn{1}{c|}{0.439} & \multicolumn{1}{c|}{0.964} & \multicolumn{1}{c|}{1.000} & \multicolumn{1}{c|}{} & \multicolumn{1}{c|}{} & \multicolumn{1}{c|}{0.143} & \multicolumn{1}{c|}{0.112} &  \\ \hline
\multirow{4}{*}{time bin} & bin 1 & \multicolumn{1}{c|}{0.690} & \multicolumn{1}{c|}{0.922} & \multicolumn{1}{c|}{0.861} & \multicolumn{1}{c|}{\multirow{4}{*}{0.738}} & \multicolumn{1}{c|}{\multirow{4}{*}{0.961}} & \multicolumn{1}{c|}{\textbf{0.714}} & \multicolumn{1}{c|}{0.938} & \multicolumn{1}{c|}{0.944} & \multicolumn{1}{c|}{\multirow{4}{*}{\textbf{0.746}}} & \multicolumn{1}{c|}{\multirow{4}{*}{0.982}} & \multicolumn{1}{c|}{0.441} & \multicolumn{1}{c|}{0.474} & \multirow{4}{*}{0.744} \\ \cline{2-5} \cline{8-10} \cline{13-14}
 & bin 2 & \multicolumn{1}{c|}{0.842} & \multicolumn{1}{c|}{0.864} & \multicolumn{1}{c|}{0.983} & \multicolumn{1}{c|}{} & \multicolumn{1}{c|}{} & \multicolumn{1}{c|}{\textbf{0.836}} & \multicolumn{1}{c|}{0.881} & \multicolumn{1}{c|}{0.983} & \multicolumn{1}{c|}{} & \multicolumn{1}{c|}{} & \multicolumn{1}{c|}{0.557} & \multicolumn{1}{c|}{0.502} &  \\ \cline{2-5} \cline{8-10} \cline{13-14}
 & bin 3 & \multicolumn{1}{c|}{-} & \multicolumn{1}{c|}{0.389} & \multicolumn{1}{c|}{1.000} & \multicolumn{1}{c|}{} & \multicolumn{1}{c|}{} & \multicolumn{1}{c|}{-} & \multicolumn{1}{c|}{\textbf{0.389}} & \multicolumn{1}{c|}{1.000} & \multicolumn{1}{c|}{} & \multicolumn{1}{c|}{} & \multicolumn{1}{c|}{0.411} & \multicolumn{1}{c|}{1.000} &  \\ \cline{2-5} \cline{8-10} \cline{13-14}
 & bin 4 & \multicolumn{1}{c|}{-} & \multicolumn{1}{c|}{0.775} & \multicolumn{1}{c|}{1.000} & \multicolumn{1}{c|}{} & \multicolumn{1}{c|}{} & \multicolumn{1}{c|}{-} & \multicolumn{1}{c|}{\textbf{0.775}} & \multicolumn{1}{c|}{1.000} & \multicolumn{1}{c|}{} & \multicolumn{1}{c|}{} & \multicolumn{1}{c|}{0.275} & \multicolumn{1}{c|}{1.000} &  \\ \hline
\end{tabular}
}
\end{table}

\begin{table}[!ht]
    \centering
    \captionsetup{labelformat=default,labelsep=colon,name=Supplementary Table}
    \caption{\label{result_TUCEC} Table of uncertainty and fairness evaluation for \textbf{TCGA-UCEC} dataset. p-values are computed using the two-sided Wilcoxon rank-sum test to compare the grades and survival risks predicted using real and synthesized transcriptomic data, in addition to WSIs. \munc, \mcov, \cunc and \ccov refer to marginal uncertainty, marginal coverage, conditional uncertainty and conditional coverage, respectively. \textuparrow \ indicates the larger the better, and \textdownarrow \ indicates the smaller the better. Equivalent or close ($\pm 0.01$) performance using real and synthesized data is marked in \textbf{bold}. \\}
    \resizebox{\textwidth}{!}{
\begin{tabular}{|r|l|ccccccccccccc|}
\hline
\multicolumn{1}{|c|}{\multirow{3}{*}{Category}} & \multicolumn{1}{c|}{\multirow{3}{*}{Group}} & \multicolumn{13}{c|}{Gradation} \\ \cline{3-15} 
\multicolumn{1}{|c|}{} & \multicolumn{1}{c|}{} & \multicolumn{5}{c|}{Real} & \multicolumn{5}{c|}{Synthesized} & \multicolumn{3}{c|}{p-value \textuparrow} \\ \cline{3-15} 
\multicolumn{1}{|c|}{} & \multicolumn{1}{c|}{} & \multicolumn{1}{c|}{AUC \textuparrow} & \multicolumn{1}{c|}{\munc \textdownarrow} & \multicolumn{1}{c|}{\mcov \textuparrow} & \multicolumn{1}{c|}{\cunc \textdownarrow} & \multicolumn{1}{c|}{\ccov \textuparrow} & \multicolumn{1}{c|}{AUC \textuparrow} & \multicolumn{1}{c|}{\munc \textdownarrow} & \multicolumn{1}{c|}{\mcov \textuparrow} & \multicolumn{1}{c|}{\cunc \textdownarrow} & \multicolumn{1}{c|}{\ccov \textuparrow} & \multicolumn{1}{c|}{grade} & \multicolumn{1}{c|}{\munc} & \cunc \\ \hline
overall & - & \multicolumn{1}{c|}{0.828} & \multicolumn{1}{c|}{0.337} & \multicolumn{1}{c|}{0.854} & \multicolumn{1}{c|}{-} & \multicolumn{1}{c|}{-} & \multicolumn{1}{c|}{\textbf{0.821}} & \multicolumn{1}{c|}{\textbf{0.337}} & \multicolumn{1}{c|}{0.854} & \multicolumn{1}{c|}{-} & \multicolumn{1}{c|}{-} & \multicolumn{1}{c|}{0.941} & \multicolumn{1}{c|}{1.000} & - \\ \hline
\multirow{2}{*}{age} & 40 - 60 & \multicolumn{1}{c|}{0.739} & \multicolumn{1}{c|}{0.744} & \multicolumn{1}{c|}{0.949} & \multicolumn{1}{c|}{\multirow{2}{*}{0.489}} & \multicolumn{1}{c|}{\multirow{2}{*}{0.916}} & \multicolumn{1}{c|}{\textbf{0.739}} & \multicolumn{1}{c|}{\textbf{0.744}} & \multicolumn{1}{c|}{0.949} & \multicolumn{1}{c|}{\multirow{2}{*}{\textbf{0.489}}} & \multicolumn{1}{c|}{\multirow{2}{*}{0.916}} & \multicolumn{1}{c|}{1.000} & \multicolumn{1}{c|}{1.000} & \multirow{2}{*}{1.000} \\ \cline{2-5} \cline{8-10} \cline{13-14}
 & \textgreater{}60 & \multicolumn{1}{c|}{0.885} & \multicolumn{1}{c|}{0.235} & \multicolumn{1}{c|}{0.882} & \multicolumn{1}{c|}{} & \multicolumn{1}{c|}{} & \multicolumn{1}{c|}{0.874} & \multicolumn{1}{c|}{\textbf{0.235}} & \multicolumn{1}{c|}{0.882} & \multicolumn{1}{c|}{} & \multicolumn{1}{c|}{} & \multicolumn{1}{c|}{0.912} & \multicolumn{1}{c|}{1.000} &  \\ \hline
magni- & level1 & \multicolumn{1}{c|}{0.881} & \multicolumn{1}{c|}{0.567} & \multicolumn{1}{c|}{0.933} & \multicolumn{1}{c|}{\multirow{3}{*}{0.519}} & \multicolumn{1}{c|}{\multirow{3}{*}{0.915}} & \multicolumn{1}{c|}{0.870} & \multicolumn{1}{c|}{0.544} & \multicolumn{1}{c|}{0.933} & \multicolumn{1}{c|}{\multirow{3}{*}{\textbf{0.511}}} & \multicolumn{1}{c|}{\multirow{3}{*}{0.915}} & \multicolumn{1}{c|}{0.912} & \multicolumn{1}{c|}{0.877} & \multirow{3}{*}{0.959} \\ \cline{2-5} \cline{8-10} \cline{13-14}
fication & level2 & \multicolumn{1}{c|}{0.810} & \multicolumn{1}{c|}{0.278} & \multicolumn{1}{c|}{0.812} & \multicolumn{1}{c|}{} & \multicolumn{1}{c|}{} & \multicolumn{1}{c|}{\textbf{0.810}} & \multicolumn{1}{c|}{\textbf{0.278}} & \multicolumn{1}{c|}{0.812} & \multicolumn{1}{c|}{} & \multicolumn{1}{c|}{} & \multicolumn{1}{c|}{1.000} & \multicolumn{1}{c|}{1.000} &  \\ \cline{2-5} \cline{8-10} \cline{13-14}
\multicolumn{1}{|l|}{} & level3 & \multicolumn{1}{c|}{0.775} & \multicolumn{1}{c|}{0.711} & \multicolumn{1}{c|}{1.000} & \multicolumn{1}{c|}{} & \multicolumn{1}{c|}{} & \multicolumn{1}{c|}{\textbf{0.775}} & \multicolumn{1}{c|}{\textbf{0.711}} & \multicolumn{1}{c|}{1.000} & \multicolumn{1}{c|}{} & \multicolumn{1}{c|}{} & \multicolumn{1}{c|}{1.000} & \multicolumn{1}{c|}{1.000} &  \\ \hline
vital & alive & \multicolumn{1}{c|}{0.811} & \multicolumn{1}{c|}{0.350} & \multicolumn{1}{c|}{0.866} & \multicolumn{1}{c|}{\multirow{2}{*}{0.300}} & \multicolumn{1}{c|}{\multirow{2}{*}{0.850}} & \multicolumn{1}{c|}{\textbf{0.803}} & \multicolumn{1}{c|}{\textbf{0.350}} & \multicolumn{1}{c|}{0.866} & \multicolumn{1}{c|}{\multirow{2}{*}{\textbf{0.300}}} & \multicolumn{1}{c|}{\multirow{2}{*}{0.850}} & \multicolumn{1}{c|}{0.941} & \multicolumn{1}{c|}{1.000} & \multirow{2}{*}{1.000} \\ \cline{2-5} \cline{8-10} \cline{13-14}
status & dead & \multicolumn{1}{c|}{0.917} & \multicolumn{1}{c|}{0.250} & \multicolumn{1}{c|}{0.833} & \multicolumn{1}{c|}{} & \multicolumn{1}{c|}{} & \multicolumn{1}{c|}{\textbf{0.917}} & \multicolumn{1}{c|}{\textbf{0.250}} & \multicolumn{1}{c|}{0.833} & \multicolumn{1}{c|}{} & \multicolumn{1}{c|}{} & \multicolumn{1}{c|}{1.000} & \multicolumn{1}{c|}{1.000} &  \\ \hline
\multirow{3}{*}{grade} & grade I & \multicolumn{1}{c|}{0.971} & \multicolumn{1}{c|}{1.000} & \multicolumn{1}{c|}{1.000} & \multicolumn{1}{c|}{\multirow{3}{*}{0.651}} & \multicolumn{1}{c|}{\multirow{3}{*}{0.984}} & \multicolumn{1}{c|}{\textbf{0.971}} & \multicolumn{1}{c|}{\textbf{1.000}} & \multicolumn{1}{c|}{1.000} & \multicolumn{1}{c|}{\multirow{3}{*}{\textbf{0.651}}} & \multicolumn{1}{c|}{\multirow{3}{*}{0.984}} & \multicolumn{1}{c|}{1.000} & \multicolumn{1}{c|}{1.000} & \multirow{3}{*}{1.000} \\ \cline{2-5} \cline{8-10} \cline{13-14}
 & grade II & \multicolumn{1}{c|}{0.805} & \multicolumn{1}{c|}{0.556} & \multicolumn{1}{c|}{1.000} & \multicolumn{1}{c|}{} & \multicolumn{1}{c|}{} & \multicolumn{1}{c|}{0.768} & \multicolumn{1}{c|}{\textbf{0.556}} & \multicolumn{1}{c|}{1.000} & \multicolumn{1}{c|}{} & \multicolumn{1}{c|}{} & \multicolumn{1}{c|}{0.893} & \multicolumn{1}{c|}{1.000} &  \\ \cline{2-5} \cline{8-10} \cline{13-14}
 & grade III & \multicolumn{1}{c|}{0.823} & \multicolumn{1}{c|}{0.397} & \multicolumn{1}{c|}{0.952} & \multicolumn{1}{c|}{} & \multicolumn{1}{c|}{} & \multicolumn{1}{c|}{\textbf{0.823}} & \multicolumn{1}{c|}{\textbf{0.397}} & \multicolumn{1}{c|}{0.952} & \multicolumn{1}{c|}{} & \multicolumn{1}{c|}{} & \multicolumn{1}{c|}{1.000} & \multicolumn{1}{c|}{1.000} &  \\ \hline
\multirow{2}{*}{time bin} & bin1 & \multicolumn{1}{c|}{0.761} & \multicolumn{1}{c|}{0.661} & \multicolumn{1}{c|}{0.968} & \multicolumn{1}{c|}{\multirow{2}{*}{0.470}} & \multicolumn{1}{c|}{\multirow{2}{*}{0.984}} & \multicolumn{1}{c|}{\textbf{0.752}} & \multicolumn{1}{c|}{\textbf{0.651}} & \multicolumn{1}{c|}{0.968} & \multicolumn{1}{c|}{\multirow{2}{*}{\textbf{0.464}}} & \multicolumn{1}{c|}{\multirow{2}{*}{0.984}} & \multicolumn{1}{c|}{0.932} & \multicolumn{1}{c|}{0.901} & \multirow{2}{*}{0.950} \\ \cline{2-5} \cline{8-10} \cline{13-14}
 & bin2 & \multicolumn{1}{c|}{0.955} & \multicolumn{1}{c|}{0.278} & \multicolumn{1}{c|}{1.000} & \multicolumn{1}{c|}{} & \multicolumn{1}{c|}{} & \multicolumn{1}{c|}{\textbf{0.955}} & \multicolumn{1}{c|}{\textbf{0.278}} & \multicolumn{1}{c|}{1.000} & \multicolumn{1}{c|}{} & \multicolumn{1}{c|}{} & \multicolumn{1}{c|}{1.000} & \multicolumn{1}{c|}{1.000} &  \\ \hline \hline
\multicolumn{1}{|c|}{\multirow{3}{*}{Category}} & \multicolumn{1}{c|}{\multirow{3}{*}{Group}} & \multicolumn{13}{c|}{Risk Estimation} \\ \cline{3-15} 
\multicolumn{1}{|c|}{} & \multicolumn{1}{c|}{} & \multicolumn{5}{c|}{Real} & \multicolumn{5}{c|}{Synthesized} & \multicolumn{3}{c|}{p-value \textuparrow} \\ \cline{3-15} 
\multicolumn{1}{|c|}{} & \multicolumn{1}{c|}{} & \multicolumn{1}{c|}{C Index \textuparrow} & \multicolumn{1}{c|}{\munc \textdownarrow} & \multicolumn{1}{c|}{\mcov \textuparrow} & \multicolumn{1}{c|}{\cunc \textdownarrow} & \multicolumn{1}{c|}{\ccov \textuparrow} & \multicolumn{1}{c|}{C Index \textuparrow} & \multicolumn{1}{c|}{\munc \textdownarrow} & \multicolumn{1}{c|}{\mcov \textuparrow} & \multicolumn{1}{c|}{\cunc \textdownarrow} & \multicolumn{1}{c|}{\ccov \textuparrow} & \multicolumn{1}{c|}{risk} & \multicolumn{1}{c|}{\munc} & \cunc \\ \hline
overall & - & \multicolumn{1}{c|}{0.680} & \multicolumn{1}{c|}{1.000} & \multicolumn{1}{c|}{0.933} & \multicolumn{1}{c|}{-} & \multicolumn{1}{c|}{-} & \multicolumn{1}{c|}{\textbf{0.673}} & \multicolumn{1}{c|}{\textbf{1.000}} & \multicolumn{1}{c|}{0.933} & \multicolumn{1}{c|}{-} & \multicolumn{1}{c|}{-} & \multicolumn{1}{c|}{0.887} & \multicolumn{1}{c|}{0.992} & - \\ \hline
\multirow{2}{*}{age} & 40 - 60 & \multicolumn{1}{c|}{-} & \multicolumn{1}{c|}{1.000} & \multicolumn{1}{c|}{0.897} & \multicolumn{1}{c|}{\multirow{2}{*}{1.000}} & \multicolumn{1}{c|}{\multirow{2}{*}{0.863}} & \multicolumn{1}{c|}{-} & \multicolumn{1}{c|}{\textbf{1.000}} & \multicolumn{1}{c|}{0.897} & \multicolumn{1}{c|}{\multirow{2}{*}{\textbf{1.000}}} & \multicolumn{1}{c|}{\multirow{2}{*}{0.863}} & \multicolumn{1}{c|}{0.912} & \multicolumn{1}{c|}{0.980} & \multirow{2}{*}{0.986} \\ \cline{2-5} \cline{8-10} \cline{13-14}
 & \textgreater{}60 & \multicolumn{1}{c|}{0.606} & \multicolumn{1}{c|}{0.999} & \multicolumn{1}{c|}{0.829} & \multicolumn{1}{c|}{} & \multicolumn{1}{c|}{} & \multicolumn{1}{c|}{0.592} & \multicolumn{1}{c|}{\textbf{0.999}} & \multicolumn{1}{c|}{0.829} & \multicolumn{1}{c|}{} & \multicolumn{1}{c|}{} & \multicolumn{1}{c|}{0.823} & \multicolumn{1}{c|}{0.992} &  \\ \hline
magni- & level1 & \multicolumn{1}{c|}{0.720} & \multicolumn{1}{c|}{1.000} & \multicolumn{1}{c|}{0.900} & \multicolumn{1}{c|}{\multirow{3}{*}{1.000}} & \multicolumn{1}{c|}{\multirow{3}{*}{0.922}} & \multicolumn{1}{c|}{0.680} & \multicolumn{1}{c|}{\textbf{1.000}} & \multicolumn{1}{c|}{0.900} & \multicolumn{1}{c|}{\multirow{3}{*}{\textbf{1.000}}} & \multicolumn{1}{c|}{\multirow{3}{*}{0.922}} & \multicolumn{1}{c|}{0.767} & \multicolumn{1}{c|}{0.971} & \multirow{3}{*}{0.982} \\ \cline{2-5} \cline{8-10} \cline{13-14}
fication & level2 & \multicolumn{1}{c|}{0.707} & \multicolumn{1}{c|}{1.000} & \multicolumn{1}{c|}{0.933} & \multicolumn{1}{c|}{} & \multicolumn{1}{c|}{} & \multicolumn{1}{c|}{\textbf{0.707}} & \multicolumn{1}{c|}{\textbf{1.000}} & \multicolumn{1}{c|}{0.933} & \multicolumn{1}{c|}{} & \multicolumn{1}{c|}{} & \multicolumn{1}{c|}{0.900} & \multicolumn{1}{c|}{0.982} &  \\ \cline{2-5} \cline{8-10} \cline{13-14}
\multicolumn{1}{|l|}{} & level3 & \multicolumn{1}{c|}{0.547} & \multicolumn{1}{c|}{1.000} & \multicolumn{1}{c|}{0.933} & \multicolumn{1}{c|}{} & \multicolumn{1}{c|}{} & \multicolumn{1}{c|}{\textbf{0.547}} & \multicolumn{1}{c|}{\textbf{1.000}} & \multicolumn{1}{c|}{0.933} & \multicolumn{1}{c|}{} & \multicolumn{1}{c|}{} & \multicolumn{1}{c|}{0.918} & \multicolumn{1}{c|}{0.994} &  \\ \hline
vital & alive & \multicolumn{1}{c|}{-} & \multicolumn{1}{c|}{1.000} & \multicolumn{1}{c|}{0.936} & \multicolumn{1}{c|}{\multirow{2}{*}{0.998}} & \multicolumn{1}{c|}{\multirow{2}{*}{0.968}} & \multicolumn{1}{c|}{-} & \multicolumn{1}{c|}{\textbf{1.000}} & \multicolumn{1}{c|}{0.936} & \multicolumn{1}{c|}{\multirow{2}{*}{\textbf{0.998}}} & \multicolumn{1}{c|}{\multirow{2}{*}{0.968}} & \multicolumn{1}{c|}{0.897} & \multicolumn{1}{c|}{0.992} & \multirow{2}{*}{0.961} \\ \cline{2-5} \cline{8-10} \cline{13-14}
status & dead & \multicolumn{1}{c|}{0.204} & \multicolumn{1}{c|}{0.996} & \multicolumn{1}{c|}{1.000} & \multicolumn{1}{c|}{} & \multicolumn{1}{c|}{} & \multicolumn{1}{c|}{\textbf{0.204}} & \multicolumn{1}{c|}{\textbf{0.996}} & \multicolumn{1}{c|}{1.000} & \multicolumn{1}{c|}{} & \multicolumn{1}{c|}{} & \multicolumn{1}{c|}{0.729} & \multicolumn{1}{c|}{0.931} &  \\ \hline
\multirow{3}{*}{grade} & grade I & \multicolumn{1}{c|}{-} & \multicolumn{1}{c|}{1.000} & \multicolumn{1}{c|}{0.958} & \multicolumn{1}{c|}{\multirow{3}{*}{1.000}} & \multicolumn{1}{c|}{\multirow{3}{*}{0.962}} & \multicolumn{1}{c|}{-} & \multicolumn{1}{c|}{\textbf{1.000}} & \multicolumn{1}{c|}{0.958} & \multicolumn{1}{c|}{\multirow{3}{*}{\textbf{1.000}}} & \multicolumn{1}{c|}{\multirow{3}{*}{0.962}} & \multicolumn{1}{c|}{0.926} & \multicolumn{1}{c|}{0.992} & \multirow{3}{*}{0.991} \\ \cline{2-5} \cline{8-10} \cline{13-14}
 & grade II & \multicolumn{1}{c|}{0.778} & \multicolumn{1}{c|}{1.000} & \multicolumn{1}{c|}{1.000} & \multicolumn{1}{c|}{} & \multicolumn{1}{c|}{} & \multicolumn{1}{c|}{0.746} & \multicolumn{1}{c|}{\textbf{1.000}} & \multicolumn{1}{c|}{1.000} & \multicolumn{1}{c|}{} & \multicolumn{1}{c|}{} & \multicolumn{1}{c|}{0.805} & \multicolumn{1}{c|}{1.000} &  \\ \cline{2-5} \cline{8-10} \cline{13-14}
 & grade III & \multicolumn{1}{c|}{0.503} & \multicolumn{1}{c|}{1.000} & \multicolumn{1}{c|}{0.929} & \multicolumn{1}{c|}{} & \multicolumn{1}{c|}{} & \multicolumn{1}{c|}{\textbf{0.509}} & \multicolumn{1}{c|}{\textbf{1.000}} & \multicolumn{1}{c|}{0.929} & \multicolumn{1}{c|}{} & \multicolumn{1}{c|}{} & \multicolumn{1}{c|}{0.936} & \multicolumn{1}{c|}{0.982} &  \\ \hline
\multirow{2}{*}{time bin} & bin1 & \multicolumn{1}{c|}{0.714} & \multicolumn{1}{c|}{1.000} & \multicolumn{1}{c|}{0.921} & \multicolumn{1}{c|}{\multirow{2}{*}{0.996}} & \multicolumn{1}{c|}{\multirow{2}{*}{0.960}} & \multicolumn{1}{c|}{0.700} & \multicolumn{1}{c|}{\textbf{1.000}} & \multicolumn{1}{c|}{0.921} & \multicolumn{1}{c|}{\multirow{2}{*}{\textbf{0.996}}} & \multicolumn{1}{c|}{\multirow{2}{*}{0.960}} & \multicolumn{1}{c|}{0.880} & \multicolumn{1}{c|}{0.990} & \multirow{2}{*}{0.970} \\ \cline{2-5} \cline{8-10} \cline{13-14}
 & bin2 & \multicolumn{1}{c|}{-} & \multicolumn{1}{c|}{0.992} & \multicolumn{1}{c|}{1.000} & \multicolumn{1}{c|}{} & \multicolumn{1}{c|}{} & \multicolumn{1}{c|}{-} & \multicolumn{1}{c|}{\textbf{0.992}} & \multicolumn{1}{c|}{1.000} & \multicolumn{1}{c|}{} & \multicolumn{1}{c|}{} & \multicolumn{1}{c|}{0.902} & \multicolumn{1}{c|}{0.951} &  \\ \hline
\end{tabular}
}
\end{table}

\begin{table}[!ht]
    \centering
    \captionsetup{labelformat=default,labelsep=colon,name=Supplementary Table}
    \caption{\label{result_CUCEC} Table of uncertainty and fairness evaluation for \textbf{CPTAC-UCEC} dataset. p-values are computed using the two-sided Wilcoxon rank-sum test to compare the grades and survival risks predicted using real and synthesized transcriptomic data, in addition to WSIs. \munc, \mcov, \cunc and \ccov refer to marginal uncertainty, marginal coverage, conditional uncertainty and conditional coverage, respectively. \textuparrow \ indicates the larger the better, and \textdownarrow \ indicates the smaller the better. Equivalent or close ($\pm 0.01$) performance using real and synthesized data is marked in \textbf{bold}. \\}
    \resizebox{\textwidth}{!}{
\begin{tabular}{|r|l|ccccccccccccc|}
\hline
\multicolumn{1}{|c|}{\multirow{3}{*}{Category}} & \multicolumn{1}{c|}{\multirow{3}{*}{Group}} & \multicolumn{13}{c|}{Gradation} \\ \cline{3-15} 
\multicolumn{1}{|c|}{} & \multicolumn{1}{c|}{} & \multicolumn{5}{c|}{Real} & \multicolumn{5}{c|}{Synthesized} & \multicolumn{3}{c|}{p-value \textuparrow} \\ \cline{3-15} 
\multicolumn{1}{|c|}{} & \multicolumn{1}{c|}{} & \multicolumn{1}{c|}{AUC \textuparrow} & \multicolumn{1}{c|}{\munc \textdownarrow} & \multicolumn{1}{c|}{\mcov \textuparrow} & \multicolumn{1}{c|}{\cunc \textdownarrow} & \multicolumn{1}{c|}{\ccov \textuparrow} & \multicolumn{1}{c|}{AUC \textuparrow} & \multicolumn{1}{c|}{\munc \textdownarrow} & \multicolumn{1}{c|}{\mcov \textuparrow} & \multicolumn{1}{c|}{\cunc \textdownarrow} & \multicolumn{1}{c|}{\ccov \textuparrow} & \multicolumn{1}{c|}{grade} & \multicolumn{1}{c|}{\munc} & \cunc \\ \hline
overall & - & \multicolumn{1}{c|}{0.593} & \multicolumn{1}{c|}{0.821} & \multicolumn{1}{c|}{0.984} & \multicolumn{1}{c|}{-} & \multicolumn{1}{c|}{-} & \multicolumn{1}{c|}{\textbf{0.593}} & \multicolumn{1}{c|}{\textbf{0.818}} & \multicolumn{1}{c|}{0.980} & \multicolumn{1}{c|}{-} & \multicolumn{1}{c|}{-} & \multicolumn{1}{c|}{1.000} & \multicolumn{1}{c|}{0.914} & - \\ \hline
magni- & level1 & \multicolumn{1}{c|}{0.595} & \multicolumn{1}{c|}{0.761} & \multicolumn{1}{c|}{0.988} & \multicolumn{1}{c|}{\multirow{3}{*}{0.805}} & \multicolumn{1}{c|}{\multirow{3}{*}{0.976}} & \multicolumn{1}{c|}{\textbf{0.595}} & \multicolumn{1}{c|}{\textbf{0.761}} & \multicolumn{1}{c|}{0.988} & \multicolumn{1}{c|}{\multirow{3}{*}{\textbf{0.805}}} & \multicolumn{1}{c|}{\multirow{3}{*}{0.976}} & \multicolumn{1}{c|}{1.000} & \multicolumn{1}{c|}{1.000} & \multirow{3}{*}{1.000} \\ \cline{2-5} \cline{8-10} \cline{13-14}
fication & level2 & \multicolumn{1}{c|}{0.601} & \multicolumn{1}{c|}{0.824} & \multicolumn{1}{c|}{0.971} & \multicolumn{1}{c|}{} & \multicolumn{1}{c|}{} & \multicolumn{1}{c|}{\textbf{0.600}} & \multicolumn{1}{c|}{\textbf{0.824}} & \multicolumn{1}{c|}{0.971} & \multicolumn{1}{c|}{} & \multicolumn{1}{c|}{} & \multicolumn{1}{c|}{1.000} & \multicolumn{1}{c|}{1.000} &  \\ \cline{2-5} \cline{8-10} \cline{13-14}
\multicolumn{1}{|l|}{} & level3 & \multicolumn{1}{c|}{0.593} & \multicolumn{1}{c|}{0.831} & \multicolumn{1}{c|}{0.971} & \multicolumn{1}{c|}{} & \multicolumn{1}{c|}{} & \multicolumn{1}{c|}{\textbf{0.592}} & \multicolumn{1}{c|}{\textbf{0.831}} & \multicolumn{1}{c|}{0.971} & \multicolumn{1}{c|}{} & \multicolumn{1}{c|}{} & \multicolumn{1}{c|}{1.000} & \multicolumn{1}{c|}{1.000} &  \\ \hline
vital & alive & \multicolumn{1}{c|}{0.575} & \multicolumn{1}{c|}{0.826} & \multicolumn{1}{c|}{0.982} & \multicolumn{1}{c|}{\multirow{2}{*}{0.717}} & \multicolumn{1}{c|}{\multirow{2}{*}{0.959}} & \multicolumn{1}{c|}{\textbf{0.575}} & \multicolumn{1}{c|}{\textbf{0.823}} & \multicolumn{1}{c|}{0.978} & \multicolumn{1}{c|}{\multirow{2}{*}{\textbf{0.715}}} & \multicolumn{1}{c|}{\multirow{2}{*}{0.957}} & \multicolumn{1}{c|}{1.000} & \multicolumn{1}{c|}{0.908} & \multirow{2}{*}{0.954} \\ \cline{2-5} \cline{8-10} \cline{13-14}
status & dead & \multicolumn{1}{c|}{0.728} & \multicolumn{1}{c|}{0.608} & \multicolumn{1}{c|}{0.937} & \multicolumn{1}{c|}{} & \multicolumn{1}{c|}{} & \multicolumn{1}{c|}{\textbf{0.728}} & \multicolumn{1}{c|}{\textbf{0.608}} & \multicolumn{1}{c|}{0.937} & \multicolumn{1}{c|}{} & \multicolumn{1}{c|}{} & \multicolumn{1}{c|}{1.000} & \multicolumn{1}{c|}{1.000} &  \\ \hline
\multirow{3}{*}{grade} & grade I & \multicolumn{1}{c|}{0.469} & \multicolumn{1}{c|}{0.973} & \multicolumn{1}{c|}{0.960} & \multicolumn{1}{c|}{\multirow{3}{*}{0.644}} & \multicolumn{1}{c|}{\multirow{3}{*}{0.928}} & \multicolumn{1}{c|}{\textbf{0.465}} & \multicolumn{1}{c|}{\textbf{0.973}} & \multicolumn{1}{c|}{0.960} & \multicolumn{1}{c|}{\multirow{3}{*}{\textbf{0.643}}} & \multicolumn{1}{c|}{\multirow{3}{*}{0.926}} & \multicolumn{1}{c|}{0.939} & \multicolumn{1}{c|}{1.000} & \multirow{3}{*}{0.929} \\ \cline{2-5} \cline{8-10} \cline{13-14}
 & grade II & \multicolumn{1}{c|}{0.454} & \multicolumn{1}{c|}{0.746} & \multicolumn{1}{c|}{0.956} & \multicolumn{1}{c|}{} & \multicolumn{1}{c|}{} & \multicolumn{1}{c|}{\textbf{0.454}} & \multicolumn{1}{c|}{\textbf{0.751}} & \multicolumn{1}{c|}{0.956} & \multicolumn{1}{c|}{} & \multicolumn{1}{c|}{} & \multicolumn{1}{c|}{0.952} & \multicolumn{1}{c|}{0.887} &  \\ \cline{2-5} \cline{8-10} \cline{13-14}
 & grade III & \multicolumn{1}{c|}{0.887} & \multicolumn{1}{c|}{0.213} & \multicolumn{1}{c|}{0.870} & \multicolumn{1}{c|}{} & \multicolumn{1}{c|}{} & \multicolumn{1}{c|}{\textbf{0.889}} & \multicolumn{1}{c|}{\textbf{0.206}} & \multicolumn{1}{c|}{0.863} & \multicolumn{1}{c|}{} & \multicolumn{1}{c|}{} & \multicolumn{1}{c|}{1.000} & \multicolumn{1}{c|}{0.899} &  \\ \hline
\multirow{2}{*}{time bin} & bin1 & \multicolumn{1}{c|}{0.496} & \multicolumn{1}{c|}{0.812} & \multicolumn{1}{c|}{0.983} & \multicolumn{1}{c|}{\multirow{2}{*}{0.841}} & \multicolumn{1}{c|}{\multirow{2}{*}{0.986}} & \multicolumn{1}{c|}{\textbf{0.503}} & \multicolumn{1}{c|}{\textbf{0.809}} & \multicolumn{1}{c|}{0.979} & \multicolumn{1}{c|}{\multirow{2}{*}{\textbf{0.840}}} & \multicolumn{1}{c|}{\multirow{2}{*}{0.984}} & \multicolumn{1}{c|}{1.000} & \multicolumn{1}{c|}{0.905} & \multirow{2}{*}{0.953} \\ \cline{2-5} \cline{8-10} \cline{13-14}
 & bin2 & \multicolumn{1}{c|}{0.608} & \multicolumn{1}{c|}{0.870} & \multicolumn{1}{c|}{0.989} & \multicolumn{1}{c|}{} & \multicolumn{1}{c|}{} & \multicolumn{1}{c|}{\textbf{0.608}} & \multicolumn{1}{c|}{\textbf{0.870}} & \multicolumn{1}{c|}{0.989} & \multicolumn{1}{c|}{} & \multicolumn{1}{c|}{} & \multicolumn{1}{c|}{1.000} & \multicolumn{1}{c|}{1.000} &  \\ \hline \hline
\multicolumn{1}{|c|}{\multirow{3}{*}{Category}} & \multicolumn{1}{c|}{\multirow{3}{*}{Group}} & \multicolumn{13}{c|}{Risk Estimation} \\ \cline{3-15} 
\multicolumn{1}{|c|}{} & \multicolumn{1}{c|}{} & \multicolumn{5}{c|}{Real} & \multicolumn{5}{c|}{Synthesized} & \multicolumn{3}{c|}{p-value \textuparrow} \\ \cline{3-15} 
\multicolumn{1}{|c|}{} & \multicolumn{1}{c|}{} & \multicolumn{1}{c|}{C Index \textuparrow} & \multicolumn{1}{c|}{\munc \textdownarrow} & \multicolumn{1}{c|}{\mcov \textuparrow} & \multicolumn{1}{c|}{\cunc \textdownarrow} & \multicolumn{1}{c|}{\ccov \textuparrow} & \multicolumn{1}{c|}{C Index \textuparrow} & \multicolumn{1}{c|}{\munc \textdownarrow} & \multicolumn{1}{c|}{\mcov \textuparrow} & \multicolumn{1}{c|}{\cunc \textdownarrow} & \multicolumn{1}{c|}{\ccov \textuparrow} & \multicolumn{1}{c|}{risk} & \multicolumn{1}{c|}{\munc} & \cunc \\ \hline
overall & - & \multicolumn{1}{c|}{0.533} & \multicolumn{1}{c|}{1.000} & \multicolumn{1}{c|}{0.908} & \multicolumn{1}{c|}{-} & \multicolumn{1}{c|}{-} & \multicolumn{1}{c|}{\textbf{0.530}} & \multicolumn{1}{c|}{\textbf{1.000}} & \multicolumn{1}{c|}{0.908} & \multicolumn{1}{c|}{-} & \multicolumn{1}{c|}{-} & \multicolumn{1}{c|}{0.932} & \multicolumn{1}{c|}{0.918} & - \\ \hline
magni- & level1 & \multicolumn{1}{c|}{0.393} & \multicolumn{1}{c|}{0.999} & \multicolumn{1}{c|}{0.912} & \multicolumn{1}{c|}{\multirow{3}{*}{0.999}} & \multicolumn{1}{c|}{\multirow{3}{*}{0.920}} & \multicolumn{1}{c|}{\textbf{0.387}} & \multicolumn{1}{c|}{\textbf{0.999}} & \multicolumn{1}{c|}{0.912} & \multicolumn{1}{c|}{\multirow{3}{*}{\textbf{0.999}}} & \multicolumn{1}{c|}{\multirow{3}{*}{0.920}} & \multicolumn{1}{c|}{0.891} & \multicolumn{1}{c|}{0.997} & \multirow{3}{*}{0.950} \\ \cline{2-5} \cline{8-10} \cline{13-14}
fication & level2 & \multicolumn{1}{c|}{0.594} & \multicolumn{1}{c|}{1.000} & \multicolumn{1}{c|}{0.941} & \multicolumn{1}{c|}{} & \multicolumn{1}{c|}{} & \multicolumn{1}{c|}{\textbf{0.592}} & \multicolumn{1}{c|}{\textbf{1.000}} & \multicolumn{1}{c|}{0.941} & \multicolumn{1}{c|}{} & \multicolumn{1}{c|}{} & \multicolumn{1}{c|}{0.975} & \multicolumn{1}{c|}{0.925} &  \\ \cline{2-5} \cline{8-10} \cline{13-14}
\multicolumn{1}{|l|}{} & level3 & \multicolumn{1}{c|}{0.586} & \multicolumn{1}{c|}{1.000} & \multicolumn{1}{c|}{0.906} & \multicolumn{1}{c|}{} & \multicolumn{1}{c|}{} & \multicolumn{1}{c|}{\textbf{0.583}} & \multicolumn{1}{c|}{\textbf{1.000}} & \multicolumn{1}{c|}{0.906} & \multicolumn{1}{c|}{} & \multicolumn{1}{c|}{} & \multicolumn{1}{c|}{0.980} & \multicolumn{1}{c|}{0.927} &  \\ \hline
vital & alive & \multicolumn{1}{c|}{-} & \multicolumn{1}{c|}{1.000} & \multicolumn{1}{c|}{0.899} & \multicolumn{1}{c|}{\multirow{2}{*}{0.582}} & \multicolumn{1}{c|}{\multirow{2}{*}{0.934}} & \multicolumn{1}{c|}{\textbf{-}} & \multicolumn{1}{c|}{\textbf{1.000}} & \multicolumn{1}{c|}{0.895} & \multicolumn{1}{c|}{\multirow{2}{*}{\textbf{0.583}}} & \multicolumn{1}{c|}{\multirow{2}{*}{0.932}} & \multicolumn{1}{c|}{0.926} & \multicolumn{1}{c|}{0.993} & \multirow{2}{*}{0.695} \\ \cline{2-5} \cline{8-10} \cline{13-14}
status & dead & \multicolumn{1}{c|}{0.509} & \multicolumn{1}{c|}{0.165} & \multicolumn{1}{c|}{0.968} & \multicolumn{1}{c|}{} & \multicolumn{1}{c|}{} & \multicolumn{1}{c|}{\textbf{0.510}} & \multicolumn{1}{c|}{\textbf{0.166}} & \multicolumn{1}{c|}{0.968} & \multicolumn{1}{c|}{} & \multicolumn{1}{c|}{} & \multicolumn{1}{c|}{0.996} & \multicolumn{1}{c|}{0.397} &  \\ \hline
\multirow{3}{*}{grade} & grade I & \multicolumn{1}{c|}{0.471} & \multicolumn{1}{c|}{1.000} & \multicolumn{1}{c|}{0.949} & \multicolumn{1}{c|}{\multirow{3}{*}{0.999}} & \multicolumn{1}{c|}{\multirow{3}{*}{0.921}} & \multicolumn{1}{c|}{0.448} & \multicolumn{1}{c|}{\textbf{1.000}} & \multicolumn{1}{c|}{0.960} & \multicolumn{1}{c|}{\multirow{3}{*}{\textbf{0.999}}} & \multicolumn{1}{c|}{\multirow{3}{*}{0.924}} & \multicolumn{1}{c|}{0.871} & \multicolumn{1}{c|}{0.733} & \multirow{3}{*}{0.909} \\ \cline{2-5} \cline{8-10} \cline{13-14}
 & grade II & \multicolumn{1}{c|}{0.507} & \multicolumn{1}{c|}{1.000} & \multicolumn{1}{c|}{0.933} & \multicolumn{1}{c|}{} & \multicolumn{1}{c|}{} & \multicolumn{1}{c|}{\textbf{0.507}} & \multicolumn{1}{c|}{\textbf{1.000}} & \multicolumn{1}{c|}{0.933} & \multicolumn{1}{c|}{} & \multicolumn{1}{c|}{} & \multicolumn{1}{c|}{0.994} & \multicolumn{1}{c|}{0.996} &  \\ \cline{2-5} \cline{8-10} \cline{13-14}
 & grade III & \multicolumn{1}{c|}{0.445} & \multicolumn{1}{c|}{0.996} & \multicolumn{1}{c|}{0.879} & \multicolumn{1}{c|}{} & \multicolumn{1}{c|}{} & \multicolumn{1}{c|}{\textbf{0.444}} & \multicolumn{1}{c|}{\textbf{0.996}} & \multicolumn{1}{c|}{0.879} & \multicolumn{1}{c|}{} & \multicolumn{1}{c|}{} & \multicolumn{1}{c|}{0.930} & \multicolumn{1}{c|}{0.999} &  \\ \hline
\multirow{2}{*}{time bin} & bin1 & \multicolumn{1}{c|}{0.527} & \multicolumn{1}{c|}{1.000} & \multicolumn{1}{c|}{0.915} & \multicolumn{1}{c|}{\multirow{2}{*}{0.895}} & \multicolumn{1}{c|}{\multirow{2}{*}{0.957}} & \multicolumn{1}{c|}{\textbf{0.527}} & \multicolumn{1}{c|}{\textbf{1.000}} & \multicolumn{1}{c|}{0.915} & \multicolumn{1}{c|}{\multirow{2}{*}{\textbf{0.895}}} & \multicolumn{1}{c|}{\multirow{2}{*}{0.952}} & \multicolumn{1}{c|}{0.988} & \multicolumn{1}{c|}{0.999} & \multirow{2}{*}{0.950} \\ \cline{2-5} \cline{8-10} \cline{13-14}
 & bin2 & \multicolumn{1}{c|}{-} & \multicolumn{1}{c|}{0.791} & \multicolumn{1}{c|}{1.000} & \multicolumn{1}{c|}{} & \multicolumn{1}{c|}{} & \multicolumn{1}{c|}{\textbf{-}} & \multicolumn{1}{c|}{\textbf{0.790}} & \multicolumn{1}{c|}{0.989} & \multicolumn{1}{c|}{} & \multicolumn{1}{c|}{} & \multicolumn{1}{c|}{0.901} & \multicolumn{1}{c|}{0.901} &  \\ \hline
\end{tabular}
}
\end{table}

\begin{table}[!ht]
    \centering
    \captionsetup{labelformat=default,labelsep=colon,name=Supplementary Table}
    \caption{\label{table_deviation} Table of expected deviation from the desired coverage for a given number of samples in the calibration set and conformal error rate $\alpha=0.1$.  ncal denotes the calibration set size, and $\epsilon$ denotes the deviation. The less the calibration set size, the greater the expected deviation.\\}
\begin{tabular}{|c|c||c|c||c|c||c|c||c|c||c|c|}
\hline
\textbf{ncal} & \textbf{$\epsilon$} & \textbf{ncal} & \textbf{$\epsilon$} & \textbf{ncal} & \textbf{$\epsilon$} & \textbf{ncal} & \textbf{$\epsilon$} & \textbf{ncal} & \textbf{$\epsilon$} & \textbf{ncal} & \textbf{$\epsilon$} \\ \hline \hline
9 & 0.183 & 33 & 0.079 & 67 & 0.060 & 104 & 0.047 & 194 & 0.035 & 373 & 0.025 \\ \hline
12 & 0.121 & 34 & 0.076 & 69 & 0.064 & 111 & 0.048 & 201 & 0.036 & 375 & 0.026 \\ \hline
13 & 0.106 & 36 & 0.077 & 72 & 0.058 & 120 & 0.047 & 204 & 0.034 & 399 & 0.026 \\ \hline
18 & 0.097 & 37 & 0.078 & 75 & 0.055 & 126 & 0.044 & 207 & 0.035 & 447 & 0.024 \\ \hline
21 & 0.107 & 38 & 0.078 & 78 & 0.057 & 132 & 0.043 & 225 & 0.033 & 453 & 0.023 \\ \hline
24 & 0.085 & 39 & 0.087 & 80 & 0.058 & 138 & 0.043 & 228 & 0.034 & 501 & 0.022 \\ \hline
25 & 0.086 & 42 & 0.074 & 81 & 0.056 & 141 & 0.043 & 231 & 0.033 & 510 & 0.022 \\ \hline
27 & 0.087 & 43 & 0.071 & 82 & 0.054 & 144 & 0.040 & 252 & 0.031 & 531 & 0.022 \\ \hline
28 & 0.087 & 48 & 0.071 & 84 & 0.052 & 150 & 0.042 & 270 & 0.031 & 550 &  0.022 \\ \hline
29 & 0.102 & 51 & 0.071 & 86 & 0.053 & 151 & 0.041 & 273 & 0.030 & 555 & 0.021 \\ \hline
30 & 0.095 & 54 & 0.063 & 87 & 0.053 & 153 & 0.039 & 279 & 0.031 & 582 & 0.021 \\ \hline
31 & 0.089 & 55 & 0.063 & 90 & 0.055 & 165 & 0.038 & 282 & 0.030 & 693 & 0.019 \\ \hline
32 & 0.084 & 57 & 0.065 & 91 & 0.053 & 170 & 0.039 & 288 & 0.030 & 700 & 0.019 \\ \hline
33 & 0.079 & 60 & 0.067 & 93 & 0.050 & 171 & 0.039 & 300 & 0.029 & 780 & 0.018 \\ \hline
34 & 0.076 & 63 & 0.060 & 99 & 0.053 & 177 & 0.038 & 309 & 0.029 & 918 & 0.017 \\ \hline
32 & 0.084 & 66 & 0.060 & 102 & 0.049 & 192 & 0.036 & 339 & 0.028 & 1119 & 0.015 \\ \hline
\end{tabular}
\end{table}


\end{document}